\documentclass[lettersize,journal]{IEEEtran}
\usepackage{amsmath,amsfonts}
\usepackage{algorithmic}
\usepackage{algorithm}
\usepackage{array}
\usepackage[caption=false,font=normalsize,labelfont=sf,textfont=sf]{subfig}
\usepackage{textcomp}
\usepackage{stfloats}
\usepackage{url}
\usepackage{verbatim}
\usepackage{graphicx}
\usepackage{cite}
\ifCLASSINFOpdf
\else
\fi
\usepackage{caption}
\usepackage{subcaption}
\usepackage{xcolor, colortbl}

\usepackage{enumerate}
\usepackage{hyperref}
\hypersetup{hidelinks}
\usepackage{amssymb, latexsym}
\usepackage{color}
\usepackage{float}
\usepackage{mathtools}
\usepackage[algo2e,ruled,noend,linesnumbered]{algorithm2e}
\usepackage[font={small,it}]{caption}
\usepackage{makecell}
\usepackage{adjustbox}
\usepackage{placeins}
\usepackage{multirow}
\usepackage{booktabs}

\newcommand{\norm}[1]{\left\lVert#1\right\rVert}

\begin{document}

\title{Kernel Reboot: Breaking the Boundaries of Neural Tangent Kernels for Neural Fields}

\author{Amir~Mallak$^{1}$, Alaa~Maalouf$^{1,2}$, Lior~Wolf$^{3}$, Daniela~Rus$^{2}$ and~Dan~Rosenbaum$^{1}$\\
\small $^{1}$Department of Computer Science, University of Haifa\\
\small $^{2}$Computer Science and Artificial Intelligence Laboratory, Massachusetts Institute of Technology\\
\small $^{3}$School of Computer Science and AI, Tel Aviv University
\thanks{Corresponding author A. Mallak: (mallak002@gmail.com)}}

\markboth{IEEE Transactions on Pattern Analysis and Machine Intelligence, Vol. 48, No. 9, September 2026}%
{Mallak \MakeLowercase{\textit{et al.}}: Kernel Reboot: Breaking the Boundaries of Neural Tangent Kernels for Neural Fields}

\IEEEpubid{\parbox{\textwidth}{\centering\scriptsize
\textcopyright~2026 IEEE. Personal use of this material is permitted. Permission from IEEE must be obtained for all other uses, in any current or future media, including reprinting/republishing this material for advertising or promotional purposes, creating new collective works, for resale or redistribution to servers or lists, or reuse of any copyrighted component of this work in other works. DOI: 10.1109/TPAMI.2026.3692624.}}

\maketitle

\begin{abstract}
Neural fields (NFs) map continuous coordinates to signals such as color or density, but fast high-quality reconstruction from sparse observations remains difficult. Classical Neural Tangent Kernel (NTK) regression gives closed-form fits, yet it is fundamentally linear and cannot accumulate reusable task priors. We develop three algorithms that address these gaps. NTK-KIP learns a distilled support set of coordinates (and optional labels) so that a finite NTK can inpaint large missing regions from little observed data, yielding a compact non-linear representation instead of a raw kernel solve. MetaQuill meta-learns a shared initialization for an INR so that new scenes can be adapted by updating only a small task-specific weight offset, which provides true feature learning and a reusable prior. Finally, MetaQuill-KIP fuses both ideas: it seeds the task with a KIP-style non-linear warm start, then refines only that small offset around the meta-learned initialization. MetaQuill-KIP achieves high-PSNR reconstructions and semantically plausible inpainting under very sparse observations, while requiring only lightweight per-instance adaptation, whereas diffusion-style baselines typically depend on large pretrained generative priors and costly per-image tuning. This shows that NTK-driven neural fields can be made both non-linear and meta-learnable, narrowing the gap between analytic kernels and practical few-shot reconstruction.
\end{abstract}

\begin{IEEEkeywords}
Machine learning, Representations, data structures, and transforms, Vision and Scene Understanding, Computer vision, Knowledge retrieval, Neural nets.
\end{IEEEkeywords}
\IEEEpeerreviewmaketitle

\section{Introduction} \label{Introduction}
\IEEEPARstart{N}{eural} field models (NFs), also known as implicit neural representations, are continuous functions that map input positional coordinates—such as spatial or temporal coordinates—directly to outputs representing properties of a field, such as color, density, or geometry. Specifically, NFs are neural networks designed to learn this mapping~\cite{stanley2007compositional, ha2016generating}. These models are particularly useful for tasks such as image representation~\cite{ha2016generating}, 3D object reconstruction~\cite{mescheder2019occupancy, chen2019learning}, 3D scenes~\cite{mildenhall2021nerf}, audio~\cite{sitzmann2020implicit}, and more~\cite{dupont2021generative}. Typically, the model does not use spatial coordinates directly as input; instead, it leverages \textit{positional encodings} (PEs) of these coordinates~\cite{mildenhall2021nerf, tancik2020fourier}, providing a continuous input space. This approach enables the model to represent complex, high-resolution fields in a memory-efficient manner, without relying on the discretized grid representations typical in traditional models.

Critical to the success of NFs is their ability to perform representation learning, enabling them to capture useful features for various downstream tasks~\cite{dupont2022data, bauer2023spatial}. This learning process is highly non-linear and challenging to model~\cite{bauer2023spatial}. While NFs have demonstrated considerable potential, their use presents two main challenges: first, representing data as neural network weights is not inherently compatible with downstream tasks and restricts the network size; second, the training process must be repeated for each data sample, which impacts efficiency.

\noindent\textbf{Neural Tangent Kernels (NTKs).} NTKs~\cite{jacot2018neural} aim to model the learning dynamics of neural networks, explaining their evolution during training via gradient descent through kernel methods. 
The NTK approximates the dynamics of wide neural networks by their first-order Taylor expansion around their initialization. At infinite width (under suitable parameter scaling), this approximation is exact, and at finite widths, it has proven to be a useful model for network training, showing applications in enabling kernel-based dataset distillation~\cite{nguyen2021dataset,loo2022efficient,maalouf2023on}, incremental learning~\cite{liu2024ntk}, regression~\cite{qadeer2023efficient}, meta-learning \cite{zhou2021meta, yang2020feature}, and more~\cite{zhang2024improving}.

Formally, let $\mathit{N}>0$ be the total number of samples in the dataset, $d$ be the dimension of each sample, and let $x,x'\in \mathbb{R}^{d}$ be a pair of input samples. Let $\mathit{f}(\mathit{x}; \theta): \mathbb{R}^d\, \rightarrow\, \mathbb{R}^C\,$ be a fully connected neural network consisting of an arbitrary number of different-sized layers with a set of parameters $\mathit{\theta}$, and let $L(\hat{y}, y)$ be the loss function of predicted labels $\hat{y}$ with respect to ground-truth labels $y$.
The NTK derives that the evolution of $\mathit{f}$ outputs over time is represented by,
\begin{align*}    
\nabla{_t}f(\mathit{x'}; \theta) = -\frac{1}{N}\sum_{\mathit{i}=\mathit{1}}^{\mathit{n}} \mathcal{K}(x, x' ; \theta) \nabla{_f}L(f(x_i,\, \theta),\, y_i)
\end{align*}

Where,
 \begin{equation} \label{NN_Evolution_Kernel}
 \begin{split}
\mathcal{K}(x, x' ; \theta) =& \nabla{_\theta}f(\mathit{x}\, ;\, \theta)^T \ \nabla{_\theta}f(x' ;\, \theta)\ \\
&|\ \mathcal{K}: \mathbb{R}^d \, \times\,  \mathbb{R}^d \ \rightarrow{} \ \mathbb{R}^{\mathit{C} \times \mathit{C}},
 \end{split}
\end{equation}

\noindent\textbf{From infinite width neural network to kernel ridge regression.} For infinite-width neural networks under the correct parameter scaling, the NTK $\mathcal{K}(\mathit{x}, x' ; \theta)$ remains constant over time \cite{jacot2018neural}, allowing the time-dependent kernel to be replaced by a constant kernel $\mathcal{K}(x, x')$. Furthermore, for multi-class networks, the kernels are the same for each class and can be treated as $C$ independent kernels each with equal value. Furthermore,  if a network is trained under MSE loss with $N$ input data points, each with $C$ labels, this leads to a closed-form solution as Kernel-Ridge Regression. \\
\\
\\We have the full kernel matrix $\mathcal{K} \in \mathbb{R}^{N \times N}$, whose entries are the corresponding $\mathcal{K}(x_i, x_j)$ at positions $\mathcal{K}_{i, j}$ with target $Y \in \mathbb{R}^{N \times C}$. We seek an optimal parameter $\mathit{W}$ that minimizes the empirical loss. Formally, $W := \underset{W^{'}\in \mathbb{R}^{N\times C}}{\arg \min} ||\mathit{Y} - \mathit{K} W^{'}||_2^2$.  Solving for the optimal $W^{'}$:
\begin{equation}
    \label{W_Vector}
    \mathcal{W} = \mathcal{K}^{-1} \mathcal{Y}
\end{equation}

Predicting the output on a new data point $x'$ is found by compute the kernel with respect to the training dataset $\mathcal{K}(x', x) \in \mathbb{R}^{1 \times N}$, with entries at position $i$ given by $\mathcal{K}(x', x_i)$. The output is then given by

\begin{equation}
\label{Y_Reconstruction}
    \mathcal{\hat{Y}} = \mathcal{K}(x', x) \mathcal{W}
\end{equation}

This approach is useful not only as a model of neural network training, but also shortcuts the need of training a model, which can be faster, particular for very large models, at the cost of inverting the kernel matrix.

\noindent\textbf{NTK for representation learning.} Motivated by the advantages of NTK, we investigate the usage of NTKs as a model for training Neural Fields. Specifically, performing Representation Learning with NFs on a dataset requires training multiple NFs where each is a representation of a single data sample, thus, intuitively, these multi-NF models could be replaced and presented via the NTK method as a single Kernel matrix.  Moreover, this allows NFs to leverage all of  NTKs notable advantages, such as the use of infinite-width layers (providing a high-dimensional parameter space and model flexibility), closed-form kernel regression (making training and inference as straightforward as matrix inversion), and utility in certain representation learning applications (such as training multi-neural network models).

\noindent\textbf{NTK NF Matrix Completion (MC).} Neural Fields are often used for inpainting/matrix completion tasks. In this case, the model is given incomplete data, for example, only a subset of pixels of an image, and the model has to inpaint the missing pixels. For MC \cite{radhakrishnan2022simple} tasks, the algorithm for reconstructing the target image is slightly different due to Kernel matrix calculation for missing pixels.
Denoting the target image's observed (known) pixels by $o$, and assuming the ground-truth (GT) image's shape is $(h, w, C)$. For calculating the $W \in \mathbb{R}^{o \times C}$ vector,
\begin{equation}
    \label{W_Vector_MC}
    \mathcal{W} = \mathcal{K}_{train}^{-1} \mathcal{Y}\quad |\, \mathcal{K}_{train} \in \mathbb{R}^{o \times o},\, \mathcal{Y} \in \mathbb{R}^{o \times C}
\end{equation}

Where $\mathcal{K}_{train}$ takes as input the model's weights and the NF observed PEs (only the PEs that correspond to the observed target image's pixels).
For reconstructing the GT image,
\begin{equation}
    \label{Y_Reconstruct_MC}
    \mathcal{\hat{Y}} = \mathcal{K}_{test} \mathcal{W}\quad |\, \mathcal{K}_{test} \in \mathbb{R}^{(h \cdot w) \times o},\, \mathcal{\hat{Y}} \in \mathbb{R}^{(h \cdot w) \times C}
\end{equation}

Where $\mathcal{K}_{test}$ takes the model's weights as input, along with the observed and complete PEs of the NF.

\noindent\textbf{Why improve NTK instead of abandoning it.}
NTK-style neural field solvers remain attractive because they provide analytic, closed-form control of fitting a signal, extremely fast per-instance adaptation, and transparent conditioning on observed coordinates, all without requiring a large pretrained generative prior. This makes them appealing for settings such as on-device personalization or single-scene reconstruction, where we cannot assume access to a massive diffusion backbone or a text semantic prior. The problem is that classical NTK pipelines are fundamentally limited: they are linear in function space (so they cannot represent rich structure), and they cannot meta-learn reusable features across tasks. The central objective of this work is therefore not to argue that NTK alone beats modern diffusion models, but to demonstrate that these two structural gaps in NTK can be closed in practice. NTK-KIP injects nonlinear representational capacity by distilling a compact set of task-specific inducing points, and MetaQuill supplies true feature learning and meta-initialization so that new tasks can adapt in a few steps. Their fusion, MetaQuill-KIP, unifies both effects and turns NTK-style neural fields into a practical, fast-adapting, per-instance reconstructor.

\subsection{Our Contributions} \label{Our_Contributions}
We study the practical use of Neural Tangent Kernels (NTKs) for representation learning with neural fields (NFs), and we ask whether NTK-style methods can be made fast, learnable, and useful for single-image tasks such as reconstruction and inpainting. We begin by replacing slow NF training with NTK regression, which in principle provides a one-step solution for fitting an (effectively) infinite-width NF. Our initial analysis reveals that the standard NTK pipeline, when applied directly to neural fields, suffers from fundamental shortcomings. Our first contribution is a systematic study of these limitations:
\begin{itemize}
    \item \textbf{NTK representational limitations}: Because NTK fits a linear function in weight space around a fixed initialization, its predictions are structurally restricted. This prevents reconstruction of high-frequency structure and fine semantic detail in many real signals (see Section~\ref{NTK_Natural_Limitations}).
    \item \textbf{NTK lack of robustness}: The explicit kernel regression solution is highly sensitive to noise and masking, and it struggles in partially observed settings like inpainting, where even modest missing regions degrade output smoothness and semantic plausibility (see Section~\ref{NTK_Robustness}).
    \item \textbf{NTK inefficiency}: Finite-width NTK on dense neural field grids requires forming or applying large Jacobian blocks. This leads to high memory use and non-trivial solve cost, which makes naive NTK unattractive for practical, iterative adaptation (see Section~\ref{NTK_Functa}).
\end{itemize}
To address these issues, we develop and evaluate three NTK-driven algorithms:
\begin{enumerate}
    \item \textbf{NTK-KIP: Distilled nonlinear NTK representation.}  
    We introduce Neural Tangent Kernels Kernel Inducing Points (NTK-KIP), which converts NTK fitting from ``solve on every pixel'' into ``learn a small support set and solve only there.'' NTK-KIP learns a compact set of spatial support coordinates and (optionally) their associated target values, by optimizing them through the NTK. This amounts to distilling the signal and its supervision into a handful of inducing points. As a result, NTK is no longer forced to behave as a strictly linear regressor over the raw grid: instead, the learned support set itself becomes nonlinear and task-adaptive. In experiments, NTK-KIP achieves aggressive compression while preserving reconstruction quality. For example, using Fourier positional encodings, we reconstruct RGB Flowers images with $\approx 20\%$ of the pixels observed at $\text{PSNR} \approx 31$ dB, and with only $\approx 5\%$ observed at $\text{PSNR} \approx 20$ dB. With raw positional encodings, we see $\approx 80\%$ compression at $\text{PSNR} > 36$ dB and $\approx 50\%$ at $\text{PSNR} > 23$ dB (See Fig. \ref{NF_KIP_PEs_experiment}). NTK-KIP therefore addresses two core NTK weaknesses: it injects nonlinearity via learned supports, and it reduces kernel cost by collapsing the solve to a tiny subset. However, NTK-KIP by itself does not provide task-transfer or shared feature learning across images.
    \item \textbf{MetaQuill: NTK-based meta-learning with feature learning and fast adaptation.}  
    We propose MetaQuill, a meta-learning procedure for neural fields that learns a reusable shared initialization $\theta_S$ across a small training set of tasks, together with a per-task low-rank update $\Delta \theta_i$. Unlike classical MAML, MetaQuill does not require an expensive inner-loop optimization during meta-training. Instead, MetaQuill directly learns both $\theta_S$ and the structure of the task-specific updates in a way that is compatible with NTK linearization. At test time, a new task can be adapted in tens of gradient steps (or even a single step in some settings), rather than training a fresh NF from scratch. MetaQuill enables two properties that standard NTK lacks: (i) reusable features that transfer across tasks, and (ii) data-efficient per-task adaptation. On MNIST-style digit fields, MetaQuill reconstructs new digits at $\text{PSNR} \approx 20$ dB after only $\approx 50$ update steps on the task-specific $\Delta \theta_i$ (See Table~\ref{tab:mnist_combined_runtime}), while a naive ``Single-INR'' trained from scratch requires similar or more optimization effort just to approach that quality. MetaQuill, however, is still limited by NTK linearity during inference, and its tangent-only adaptation can struggle with large missing regions (e.g., hole inpainting).
    \item \textbf{MetaQuill-KIP: Fusing nonlinear NTK-KIP priors with MetaQuill-style fast adaptation.}
    Our final contribution is a new hybrid method that unifies the strengths of NTK-KIP and MetaQuill. MetaQuill-KIP first runs a KIP-style support optimization on a single target image, producing a distilled support set and an initial task-specific update direction $\Delta \theta_i$. It then refines only that low-dimensional $\Delta \theta_i$ nonlinearly, while keeping the shared MetaQuill initialization $\theta_S$ fixed. This gives us (i) nonlinear, structure-aware inpainting and reconstruction from NTK-KIP, and (ii) the rapid, few-step adaptation and reusable feature backbone inherited from MetaQuill. In practice, MetaQuill-KIP delivers high-fidelity reconstructions and semantically plausible inpainting within a few hundred lightweight refinement steps, and does so without relying on a large pretrained diffusion prior.
\end{enumerate}
Finally, we validate all three methods, and we compare them both internally and against modern diffusion-based inpainting systems. On MNIST, MetaQuill-KIP reaches over $32$ dB PSNR after only $50$ refinement steps, substantially improving over plain kernel regression (about $11$ dB) and over MetaQuill's tangent-only update (about $18$--$19$ dB) (See T. \ref{tab:mnist_combined_runtime}). On RGB Flowers images, MetaQuill-KIP produces visually coherent hole fills from sparse masks, outperforming naive single-image baselines in both PSNR and perceptual quality while requiring only a few seconds of adaptation for the new task. We also benchmark against Stable Diffusion (SD) inpainting, StrDiffusion~\cite{Liu_2024_CVPR}, Denoising Diffusion Probabilistic Models (DDPM), and a global structure-guided diffusion model (GSDM)~\cite{zhu2024global}. Although diffusion models achieve strong PSNR when given large observed regions and pretrained semantic priors, our NTK-driven pipelines operate in a fundamentally different and complementary regime: they adapt from scratch to a single image in seconds, using only per-image supervision, without external text conditioning or massive pretraining. This establishes MetaQuill-KIP as an effective bridge between theoretical NTK structure and practical single-image neural field reconstruction and inpainting.
\noindent\textbf{Notations. } For readability, we provide a consolidated list of acronyms and abbreviations in the Appendix (List of Acronyms and Abbreviations) - See Table ~\ref{tab:acronym_list}.

\begin{figure}[t]
    \centering
    \includegraphics[width=0.45\textwidth]{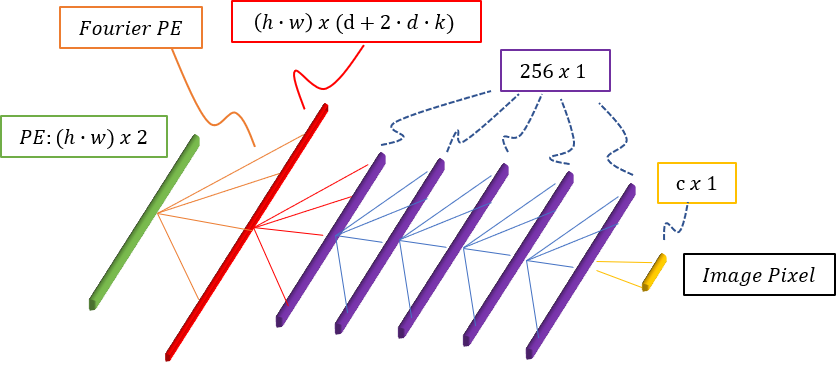}
        \caption{Neural Field architecture used throughout the paper. Input is a 2D coordinate $(a,b)\in[-1,1]^2$, encoded either by raw positional encoding (RPE) (Green) or Fourier positional encoding (FPE) (Orange) with $k$ frequency bands. The encoded vector (Red) is passed through a 5 layer MLP of width 256 with nonlinear activations (Purple) to produce an RGB output in $\mathbb{R}^3$ (or a scalar output in $\mathbb{R}$ for single channel signals) (Yellow).}
    \label{NF_Architecture_with_FPE}
\end{figure}

\section{Investigating the use of NTK in NFs} \label{Investigating the use of NTK in NFs}
We start by investigating the direct use of NTKs to release the long learning process in NFs.

\noindent\textbf{The setting. } Inspired by Functa ~\cite{dupont2022data}, throughout the experiments, we use the NF architecture described in Fig. \ref{NF_Architecture_with_FPE}, which consists of 2D spatial PEs, or alternatively Fourier positional encodings (FPEs), $\mathit{5}$ hidden layers (Each of width $\mathit{256}$ neurons), and one RGB or a Single-value output layer.
Unless otherwise specified, we use FPE with frequency bands set to $k_{bands} = 20$ as the default.
This value was determined from an experiment set to find the optimal FPE band; Figs.~\ref{Optimal_K_Band} and~\ref{Elbow_K_Band_Graph} in the appendix illustrates the effect of different frequency bands $\mathit{k}$. See Appendix~\ref{Ablation} for more details.

\begin{figure*}[!t]
    \centering
        \includegraphics[width=0.98\textwidth]{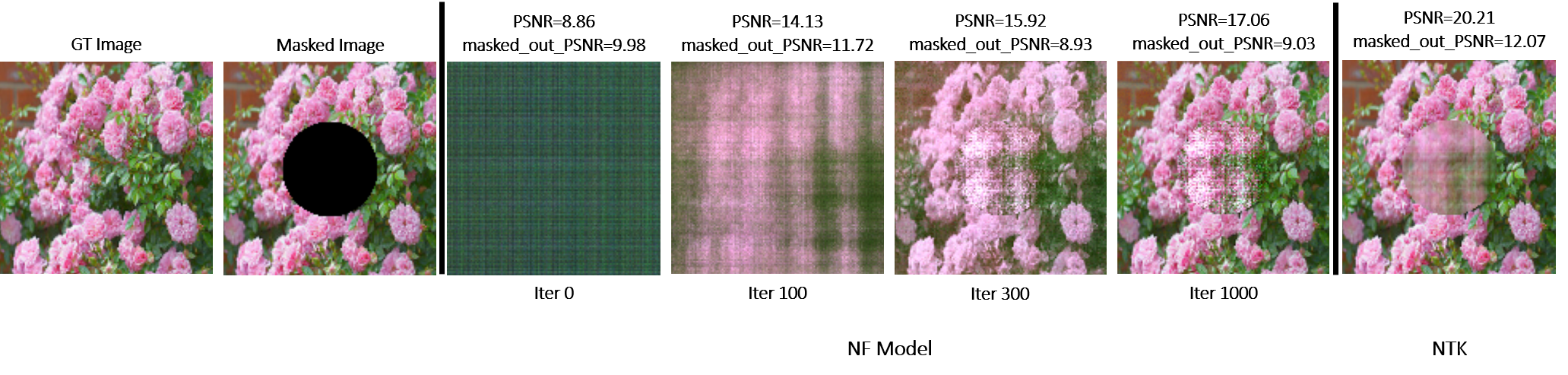}
        \caption{Inpainting experiment. From left to right: ground truth image, masked image, NF at iterations $0$, $100$, $300$, $1000$ (PSNR $\approx 17.06$ dB, masked PSNR $\approx 9.03$ dB), and Infinite NTK reconstruction (PSNR $\approx 20.21$ dB, masked PSNR $\approx 12.07$ dB).}
    \label{Inpainting_NTK_NF_Iter_1000}
\end{figure*}

\subsection{Exploring NTK-based NFs  in Representation Rigidity and Inpainting} \label{NTK_Natural_Limitations}
To investigate the capabilities and limitations of NTK-based NFs, we delve into how NTK performs in capturing and representing complex data structures.
We compare traditional finite NF to an NTK-based NF.

\begin{figure}[t]
    \centering
        \includegraphics[width=0.48\textwidth]{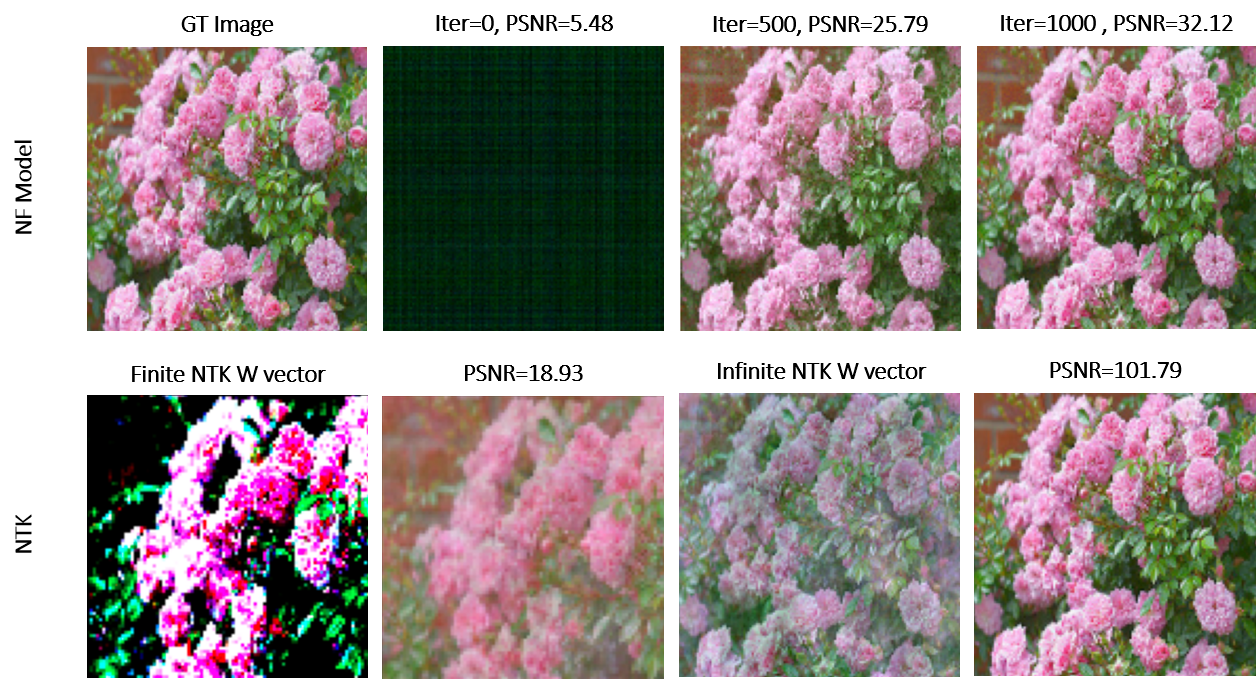}
    \caption{Representation rigidity experiment. Row 1: Neural Field training snapshots (ground truth, then reconstructions at iterations $0$, $500$, $1000$), PSNR=$32.12$ dB. Row 2, left to right: finite time NTK $W$, finite time NTK reconstruction (PSNR=$18.93$ dB), infinite NTK $W$, infinite NTK reconstruction (PSNR=$101.79$ dB).}
    \label{NF_NTK_Representation_Rigidity}
\end{figure}

\noindent\textbf{Representation Rigidity: NF vs NTK-based NF. }
We begin by examining NTK's representational rigidity ("inpainting" a fully known image). With all labels known, the Kernel can accurately capture pixel correlations. Thus, we test the infinite NTK method and compare it to standard neural network training. Specifically, we train the NF model in Fig.~\ref{NF_Architecture_with_FPE} for $\mathit{1000}$ steps with a learning rate of $\mathit{10}^{-3}$. For the NTK comparison, we use the NTK reconstruction method while the observed pixels here are the whole image pixels $(h \times w)$. Here, the PE vector and the target image are fully known.
Formally, let $\mathit{N} = \mathit{h} \cdot \mathit{w}$ be the total number of pixels, $a,\, b \in [-1,\, 1]$ be a pair of values representing a spatial 2D coordinate (NF input), and let $x, y, z \in [0,\, 255]$ be the target image's RGB values accordingly. The PE vector is then represented by $
\big((\mathit{a}_i, \mathit{b}_i)\big)^N_{i=1}
$, and the target image,
$\big((\mathit{x}_i, \mathit{y}_i, \mathit{z}_i)\big)^{N}_{i=1}$.

We chose the single flower image for this experiment because it represents high-frequency semantic information, edges, and texture, while lacking predictable spatial patterns.

\noindent\textbf{Reported results (representation rigidity). } Figure \ref{NF_NTK_Representation_Rigidity} presents the results. In the first row, from left to right, we show the ground truth image and the reconstructed image at iterations 0, 500, and 1000. In the second row, we present, left to right: the Finite-NTK reconstructed image, the Finite-NTK \(\mathit{W}\) vector, the Infinite-NTK reconstructed image, and the Infinite-NTK \(\mathit{W}\) vector. Here, Finite-NTK refers to the Neural Tangent Kernel applied within finite-width neural networks, representing finite-time training. 
The PSNR values indicate the performance differences: NF model at 32.12, infinite NTK-based NF at 101.79, and finite NTK-based NF at 18.93. With all labels known, NTK excels in reconstruction due to its kernel correlation calculation, allowing nearly perfect reconstruction when all pixel relationships in the target image are known. 
Visually, there is a minimal difference, thus, \(\mathit{PSNR} \approx 30\), suggesting visual saturation. Additionally, the \(\mathit{W}\) vectors differ between Finite and Infinite NTK: the infinite-time NTK captures more details, as expected since Finite-time NTK illustrates training for only a limited time period (here, 1000 iterations) without full convergence.

\noindent\textbf{NF vs NTK Inpainting. } In this experiment, we aim to study the NTK's capabilities in a more complex task: inpainting, where some labels are masked, and the Kernel's pixel correlations can be partially disrupted. We apply the infinite (width and time) NTK reconstruction method to inpaint a masked image, comparing it to the standard NFs training approach, where we train the same NF model (Fig. \ref{NF_Architecture_with_FPE}) for 1000 steps with a learning rate of \(10^{-3}\). For NTK, we use the reconstruction method for MC task (Eqs. \ref{W_Vector_MC} and \ref{Y_Reconstruct_MC}). Here, the PE vector and the target image are masked in a circular 'hole' centered shape.
Define $r \in \mathbb{R}$ to be the centric hole masking radius, and set $$\mathcal{L}_2^r=\{i \mid i\in \{1,\cdots, N\}, \norm{(\mathit{a}_i, \mathit{b}_i)}_2^2 \geq \mathit{r}\}$$ to be the set of unmasked coordinate pairs. Then, the NF PE vector is $
    \big((\mathit{a}_i, \mathit{b}_i)\big)_{i\in \mathcal{L}_2^r}$ , and the target image $\big((\mathit{x}_i, \mathit{y}_i, \mathit{z}_i)\big)_{i\in \mathcal{L}_2^r}$ .

\noindent\textbf{Reported results (inpainting). } The results are shown in Fig. \ref{Inpainting_NTK_NF_Iter_1000}. The NF training results are presented from left to right: the GT image, the masked GT image, and reconstructed images at iterations 0, 100, 300, and 1000. The Infinite-NTK reconstruction result is the last image on the right.

The NF-trained model and NTK method yield relatively close results ($\text{PSNR} \approx 17.06$ vs. $\text{PSNR} \approx 20.21$) despite only training the NF model for 1000 iterations. In the masked pixel region, the scores are also comparable ($\text{PSNR} \approx 9.03$ for NF vs. $\text{PSNR} \approx 12.07$ for NTK). However, several distinctions emerge: first, the NF model predicts flowers in the masked region—reasonable given the lack of a fixed spatial pattern—while NTK does not, unable to capture object shape connections. Second, NTK's reconstruction lacks semantic coherence, with edge transitions from observed to masked areas being rough and inconsistent. The NTK method also fails to capture semantic details and texture of the flowers, unlike the NF model, which preserves semantic edges, details, and texture, yielding visually superior results. This experiment highlights a few critical points. Flower images often feature objects with unpredictable patterns and high-frequency signals, making pixel prediction challenging. Additionally, the hole-shaped mask introduces hidden neighboring pixels, where NTK’s kernel matrix struggles to approximate correlations, resulting in blurred reconstructions.

\begin{figure}
    \centering
        \includegraphics[width=0.48\textwidth]{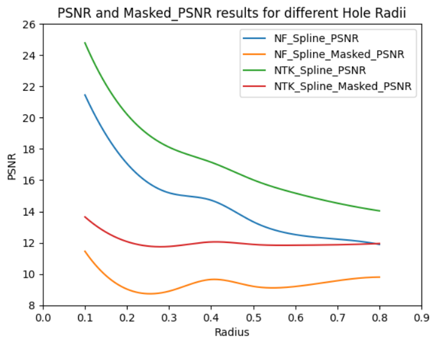}
        \caption{PSNR and masked PSNR versus hole radius. Green: NTK based NF PSNR. Blue: NF PSNR. Red: NTK based NF masked PSNR. Orange: NF masked PSNR. Masked PSNR is computed only over the masked region.}
    \label{Inpainting_NF_NTK_Radii_PSNR_Graph}
\end{figure}

\noindent\textbf{The effect of missing pixels. }
To analyze the impact of hole radius on the center hole inpainting task, we evaluate reconstruction PSNR as a function of the masked hole radius for both the NF model and the NTK method. For each method, we report PSNR on the full image as well as masked PSNR computed only over the hole region. The results are shown in Fig.~\ref{Inpainting_NF_NTK_Radii_PSNR_Graph}.

Overall, NF and NTK exhibit similar full image PSNR and masked PSNR trends across radii. However, NTK might be expected to better reconstruct small masked regions if local pixel correlations alone were sufficient. Instead, we observe a clear gap between NTK performance in representational rigidity and inpainting, indicating that the fixed kernel correlations that enable near perfect reconstruction when all labels are observed do not transfer reliably to partially observed settings.

These results suggest a key limitation of NTK in this context: its fixed kernel behaves primarily as a correlation structure rather than a feature-adaptive representation. As a result, NTK can reconstruct pixels in representation rigidity via kernel inversion, yet struggles in settings that require learning semantic structure and long range context, such as edge continuity, texture, and object level consistency. The disparity between representation rigidity and inpainting therefore highlights the limitations of the NTK regime for tasks that benefit from adaptive feature learning.

\begin{figure}[t]
    \centering
        \includegraphics[width=0.48\textwidth]{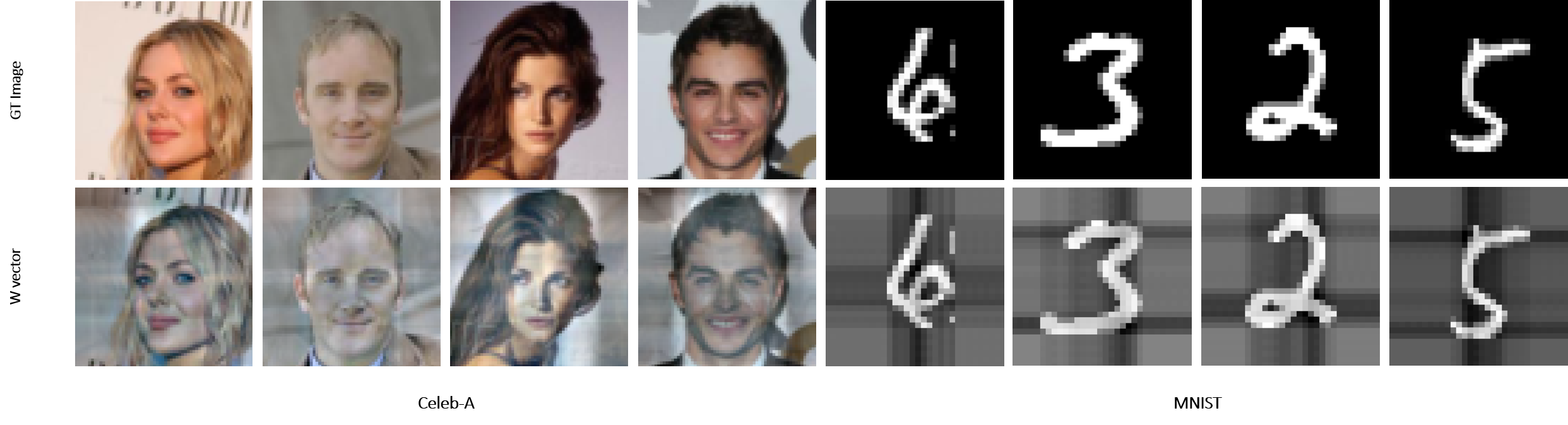}
        \caption{Ground truth image versus NTK coefficient representation $W$ on CelebA and MNIST. First row: ground truth images. Second row: corresponding $W$ representations.}
    \label{W_vs_GT_CelebA_MNIST}
\end{figure}

\noindent\textbf{Mathematical derived explanation. } To better understand the results, we delve into the core Kernel representation and present theoretical findings and explanations.
From equation~\ref{NN_Evolution_Kernel}, if to consider a specific cell \([(i, j)\, |\, 1 \leq i,j \leq C]\) in the NTK,
\begin{align*}
\mathcal{K}_{(\mathit{i}, \mathit{j})}(\mathit{x}, \mathit{x}^{'} ;\, \theta) &= \nabla{_\theta}\mathit{f}_{\mathit{i}}(\mathit{x}\, ;\, \theta)^T \ \cdot \ \nabla{_\theta}\mathit{f}_{\mathit{j}}(\mathit{x}^{'} ;\, \theta) = \\ &= \sum_{\mathit{q}=\mathit{1}}^{\mathit{Q}} \frac{\partial \mathit{f}_{\mathit{i}}(\mathit{x}\, ;\, \theta)^\mathit{T}}{\partial \theta_{\mathit{q}}} \cdot \frac{\partial \mathit{f}_{\mathit{j}}(\mathit{x}^{'} ;\, \theta)}{\partial \theta_{\mathit{q}}},
\end{align*}

where $\mathit{Q}$ is the total number of parameters in the NN $\mathit{f}$. See Fig. \ref{NTK_Matrix_Illustrarion} in the Appendix for illustration.

Calculating correlations through inner products of neural network gradients compresses meaningful information, neglecting dynamic architectural effects and resulting in a low-dimensional projection rather than a robust representation. While NTK offers advantages in training flexibility, it lacks effective feature extraction and true learning capability. To explore these limitations, we propose the concept of "NTK Representational Linearity", which addresses the fundamental constraints in NTK’s approach. This analysis contributes to understanding and mitigating limitations in the NTK theorem, highlighting its impact on representation learning.

\noindent\textbf{NTK representational linearity. } Recall the coefficient representation $\mathcal{W}$ in Eq.~\ref{W_Vector}, obtained by solving a kernel regression system on the observed coordinates. Denoting by $\mathcal{K}_{\mathrm{train}}$ the kernel matrix over the observed coordinates, the solution satisfies
$\mathcal{W} = \mathcal{K}_{\mathrm{train}}^{-1}\mathcal{Y}$,
hence $\mathcal{W}$ depends linearly on the labels once $\mathcal{K}_{\mathrm{train}}$ is fixed. Predictions at query coordinates are then reconstructed via kernel evaluations against the observed coordinates, i.e.,
$\tilde{\mathcal{Y}} = \mathcal{K}_{\mathrm{test}}\mathcal{W}$,
where $\mathcal{K}_{\mathrm{test}}$ denotes the cross kernel between query and observed coordinates. Therefore, $\mathcal{W}$ is best interpreted as a coefficient space solution to a fixed kernel regression problem, rather than as a task adaptive learned feature representation. Empirically, Fig.~\ref{W_vs_GT_CelebA_MNIST} shows that $\mathcal{W}$ is highly correlated with the target image in our setting, consistent with this coefficient space interpretation. In addition, the coefficient representation can preserve label noise and local artifacts, rather than encoding semantically organized features produced by nonlinear representation learning.

\noindent\textbf{In short. } NTK provides a powerful framework for analyzing neural network training dynamics and, in the infinite width regime, can approximate neural network evolution. However, in the kernel regression form used here, the resulting coefficient representation remains a linear, non adaptive mapping with respect to the labels once the kernel is fixed. This limits its ability to model richer task dependent feature transformations that are often important in neural field generation and structured reconstruction tasks. In representation rigidity and inpainting settings, this limitation is reflected in weaker semantic continuity and texture recovery. These observations motivate the development of NTK-KIP, MetaQuill, and MetaQuill-KIP.

\begin{figure}[t]
    \centering
    \includegraphics[width=\columnwidth]{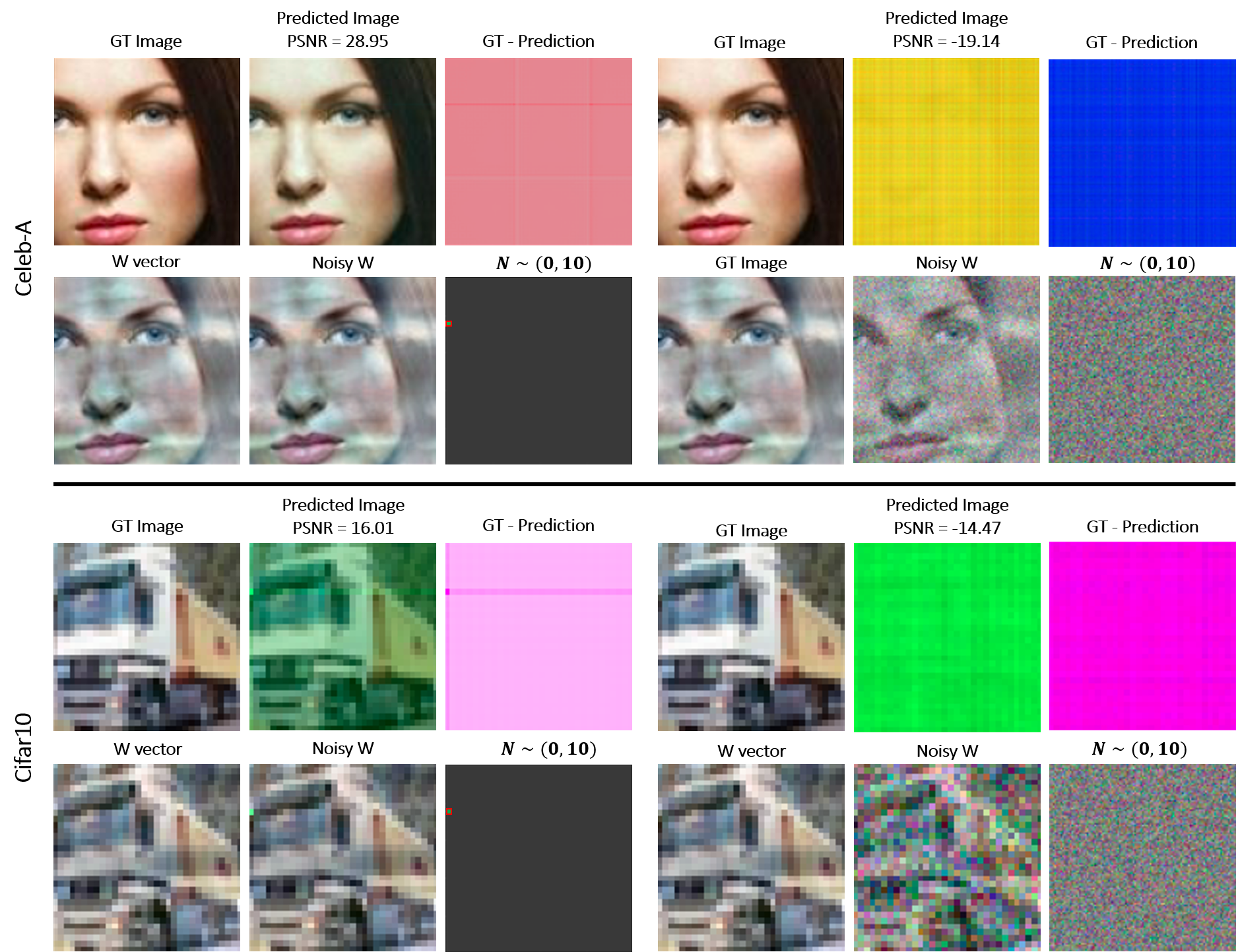}
    \caption{
    NTK robustness analysis on CelebA (top) and CIFAR-10 (bottom). 
    Left: perturbing a single coefficient in $W$. 
    Right: perturbing all coefficients in $W$. 
    Within each block, we show: ground truth image (top-left), original $W$ (bottom-left), Gaussian noise ($\mathcal{N}(0,10)$), perturbed $W$, reconstruction from perturbed $W$ (top-middle), and reconstruction error heatmap (top-right). 
    Small perturbations in $W$ cause large, structured artifacts in pixel space, highlighting noise sensitivity of the explicit NTK coefficient representation.
    }
    \label{fig:ntk_robustness}
\end{figure}

\subsection{Exploring NTK's Robustness to Noise} 
\label{NTK_Robustness}
While classic NNs can learn and adapt to noisy or imperfect data, NTK struggles with such robustness, especially in very deep networks and complex, large-scale data. This limitation stems from NTK's core theory. For example, in Matrix Completion (MC) scenarios, added noise in the samples propagates through the kernel, significantly affecting reconstruction due to NTK’s linear matrix transformations and pixel-space correlations. Consequently, NTK's performance suffers in real-world tasks with noise, skewed samples, or varying data distributions.

\noindent{\textbf{Image reconstruction under $W$ perturbations}.} To demonstrate this claim, we examine image reconstruction under perturbations to the $W$ vector. In this experiment, we compute the $W$ vector for a set of GT images and then add to it stochastic noise from a normal distribution ($\varphi \sim N(\mu, \sigma)$).
We conduct two experiments: first, perturb the $W$ vector at a single pixel, and second, across the whole vector. Then use NTK to reconstruct the fully observed image. Results for different perturbation levels are in Fig.~\ref{fig:ntk_robustness}, where we can notice:
\begin{enumerate}
    \item A single noisy pixel in $W$ affects not just the target pixel, but also impacts rows and columns at fixed frequency in the reconstructed image.
\item When adding noise to the entire vector, the image structure remains somewhat visible, but the reconstruction is highly perturbed, with minimal preservation of the GT image's structure.
\end{enumerate}

This leads to a significant vulnerability in the NTK method, specifically in such tasks and applications that rely on a high $SNR$ yielded results.

\noindent\textbf{In short.} While classic NNs handle noise, NTK faces challenges. To test NTK’s robustness, we perturb its $\mathcal{W}$ vector with various noise distributions. 
The reconstruction of labels via NTK's $\mathcal{W}$ vector is as follows, 
$\hat{\mathcal{Y}} = \mathcal{K}_{\mathit{T}}(\mathit{x}, \mathit{x}^{'} ; \theta) \times \mathcal{W}\ |\ \mathcal{K}: \mathbb{R}^d \ \times\  \mathbb{R}^d \ \rightarrow{} \ \mathbb{R}^{\mathit{C} \times \mathit{C}}$. We found that NTK's representation is extremely noise-sensitive, leading to a blurry reconstruction containing Kernel-resulting frequencies.

\begin{figure*}[ht]
    \centering
        \includegraphics[width=1.0\textwidth]{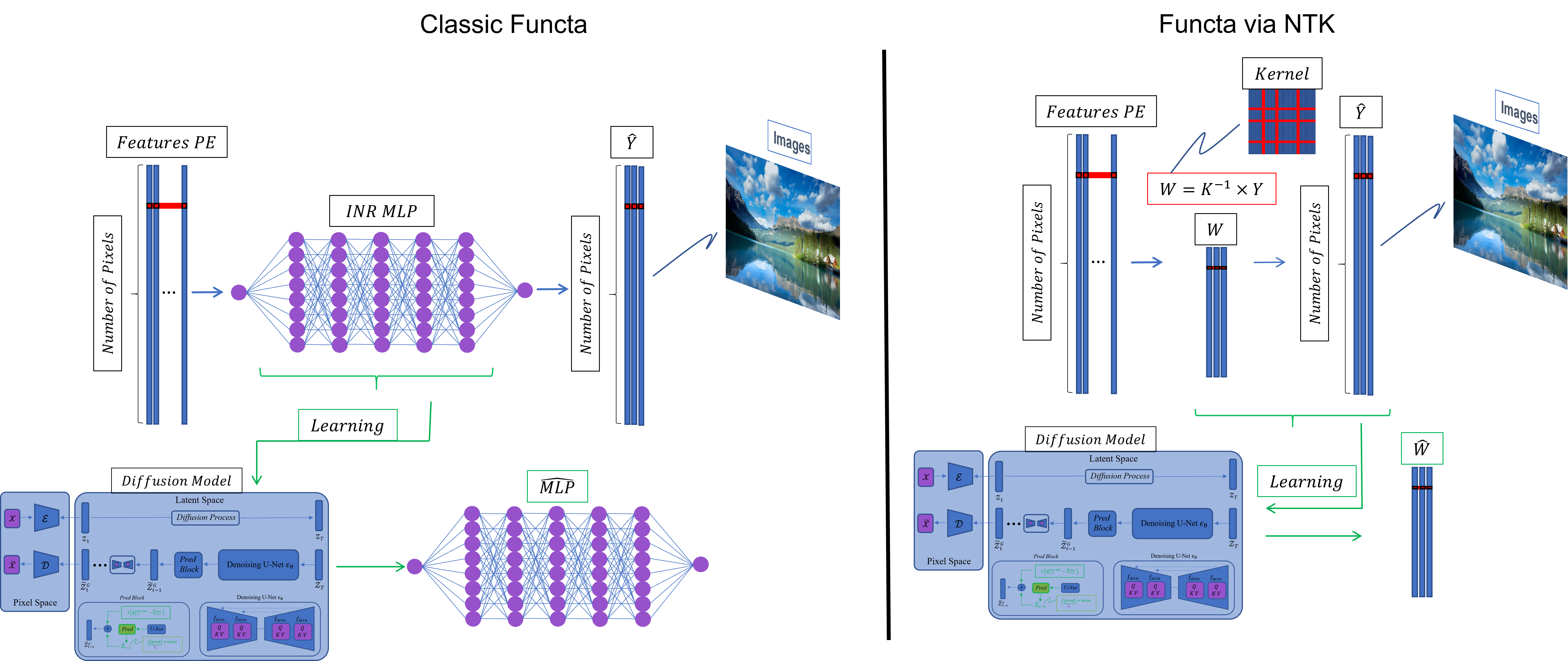}
        \caption{Left: Classic Functa. Right: Functa via NTK. In Functa via NTK, each neural field in the classic Functa pipeline is replaced by its NTK coefficient representation $W$.}
    \label{Functa_NTK_Illustration}
\end{figure*}
\begin{figure}[h!]
    \centering
        \includegraphics[width=0.48\textwidth]{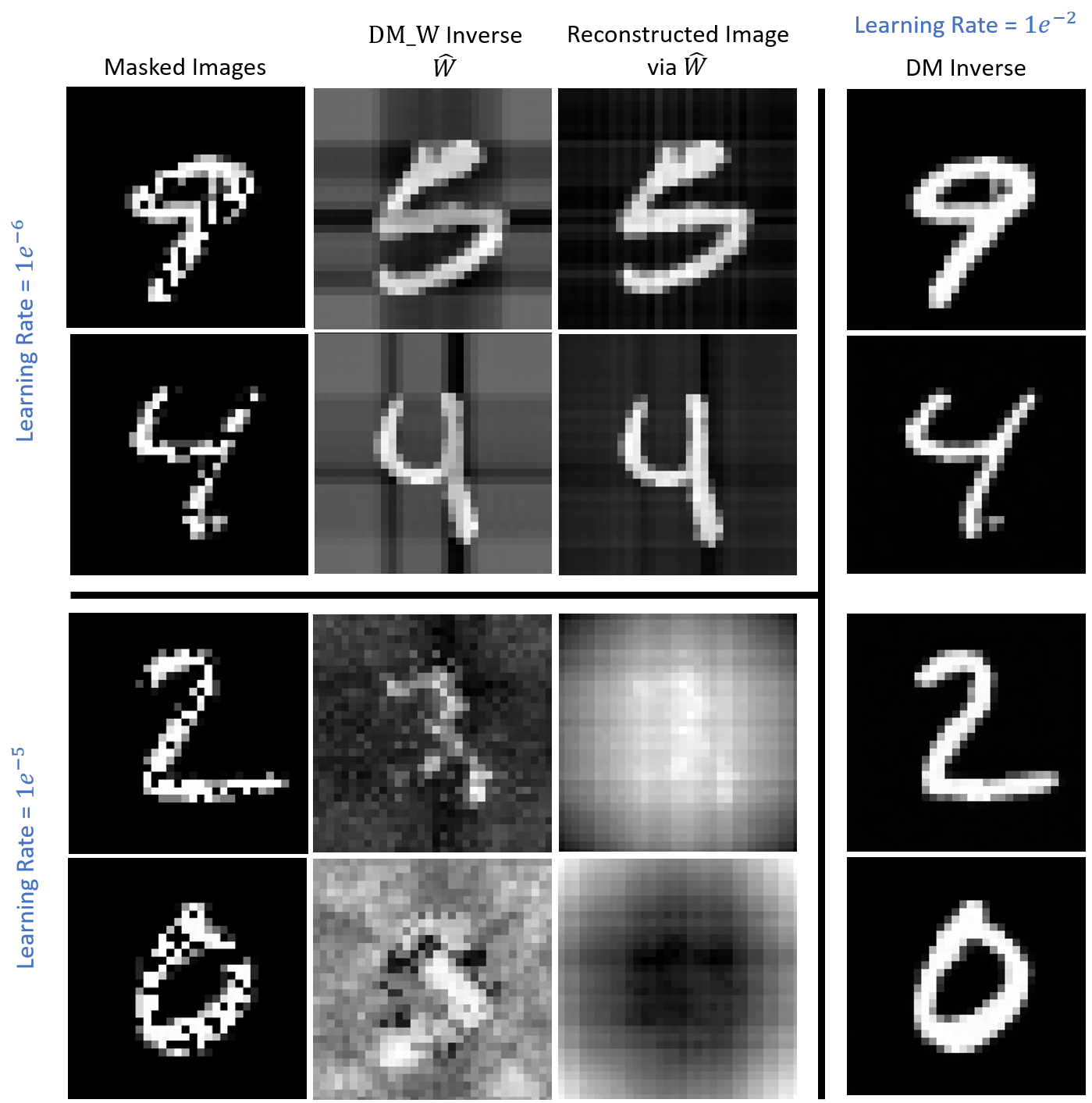}
        \caption{Guided diffusion reconstruction results for DM and DM\_W. Columns from left to right: masked image, DM\_W inverse estimate $\hat{W}$, reconstructed image via $\hat{W}$, and DM inverse. Learning rates: left two rows use $10^{-6}$ (top) and $10^{-5}$ (bottom), right column uses $10^{-2}$.}
    \label{Guided_DM_&_WDM}
\end{figure}

\subsection{A step forward: NTK for Functa} \label{NTK_Functa}
\noindent\textbf{Functa. }Deep learning often represents data on discrete grids (e.g., pixels), though signals are continuous. NFs offer an alternative, predicting values at any spatial location, $NF: l \rightarrow v$. Functa \cite{dupont2022data} explored using NFs as data, showing that datasets can be distilled into compact NF representations, reducing the size by $\sim 0.5\%$ while preserving key features for accurate reconstruction, highlighting the importance of feature learning, which we benchmarked to evaluate NTK's representation efficiency (see Fig.~\ref{Functa_NTK_Illustration} for an illustration). We now experiment with NTK for representation learning via the Functa setting. Our downstream chosen task is MC, and more specifically, Inpainting.

\noindent\textbf{The setting. }In the NTK Functa method, unlike classic Functa, where a Diffusion Model (DM) is being trained on a compact form of the NFs weights, we train the DM on infinite-width NFs data. And the way to make this feasible is via NTK kernel trick; see Fig.~\ref{Functa_NTK_Illustration}. We replace each NF with a $W$ vector (via Eq.~\ref{W_Vector}), and train the DM upon. The calculation process is convenient due to the NTK Kernel being shared upon all $W$s (Same NF model and fixed image size, thus fixed PEs). At inference, We'll generate a novel $\mathcal{\hat{W}}$ vector via our trained DM, and predict the missing pixels via the pre-calculated NTK Kernel (Eq. \ref{Y_Reconstruction}).
To use the DM for specific tasks such as inpainting, we use a guidance algorithm where we pass the observed labels as a guided input for the DM. Formally,
$$x_{t-1} = \frac{1}{\sqrt{{\alpha_t}}}(x_t\, -\, \frac{1-\alpha_t}{\sqrt{1 - \overline{\alpha_t}}} \cdot \hat{\epsilon}_{t-1})\, +\, S \cdot \nabla P(y|x)\, +\, \sqrt{\overline{\beta}_t} \cdot z$$

The experiment was conducted on the MNIST dataset. In which, we trained two DMs. First, directly on the image domain, and second, on the $W$s dataset. Example results of the calculated $W$s dataset, can be seen in Fig, \ref{W_vs_GT_CelebA_MNIST}. For the full formulation and notation, see Appendix \ref{functa_NTK_appendix}.
We evaluate NTK's representational performance by comparing the W vector representation with the original image domain in an MC inpainting task, using reconstructed images from Guided-DMs trained on the Ws dataset and directly on GT images. Results can be seen in Fig \ref{Guided_DM_&_WDM}.

\noindent\textbf{Representational robustness.} 
Results in the GT image space are smoother and more refined (Right-most column), whereas training in the $W$ vector space exhibits noise sensitivity (Bottom-left section - $2$ and $0$ digits) and inconsistent guidance (Top-left and bottom-left results gap with respect to the guidance learning rate) while showing poor results (See Section \ref{NTK_Robustness}), resulting in blurry reconstructions.
The denoising process in the diffusion trajectory is complex, and added noise in learned representations further complicates the inversion. This suggests that the NTK representation does not add valuable features and may even degrade results. Even successful reconstructions with NTK exhibited noise issues.

\noindent\textbf{Run-Time efficiency.} Considering inference runtime efficiency, we compared the representational size and computation of both methods (Functa vs. NTK) using the basic NF in both algorithms. In Functa, the base NF has $5 \times 256$ hidden layers and an RGB output ($1\times 3$), allowing us to estimate the runtime complexity (ignoring bias). The results:
With NF,
\begin{align*}
    \underbrace{\overbrace{(h \cdot w) \times d}^{PE} \cdot l}_{1^{st}\, hidden\, layer}\, +\, \underbrace{(l \times l \times N_l)}_{mid\, hidden\, layers}\, &+\, \underbrace{(l \times c)}_{final\, output\, layer}   \\&
    \cong 8 \times 10^5 = \textbf{0.8M}
\end{align*}

In NTK (Neglecting FPE and kernel mapping calculation),
\begin{align*}
    (*)=\underbrace{\overbrace{(h \cdot w) \times d}^{PE} \times (d\, +\, 2 \cdot d \cdot \overbrace{k}^{Fourier\, bands})}_{FPE}\, &+\, \underbrace{(K_{train}^{-1} \times Y)}_{W_{vector}}\, \\&+\,  \underbrace{K_{test} \times W}_{Reconstruct\, \hat{Y}} 
\end{align*}

where, $(*) \geq \textbf{2G}$ (For general matrix multiplication), and $(*) \geq \textbf{6M}$ (For matrix-Vector multiplication). For the full calculation, see Appendix~\ref{functa_NTK_appendix}.

\noindent\textbf{In summary:} NTK is unsuitable for practical representation tasks due to weaknesses in structure, representation, and runtime, posing limitations for applications like model scaling and edge AI (energy-efficient systems). We observe three main weaknesses in the NTK method:
\begin{enumerate}
    \item \textbf{Representational structure efficiency:} The NTK matrix $W$ has a representational size of $\mathbb{R}^{(h \cdot w) \times c}$, matching the ground truth (GT) image size but lacking efficient structure.
    \item \textbf{Representational effectiveness:} Although $W$ matches the image size, it fails to achieve effective representation in terms of performance results.
    \item \textbf{Time-complexity:} The NTK method exhibits high time complexity and computational cost, being at least one order of magnitude less efficient in runtime.
\end{enumerate}

\section{Method}
The limitations outlined in the previous section indicate that the straightforward application of NTK as a substitute for training neural fields is inadequate for effective representation learning. To address these issues, we propose two novel approaches that can be used independently or combined to create efficient NTK representation learning algorithms. Each approach targets a unique, previously unresolved challenge within NTK, and their integration could overcome existing limitations.

\begin{figure}[!t]
    \centering
        \includegraphics[width=0.48\textwidth]{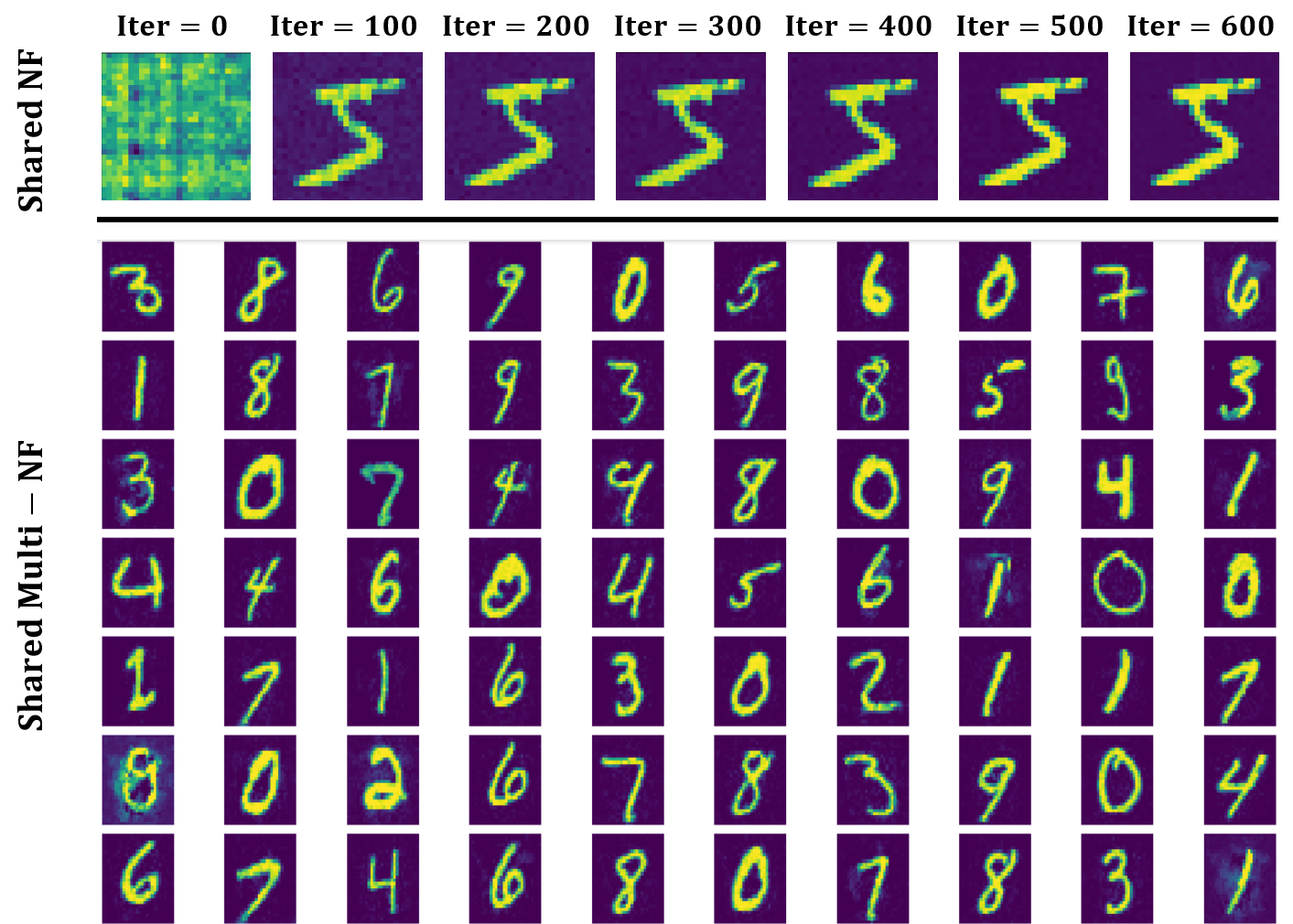}
        \caption{Top section: Training a Shared NF from a Single Image. Iterations from left to right: 0, 100, 200, 300, 400, 500, 600. Bottom section: Last iteration in training a Multi-Shared NF (From 100 images)}
    \label{Training_Shared_NF}
\end{figure}
\begin{figure}[!t]
    \centering
        \includegraphics[width=0.44\textwidth]{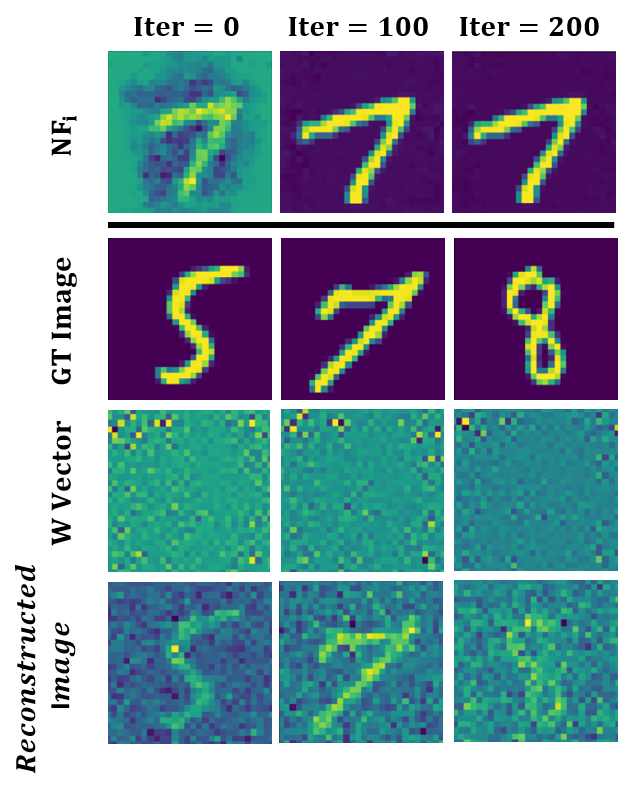}
        \caption{Top section: Training a $NF_i$ from $NF_{\theta_S}$ (Shared NF) for 200 iterations, the iterations from left to right: 0, 100, 200. Bottom section: Reconstructed Images via 
$W$ Vector and $NF_{\theta_S}$. Bottom section, from top to bottom: GT image, W vector, Reconstructed image}
    \label{Train_NFi_Reconstruct_via_w}
\end{figure}

\subsection{Neural Field Kernel Inducing Points (NTK-KIP)} \label{Neural Field Kernel Inducing Points (NTK-KIP)}
As opposed to using the NTK as a direct mapping between data $Y$ (e.g. images) and representation $W$ (creating a linear transformation representation. See Section \ref{NTK_Natural_Limitations}), we propose to use the NTK as a mean to accelerate a different method of extracting useful representation, namely, finding inducing points in images. This will yield distilled data, and also a non-linear NTK representation ($W$ vector).

\noindent{\textbf{NTK-KIP.}} While the typical way to train a NF is to use all the image pixels as the training set, here we seek to find a smaller set of induced pixels that can serve as a compressed training set for the neural field. In other words, we are optimizing the values of a small training set, such that training a NF on it will result in a good generalization of the real pixels in the image. Notably, each update of the inducing point optimization requires full training of the NF, which renders this unrealistic. To solve this, we propose to replace the NF training process with the NTK regression approximation and optimize the inducing points by differentiating through the NTK regression. Formally, let $N$ be the total number of images in our dataset. Given image $i \in [1,\, ...,\, N]$ of size $(h,\, w,\, c)$, let $P = h \cdot w$ be the total number of pixels in $i$, $p \in [1,\, ...,\, P]$ be a pixel in $i$, $x \in \mathbb{R}^{N \times 2}$ be the PE vector, $y \in \mathbb{R}^{(P \times c)}$ be the image labels (values). We then define $x_i^p \in \mathbb{R}^2$ to be the PE coordinate which corresponds to pixel $p$ in image $i$, and $y_i^p$ to be the value (label) of pixel $p$ in image $i$. In the Raw PE (RPE) example, we optimize $x_i^p$ and $y_i^p$ to minimize the reconstruction error for the full image:
\begin{align*}
    \arg \min_{x_i^p, y_i^p}
    \left\| \mathcal{K}(x, x_i^p) \; \mathcal{K}(x_i^p, x_i^p)^{-1} \; y_i^p   - y \right\|
\end{align*}

This representation is therefore no longer a simple linear mapping of the signal, and can potentially lead to extracting more meaningful and useful information from the image. Previous work has shown that this approach is effective for dataset distillation in classification problems~\cite{nguyen2020dataset,loo2022efficient}. Here we apply the same method on NF training, where our dataset consists of image coordinates and pixels, and therefore `dataset distillation` translates to finding inducing points for an image.
For the full algorithm elaboration and formulation, please refer to Appendix \ref{NF_KIP_Algorithm_Appendix}.

\subsection{MetaQuill Algorithm} \label{MetaQuill_Algorithm}
One of the limitations of replacing NF training with NTK separately for each image is that no global information at the level of the dataset is extracted. In comparison, in the standard implementation of Functa \cite{dupont2022data} with finite NFs, a shared set of weights for the whole dataset is used, and the representation of each image consists of a latent vector that is used to modulate the shared network. The advantage of this is that global information that is common to the whole dataset does not need to be stored in the representation of each image.

\noindent{\textbf{MetaQuill.}} We propose $\textit{MetaQuill}$ Algorithm. An algorithm for Neural Field meta-learning through NTK, which enhances the NTK method with feature-learning capabilities in an efficient manner.
We propose to use a similar approach to Functa, using Model Agnostic meta-learning (MAML) ~\cite{finn2017model} and finite width NTK ~\cite{novak2022fast}.  Our method is similar to MAML, where the initial values of the network weights are optimized through the (short) training of the network for different images. In order to accelerate the inner loop of MAML, we use finite-width NTK, a method that approximates the training process of a finite network, by linearizing the weights updates around some initial value of the weights. Formally, we approximate a few steps $t$ of gradient descent on the parameters $\theta$ around the parameters $\theta_\tau$  by:
\begin{align*}
    f_{\theta_{\tau+t}}(x) - f_{\theta_{\tau}}(x) &\approx (\theta_{\tau+t} - \theta_{\tau})^T  \nabla_{\theta_{\tau}}f_{\theta_{\tau}}(x) \\& =  \Delta \theta ^T \nabla_{\theta_{\tau}}f_{\theta_{\tau}}(x)
\end{align*}

In our context, $\theta_\tau$ is shared for all images and serves as a starting point for the above approximation and is trained simultaneously with separate values of $\Delta \theta$ corresponding to each image in the dataset.
From the shared NF, we can calculate each specific NF model which corresponds to an image $i \in N$ by,
\begin{align*}
    NF_i(x) = \triangle \theta_i^T \cdot \nabla_{\theta_S} f_{\theta_S}(x)
\end{align*}

The values of $\Delta \theta$ can then serve as the representation of the specific image, or mapped to the kernel coefficient $W$. Formally,
\begin{align*}
    \triangle \theta_i = \sum_{W_i^p,\, x^p} W_i^p \cdot \nabla_{\theta_S} f_{\theta_S}(x^p),
\end{align*}
such that $ W_i = K_{\theta_S}^{-1} \cdot Y_i$ , where $Y_i$ is GT image $i$ and $p$ is an image pixel.
For the full algorithm development and the theoretical analysis, please refer to Appendix \ref{MetaQuill_Algorithm_Appendix}. In Section \ref{Results}, Fig.~\ref{Training_Shared_NF} shows example of our method applied to a dataset of 100 images from MNIST, achieving shared NF mapping $\theta_{\tau}$. And Fig. \ref{Train_NFi_Reconstruct_via_w} shows results of both achieving $\triangle \theta$ and extracting the $W$ vector (via finite-NTK) from the shared NF model.

\subsection{MetaQuill-KIP Hybrid Adaptation}
\label{MetaQuillKIP}
NTK-KIP and MetaQuill address two different failure modes of applying NTK regression to neural fields. NTK-KIP learns a compact, task-specific support set of synthetic coordinates and labels that makes NTK fitting nonlinear, efficient, and robust to sparse observations. MetaQuill learns a reusable shared initialization $\theta_S$ across tasks and adapts new images by optimizing only a lightweight residual $\Delta \theta$, rather than re-training a full network. However, each method alone still retains a weakness: NTK-KIP does not transfer shared structure across tasks, and MetaQuill remains locally linear around $\theta_S$ and can struggle to hallucinate large missing regions.
We therefore propose a hybrid procedure, \emph{MetaQuill-KIP}, which fuses these strengths.
\paragraph{Setup.}
Let $f_\theta : \mathbb{R}^d \to \mathbb{R}^c$ be the neural field (for example, an MLP mapping positional encodings to RGB). MetaQuill meta-learns a shared parameter vector $\theta_S$ across a pool of training tasks. For a new target task $\mathcal{T}$ (a single image), we seek a small per-task update $\Delta \theta_{\mathcal{T}}$ such that
\begin{equation}
    f_{\theta_S + \Delta \theta_{\mathcal{T}}}(x) \approx y(x)
\end{equation}
for all spatial coordinates $x$ in the image, including coordinates that are never observed during inpainting.
\paragraph{Step 1: Shared backbone from MetaQuill.}
MetaQuill provides $\theta_S$ by minimizing, over a set of training tasks $\{\mathcal{T}_j\}$,
\begin{equation}
    \min_{\theta_S} \sum_j \mathcal{L}\Big( f_{\theta_S + \Delta \theta_j}(x),\, y_j(x)\Big).
\end{equation}
Here $\Delta \theta_j$ is the task-specific residual obtained from a short adaptation starting at $\theta_S$, approximated via a finite-width NTK linearization so that no expensive full inner loop retraining is required.
In practice, MetaQuill approximates the inner adaptation with a finite-width NTK linearization around $\theta_S$, so that each task learns only its residual $\Delta \theta_j$ without re-training the full network. This makes $\theta_S$ a meta-learned feature extractor that already encodes global structure (for example, digit strokes in MNIST or petal texture in Flowers).
\paragraph{Step 2: KIP-style distilled initialization.}
For a new task $\mathcal{T}$ with observed pixels $\{(x_i, y_i)\}_{i \in \mathcal{O}}$ and possibly large missing regions, we first run a Kernel Inducing Point (KIP) procedure to produce a compact distilled support set
\begin{equation}
    \mathcal{S}_\mathcal{T} = \{(\tilde{x}_m, \tilde{y}_m)\}_{m=1}^{M},
\end{equation}
where $M \ll |\mathcal{O}|$. These $\tilde{x}_m$ are synthetic coordinates (learned positional encodings), and $\tilde{y}_m$ are synthetic labels, optimized so that solving an NTK-style regression on $\mathcal{S}_\mathcal{T}$ alone already approximates the full image.
Concretely, we optimize $\mathcal{S}_\mathcal{T}$ to minimize
\begin{equation}
    \mathcal{L}_\mathrm{KIP} =
    \left\| \mathcal{K}_{\theta_S}(X, \tilde{X}) 
    \Big(\mathcal{K}_{\theta_S}(\tilde{X}, \tilde{X}) + \lambda I\Big)^{-1} \tilde{Y}
    - Y \right\|^2,
\end{equation}
where $X$ and $Y$ are the coordinates and pixel values of either all known pixels or just the observed (unmasked) pixels, $\tilde{X}$ and $\tilde{Y}$ stack the distilled supports $(\tilde{x}_m, \tilde{y}_m)$, and $\mathcal{K}_{\theta_S}$ is the finite-width NTK computed around $\theta_S$. The regularizer $\lambda I$ is the usual ridge term.
This step gives us two things:
(i) an NTK predictor conditioned on a tiny distilled set, and
(ii) an implicit task-specific direction in parameter space.
\paragraph{Step 3: Lifting the KIP solution into parameter space.}
Because $\mathcal{K}_{\theta_S}$ is the tangent kernel of $f_\theta$ at $\theta_S$, its solution corresponds to a first-order update of the weights. In other words, there exists a residual $\Delta \theta_{\mathrm{KIP}}$ such that, to first order,
\begin{equation}
    f_{\theta_S + \Delta \theta_{\mathrm{KIP}}}(x)
    \approx
    \mathcal{K}_{\theta_S}(x, \tilde{X})
    \Big( \mathcal{K}_{\theta_S}(\tilde{X}, \tilde{X}) + \lambda I \Big)^{-1}
    \tilde{Y}.
\end{equation}
We take this $\Delta \theta_{\mathrm{KIP}}$ as the \emph{initialization} for our per-task residual, instead of starting from $\Delta \theta = 0$ as in vanilla MetaQuill.
\paragraph{Step 4: Fast nonlinear residual refinement.}
Finally, we refine only the residual $\Delta \theta$ by standard gradient descent for a short budget of steps $T$:
\begin{equation}
    \Delta \theta^{(t+1)} 
    = \Delta \theta^{(t)} 
    - \eta \, \nabla_{\Delta \theta}
    \mathcal{L}\Big(f_{\theta_S + \Delta \theta^{(t)}}(x_i),\, y_i\Big)_{i \in \mathcal{O}}.
\end{equation}
Here $t = 0,1,\ldots,T-1$, $\eta$ is a small learning rate, and $\mathcal{O}$ indexes only the observed (unmasked) pixels.
with $\Delta \theta^{(0)} = \Delta \theta_{\mathrm{KIP}}$ and $\theta_S$ frozen.
Because $\Delta \theta$ is typically orders of magnitude smaller than $\theta_S$, this refinement is extremely fast in wall-clock time (on the order of a few seconds for $T \in [50,500]$ in our experiments). Importantly, this last stage is \emph{nonlinear}: we are no longer constrained to the pure tangent regime of MetaQuill's linear approximation. This lets the model hallucinate plausible structure in large missing regions.
\paragraph{Why this hybrid matters.}
MetaQuill-KIP combines:
(i) a meta-learned backbone $\theta_S$ that captures globally reusable structure across tasks,
(ii) a distilled, task-specific inducing set from KIP that encodes the identity of the new instance using only a handful of synthetic coordinates and values, and
(iii) a fast, nonlinear residual refinement that updates only $\Delta \theta$ instead of re-training the full network.
Empirically, on MNIST, MetaQuill-KIP reaches over $32$\,dB PSNR after only $50$ residual-update steps, compared to about $11$\,dB for kernel ridge regression alone and under $19$\,dB for a tangent-only MetaQuill-style adaptation with no KIP initialization (See Table~\ref{tab:mnist_combined_runtime}). On a Flowers inpainting task with large missing regions, MetaQuill-KIP reconstructs semantically consistent petal and texture structure while adapting in seconds per image, despite observing as little as $1\%$ to $10\%$ of the pixels. Competing diffusion models in our study either require heavy pretrained generative priors (Stable Diffusion, StrDiffusion) or significantly longer per-image adaptation (for example, full diffusion or many-step score refinement), and they cannot reach comparable reconstruction quality under the same lightweight adaptation budget.

\begin{figure*}[ht!]
    \centering
    \includegraphics[width=0.98\textwidth]{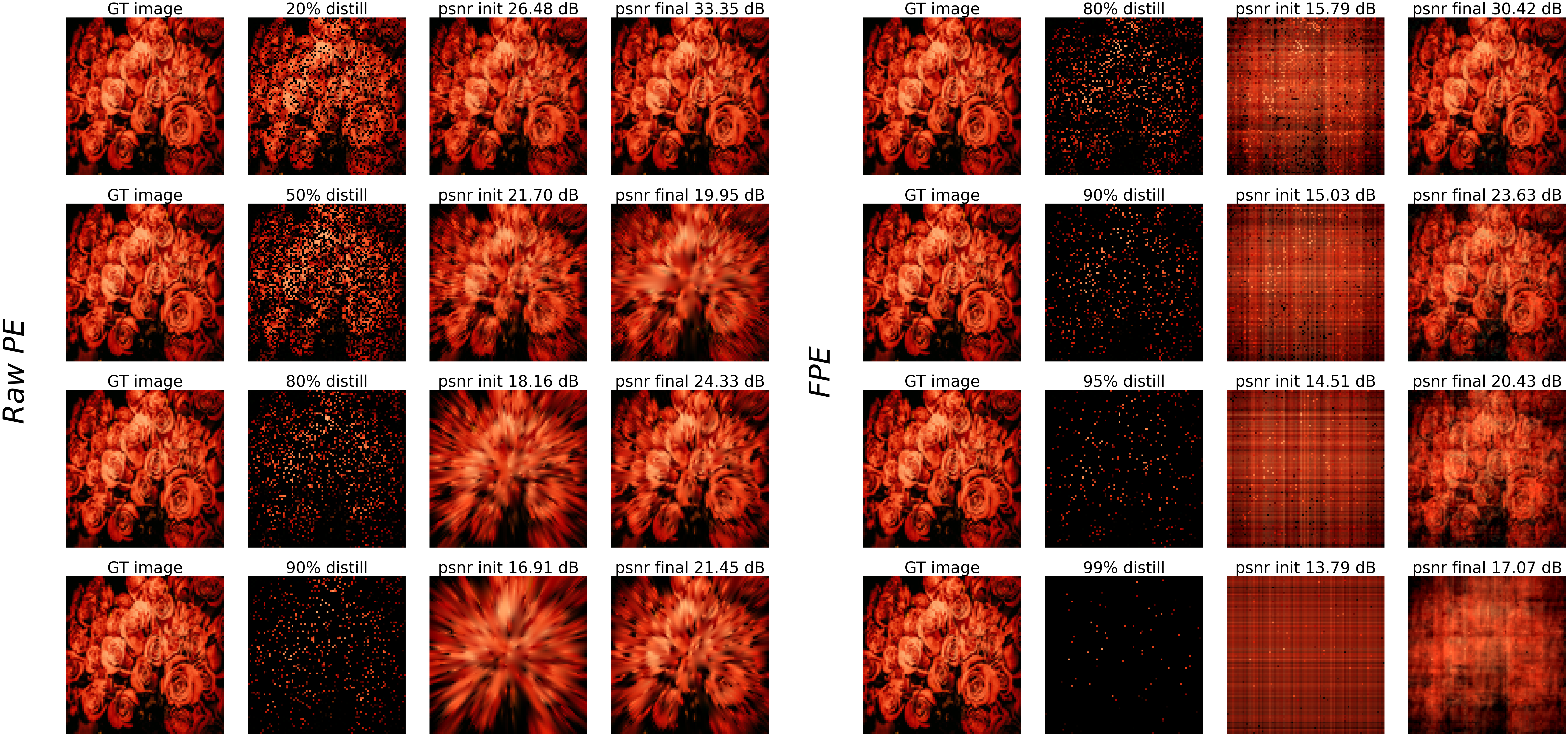}
        \caption{NTK-KIP experiment. Left section: Raw PEs. Right section: FPEs. In each section, columns from left to right: ground truth image, distilled image, initial PSNR, final PSNR. Left section distillation percentages (top to bottom: 20\%, 50\%, 80\%, 90\%). Right section distillation percentages (top to bottom: 80\%, 90\%, 95\%, 99\%).}
    \label{NF_KIP_PEs_experiment}
\end{figure*}

\section{Results} \label{Results}
We present the results of our three algorithms: \textit{NTK-KIP}, \textit{MetaQuill}, and \textit{MetaQuillKIP}.

\subsection{NTK-KIP Results}
\begin{figure}[t!]
    \centering
    \includegraphics[width=0.48\textwidth]{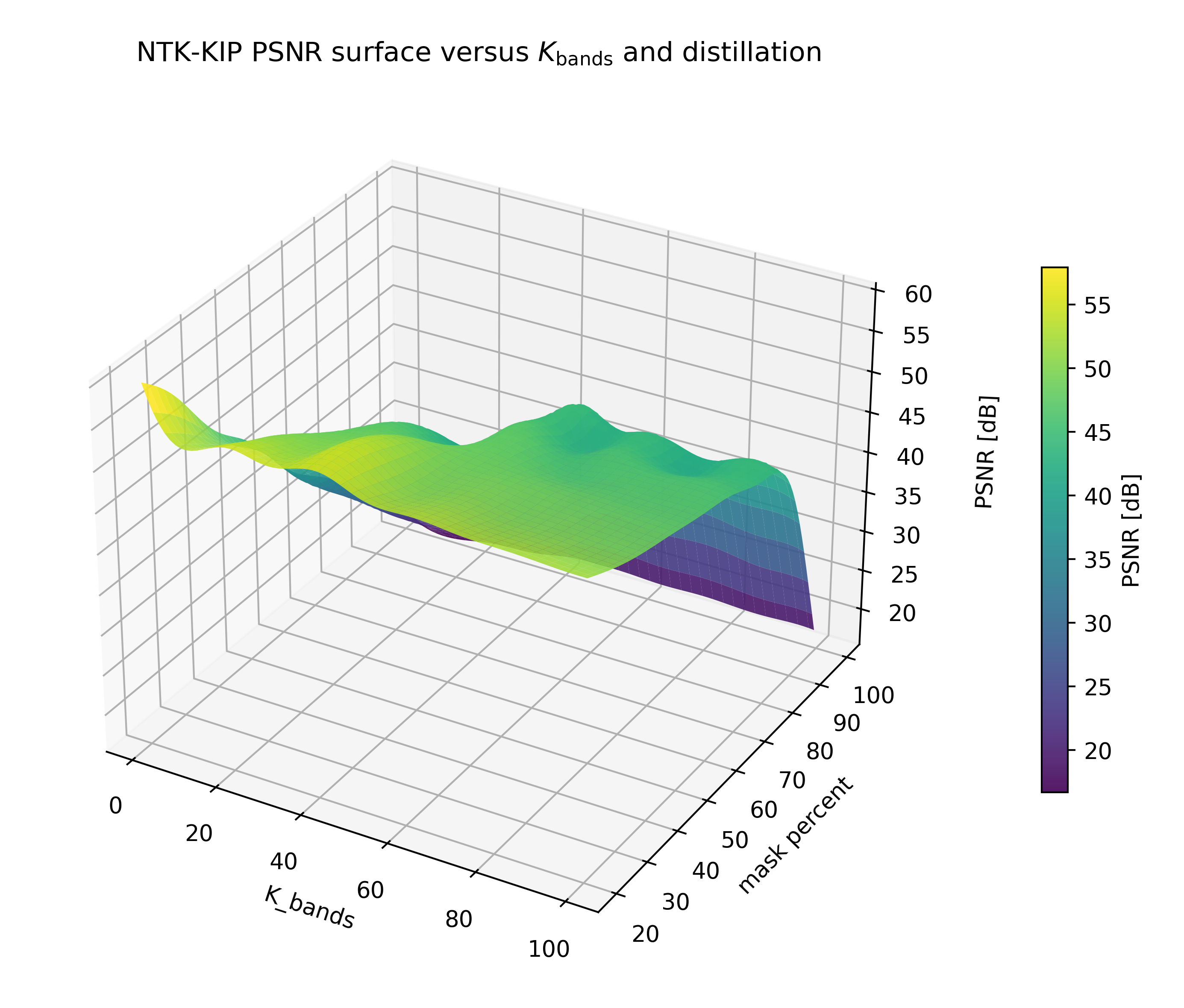}
    \caption{
    NTK-KIP PSNR surface as a function of Fourier $K_{\text{bands}}$ (horizontal axis)
    and observed fraction / distillation percentage (depth axis). 
    The vertical axis is PSNR [dB]. 
    This surface shows a clear ridge: performance improves sharply as we (i) raise $K_{\text{bands}}$ to allow richer Fourier positional encoding
    and (ii) allow a modest increase in observed fraction, then saturates.
    The high-PSNR band corresponds to the sweet spot where NTK-KIP has both expressive positional features and enough distilled samples to remain numerically stable.
    }
    \label{fig:kip_surface_fourier}
\end{figure}

\begin{figure}
    \centering
        \includegraphics[width=0.48\textwidth]{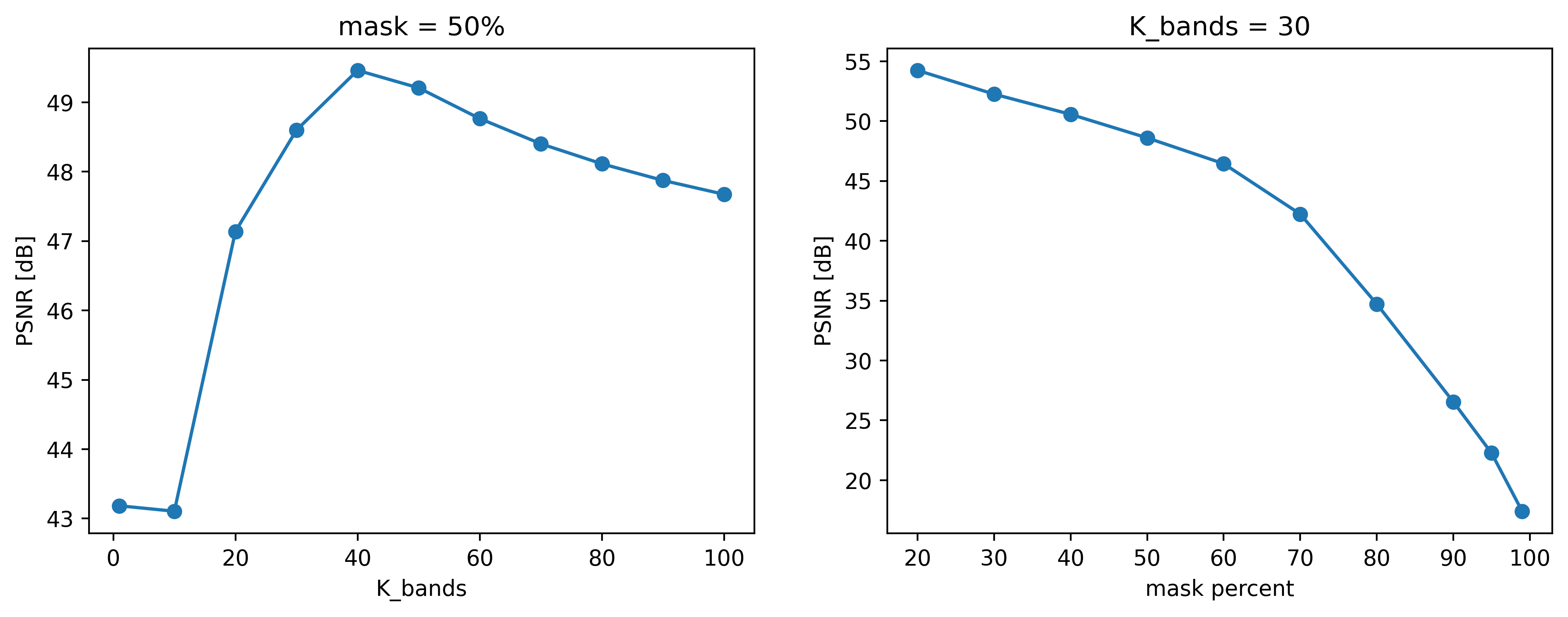}
        \caption{Left: NTK-KIP PSNR [dB] versus Fourier $K_{\mathrm{bands}}$ at distillation $50\%$, with best PSNR $\approx 48.40$ at $K_{\mathrm{bands}}=30$. Right: NTK-KIP PSNR [dB] versus distillation percentage at $K_{\mathrm{bands}}=30$, with best PSNR $\approx 54.24$ at distillation $20\%$.}
    \label{NF_KIP_2D_Graph_projections}
\end{figure}

\noindent{\textbf{The setting.}} We train an NF in such a manner where we seek to find a small, distilled set of induced pixel's coordinates that can serve as a compressed training set for the NF. In other words, we are optimizing the values of a small training set, such that training an NF on it will result in a good generalization of the real pixels in the image.
We show two experiments: in the first, the pixel position is optimized, and in the second, the Fourier positional feature is optimized. 
For Fig.~\ref{NF_KIP_PEs_experiment} and Table~\ref{PE_(Raw and FPE)_vs_Distillation} we report a qualitative slice at $K_{\mathrm{bands}}=20$.
For the hyperparameter study in Fig.~\ref{fig:kip_surface_fourier} and Fig.~\ref{NF_KIP_2D_Graph_projections}, we sweep $K_{\mathrm{bands}}$ and distillation percentage while keeping the remaining settings fixed.

\noindent{\textbf{Distilled representation.}} Fig.~\ref{NF_KIP_PEs_experiment} shows the reconstruction results of our method using different number of inducing points. 
Fig.~\ref{NF_KIP_PEs_experiment} shows representative reconstructions across distillation levels for Raw PE and FPE at $K_{\mathrm{bands}}=20$.
To summarize the quantitative trend across distillation percentages and encoding types, see Table~\ref{PE_(Raw and FPE)_vs_Distillation}. At $K_{\mathrm{bands}}=20$, Raw PE reaches about $33.35$ dB at $20\%$ distillation and about $19.95$ dB at $50\%$, while FPE maintains about $20.43$ dB at $95\%$ and about $17.07$ dB at $99\%$.

Optimizing the Fourier positional encoding improves reconstruction quality at high distillation rates and increases representational capacity through $K_{\mathrm{bands}}$.
Fig.~\ref{NF_KIP_PEs_experiment} shows qualitative reconstructions at a fixed $K_{\mathrm{bands}}$, while Fig.~\ref{fig:kip_surface_fourier} and Fig.~\ref{NF_KIP_2D_Graph_projections} summarize the corresponding hyperparameter sweep over $K_{\mathrm{bands}}$ and distillation.

\noindent{\textbf{3D effect.}} To study sensitivity to hyperparameters, we sweep Fourier $K_{\mathrm{bands}}$ and distillation percentage and record the final PSNR.
Fig.~\ref{fig:kip_surface_fourier} visualizes a continuous PSNR surface estimated from the sweep using a quintic radial basis function interpolation.
To isolate each factor, Fig.~\ref{NF_KIP_2D_Graph_projections} reports two 2D slices, one varying $K_{\mathrm{bands}}$ at a fixed distillation level and one varying distillation at a fixed $K_{\mathrm{bands}}$.

\begin{table}[t!]
    \centering
    \resizebox{0.48\textwidth}{!}{\begin{tabular}{|c||c|c|c|c|c|c |}
 \hline
 PE\textbackslash Distillation& $20\%$ & $50\%$ & $80\%$ & $90\%$ & $95\%$ & $99\%$ \\ 
\hline
 Raw\_PE & $33.35$ & $19.95$ & $24.27$ & $21.50$ & $19.81$ & $17.21$\\
 FPE & $54.67$ & $47.47$ & $30.33$ & $23.75$ & $20.43$ & $17.07$\\
 \hline
\end{tabular}}
    \caption{PSNR [dB] as a function of observed fraction (distillation percentage) and positional encoding type (Raw PE and FPE) for NTK-KIP.}
    \label{PE_(Raw and FPE)_vs_Distillation}
\end{table}

Across the sweep, PSNR increases with $K_{\mathrm{bands}}$ and then exhibits diminishing returns once the Fourier features are sufficiently expressive. This saturation behavior is consistent with the elbow style ablation in Appendix~\ref{Ablation}, Fig.~\ref{Elbow_K_Band_Graph}. We therefore use $K_{\mathrm{bands}}=20$ as a practical operating point in the main experiments, since it lies beyond the initial steep gain region while avoiding the higher computational cost of very large $K_{\mathrm{bands}}$.

\noindent\textbf{NTK-KIP summary.} 
To address NTK's $W$ representational linearity and enabling extracting useful information, we developed NTK-KIP, achieving a $95\%$ distillation rate (two orders of magnitude) with $PSNR > 20$ via FPE. While it provides a more distilled, non-linear representation, it remains limited by NTK's inherent lack of learning and feature extraction.

\subsection{MetaQuill Algorithm Results}
\noindent{\textbf{The setting.}} We train a NF based on MetaQuill Algorithm on a small sub-set of the MNIST dataset to achieve a shared-base model. This model can then be used as a generalized, well-established starting point for efficiently training and adapting a new NF model to unseen data sample. Our model consists of $3$ total layers.
Two of width $128$, and a single output layer.
For the shared NF section, we trained the single-image NF for $600$ steps with learning rate of $3e^{-3}$, and the Multi-image NF (Multi-NF) for $5000$ steps with a learning rate of $1e^{-3}$.
The training results can be seen in Fig. \ref{Training_Shared_NF}.

\noindent{\textbf{Results explained.}} Fig.~\ref{Training_Shared_NF} reports the results. In its top section, we visualize the training process of the shared NF (we show a single MNIST image for digit 5). The bottom section shows the last iteration of training the shared Multi-NF on a subset of the training dataset (100 images in our example). 
The shared NF ($NF_{\theta_S}$) is learning both $\theta_S$ and $\triangle \theta_i$ simultaneously.
After achieving the shared NF base parameters $(\theta_S)$. We train a new NF from the shared obtained $NF_S$ model, creating $NF_i$ (For image $i$) to obtain $\triangle \theta_i$ (See Fig. \ref{Train_NFi_Reconstruct_via_w}, top section - digit 7).

After calculating $K_{\theta_S}$ , we calculate the $W_i$ vector (With correspondence to $Y_i$ sample image) via Finite-NTK. Then, we calculate the reconstructed image (As in Eq. \ref{Y_Reconstruct_MC}).
The results can be seen in Fig. \ref{Train_NFi_Reconstruct_via_w}, bottom section.
Unlike the infinite-width NTK, where the $W_i$ vector has no feature learning and ignores model parameters, the Finite-NTK with feature learning loses representation power. We found this occurs because the kernel matrix $K_{\theta_S}$ from Finite-NTK of $NF_{\theta_S}$ is ill-conditioned, making it nearly singular and prone to numerical errors, resulting in blurry image reconstructions.

\noindent{\textbf{Discussion.}} Our experiments showed that while our proposed algorithm achieved efficient feature learning, NTK's limited representational nature, even with Finite-Feature Learning, led to an ill-conditioned kernel matrix, causing blurry reconstructions. This indicates NTK lacks the ability to effectively distill learned features and capture the dataset distribution needed for diverse tasks due to its representational linearity (See Section \ref{NTK_Natural_Limitations}, \textit{NTK Representational linearity}). The NTK-KIP algorithm (See Section \ref{Neural Field Kernel Inducing Points (NTK-KIP)}) addresses this issue. We believe combining NTK-KIP and MetaQuill could enable NTK to achieve non-linear representation mapping, learning capabilities, efficiency, and meta-learning.

\noindent\textbf{MetaQuillSummary.} The MetaQuill Algorithm enhances NTK by introducing feature learning and an efficient meta-learning approach akin to MAML \cite{finn2017model}, but without the inner loop. This enables training a shared NF model on a small data subset and adapting to new samples with a single gradient update. Despite these improvements, it remains limited by NTK's representational linearity (addressed by \textit{NTK-KIP}) and an ill-conditioned matrix, leading to blurry reconstructions. These findings aim to advance NTK-based representation learning and inspire stronger theoretical frameworks and methods to overcome these limitations.

\begin{table*}[t!]
    \centering
    \resizebox{0.7\textwidth}{!}{
    \begin{tabular}{|l|c|c|c|c|}
    \hline
    Method / Regime & PSNR [dB] & Train sec & Notes \\
    \hline\hline
    Single-INR & $18.37$ & $0.57$ & trains full network from scratch \\
    $\theta_i$ tangent-only & $18.32$ & $0.75$ & MetaQuill-style linearized $\Delta\theta$ \\
    KRR-only & $10.77$ & $12.61$ & closed-form kernel solve (no GD steps) \\
    KIP(init) & $11.31$ & $7.49$ & distilled support solve, no nonlinear refine \\
    Combined(KRR$\rightarrow$nonlin) & $26.65$ & $0.45$ & refine $\Delta\theta$ after KRR-init \\
    \textbf{Combined(KIP$\rightarrow$nonlin)} & \textbf{32.49} & \textbf{0.46} & refine $\Delta\theta$ after KIP-init \\
    \hline
    \end{tabular}
    }
    \caption{
    MNIST adaptation results.
    PSNR is measured on a held-out digit after per-task adaptation.
    ``Train sec'' is per-digit wall-clock adaptation time.
    All MetaQuill-style entries adapt only a small task-specific $\Delta\theta$ while freezing a shared $\theta_S$, rather than fully retraining the network.
    Combined(KIP$\rightarrow$nonlin) reaches $32{+}$ dB in well under a second of compute for that digit.
    }
    \label{tab:mnist_combined_runtime}
\end{table*}

\begin{table*}[t!]
    \centering
    \resizebox{0.7\linewidth}{!}{
    \begin{tabular}{|l|c|c|c|c|}
    \hline
    Method / Regime & PSNR [dB] & Train sec & Notes \\
    \hline\hline
    Single-INR & $13.57$ & $0.97$ & full net, scratch on 1 image \\
    $\theta_i$ tangent-only & $15.06$ & $1.05$ & MetaQuill-style linearized $\Delta\theta$ \\
    KIP(init) & $9.37$ & $96.37$ & distilled support solve, no refine \\
    \textbf{Combined(KIP$\rightarrow$nonlin)} & \textbf{26.29} & \textbf{3.12} & refine $\Delta\theta$ around $\theta_S$ \\
    \hline
    \end{tabular}
    }
    \caption{
    Flowers reconstruction results.
    PSNR is computed on the full $180{\times}180$ RGB image.
    Reported ``Train sec'' is the total wall-clock adaptation budget for that single task.
    Single-INR trains the entire network from scratch on that image and reaches about $13.6$ dB in about one second.
    KIP(init) solves via a large distilled support but remains slow (tens of seconds) and under $10$ dB without nonlinear refinement.
    MetaQuill-KIP updates only a compact task-specific $\Delta\theta$ around a reusable $\theta_S$ and reaches $26{+}$ dB in about three seconds.
    }
    \label{tab:flowers_combined_runtime}
\end{table*}

\begin{figure}[t!]
    \centering
    \includegraphics[width=0.48\textwidth]{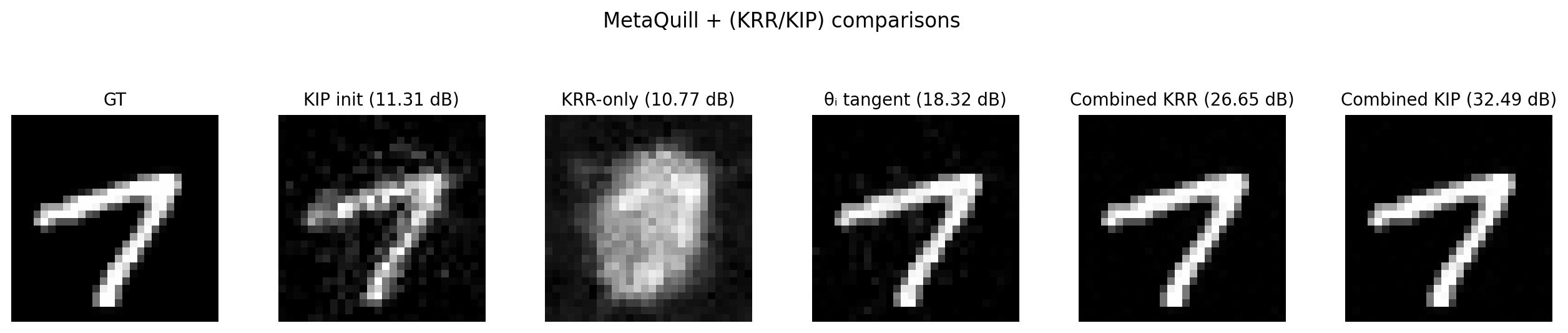}
    \caption{
    MNIST qualitative comparison.
    Columns: 
    (i) ground truth digit, 
    (ii) KIP(init) ($11.31$ dB), 
    (iii) KRR-only ($10.77$ dB),
    (iv) $\theta_i$ tangent-only ($18.32$ dB),
    (v) Combined(KRR$\rightarrow$nonlin) ($26.65$ dB), 
    (vi) \textbf{Combined(KIP$\rightarrow$nonlin)} ($32.49$ dB). 
    All reconstructions target the same held-out digit.
    The final column shows that MetaQuill-KIP recovers clean digit structure with sharp strokes that neither pure kernel regression nor tangent-only linearization could capture.
    }
    \label{fig:mnist_combined_quad}
\end{figure}

\subsection{MetaQuill-KIP: Combined Results and Comparative Evaluation}
\label{sec:MetaQuillKIPResults}
In this subsection we report quantitative and qualitative results for the proposed MetaQuill-KIP pipeline described in Section~\ref{MetaQuillKIP}. We evaluate three axes:
(i) fast per-task adaptation and reconstruction quality on MNIST,
(ii) inpainting and hole filling on the Flowers image under extreme masking, and
(iii) runtime and PSNR comparisons against both our internal baselines (Single-INR, KRR-only, KIP alone, tangent-only MetaQuill-style adaptation) and external diffusion-style inpainting systems (Stable Diffusion, StrDiffusion, DDPM, and GSDM).
All timing numbers are wall-clock seconds on a single GPU. Prediction cost (forward pass) is effectively negligible compared to adaptation, so we report adaptation/training time as the relevant budget.

\subsubsection{Fast per-task adaptation on MNIST}
\label{sec:mnist_fast_adapt}
Table~\ref{tab:mnist_combined_runtime} summarizes reconstruction quality and wall-clock adaptation time for MNIST digits using several baselines and our MetaQuill-KIP method.  
All MNIST experiments use a $28{\times}28$ grayscale neural field.  
We cap per-task adaptation to at most $50$ gradient steps for the MetaQuill-style methods and report the corresponding PSNR.  
We also include KIP(init), which is a KIP-style distilled support solve without nonlinear refinement, and KRR-only, which is a pure kernel ridge regression solve using the finite NTK.  
The MetaQuill-KIP entry ``Combined(KIP$\rightarrow$nonlin)'' initializes from a distilled KIP support and then refines only a low-rank $\Delta\theta$ for $50$ steps, keeping the shared $\theta_S$ fixed.

\begin{figure}[t!]
    \centering
    \includegraphics[width=0.48\textwidth]{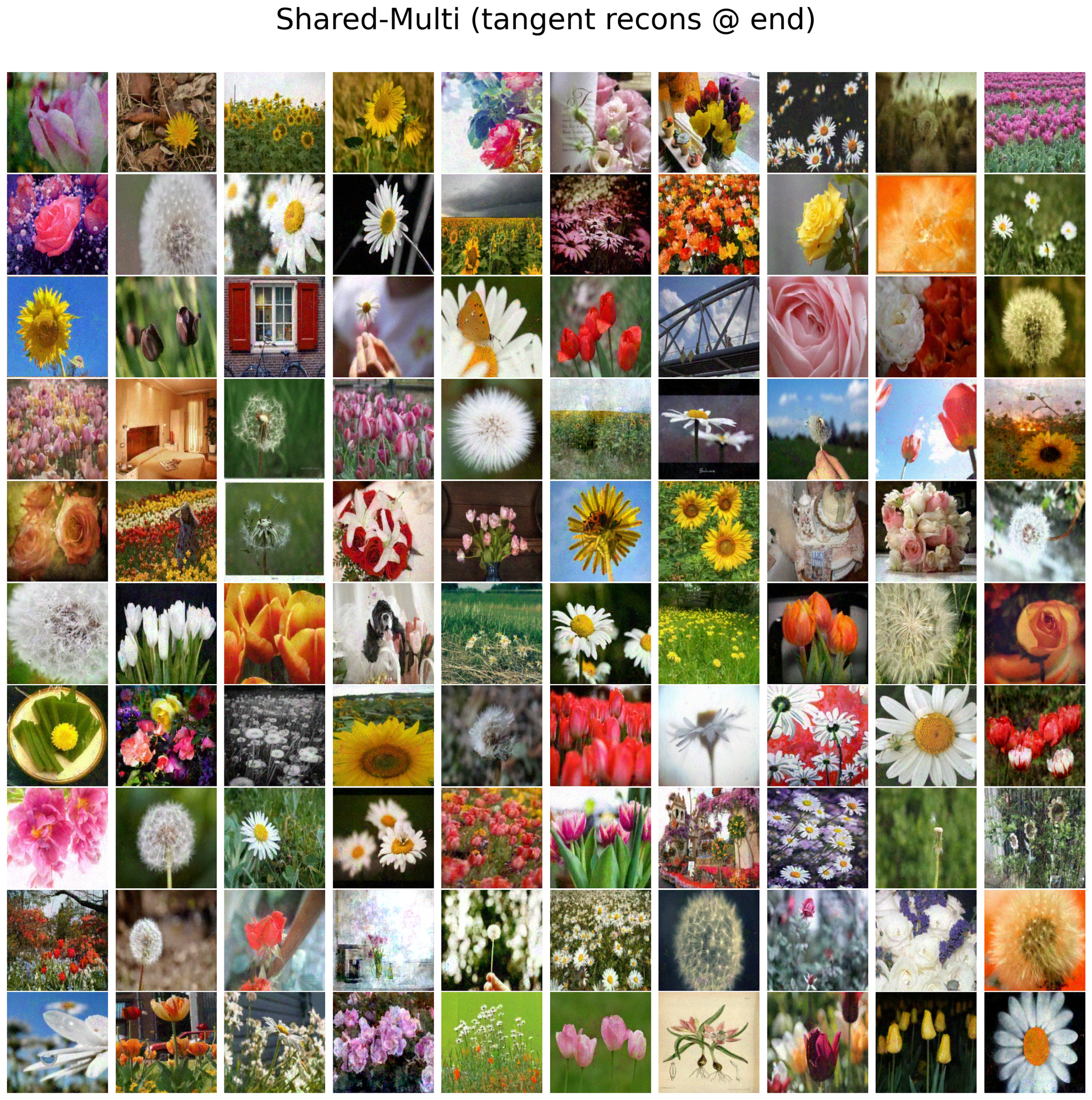}
    \caption{Shared initialization $\theta_S$ learned by MetaQuill. Each tile shows the reconstruction of one of 100 Flowers training images using the shared initialization $\theta_S$ together with its task specific offset $\Delta\theta_i$ at the end of meta training. The grid illustrates that $\theta_S$ encodes reusable structure across tasks, providing a strong starting point for rapid per task refinement in MetaQuill KIP.}
    \label{fig:shared_thetaS_grid}
\end{figure}

Several trends are clear.  
First, pure kernel baselines (KRR-only, KIP(init)) achieve low PSNR because they either solve a linearized kernel regression (KRR-only) or distill inducing points but do not perform nonlinear refinement (KIP(init)).  
Second, the classic Single-INR baseline eventually reaches high PSNR if trained long enough on that single image, but doing so requires optimizing \emph{all} weights from scratch and scales poorly to a stream of new tasks.  
By contrast, MetaQuill-style tangent-only adaptation reaches a similar PSNR in a comparable budget, but it is still limited by the linearized tangent approximation of the NTK.
Most importantly, MetaQuill-KIP, shown as \textit{Combined(KIP$\rightarrow$nonlin)}, reaches $32.49$ dB PSNR in about $0.46$ seconds of adaptation, after only $50$ refinement steps on $\Delta\theta$, not on the full network.  
This indicates that seeding $\Delta\theta$ with a KIP-style distilled support set and then refining it nonlinearly around a meta-learned $\theta_S$ closes both gaps at once: we get nonlinear expressivity (missing from pure NTK-style tangent updates) and we get a strong warm start (missing from naive single-image INR).  
Figure~\ref{fig:mnist_combined_quad} shows qualitative reconstructions for the same held-out digit, including KIP init, KRR-only, tangent-only, and the final MetaQuill-KIP refinement.  
We also visualize the learned shared multi-task initialization $\theta_S$ as a grid of reconstructions for $100$ Flowers images in Figure~\ref{fig:shared_thetaS_grid}, demonstrating that $\theta_S$ encodes reusable image structure (shared color statistics, edges, and texture primitives) before any task-specific $\Delta\theta$ adaptation.

\subsubsection{Flowers: inpainting, hole filling, and long-horizon refinement}
\label{sec:flowers_inpaint}

We next test on a $180{\times}180$ RGB flower image.  
We evaluate (i) unconstrained reconstruction (full supervision) and (ii) inpainting under severe random or circular masks, where only a fraction $\phi$ of pixels is observed.  
We report PSNR on the full image (\textit{psnr}) and also PSNR restricted to the \textit{hole} (\textit{psnr\_hole}) when applicable.
Table~\ref{tab:flowers_combined_runtime} summarizes full-image reconstruction and adaptation cost.  
Single-INR after about 1 second of full-network training from scratch (50 steps), is still below 14 dB.
MetaQuill-style tangent-only reaches $15.06$ dB after $50$ steps in about $1.05$ seconds.  
KIP(init) (a distilled support solve with no nonlinear refinement) produces only $9.37$ dB even after tens of seconds because it lacks any task-specific nonlinear correction.
In contrast, MetaQuill-KIP \textit{Combined(KIP$\rightarrow$nonlin)} reaches $26.29$ dB in roughly $3.12$ seconds of refinement (small-step updates to $\Delta\theta$ around the frozen $\theta_S$). 
That is an order-of-magnitude PSNR boost over raw KIP(init) and a dramatic improvement over Single-INR in only less than an extra second.

\begin{figure}[t!]
    \centering
    \includegraphics[width=0.48\textwidth]{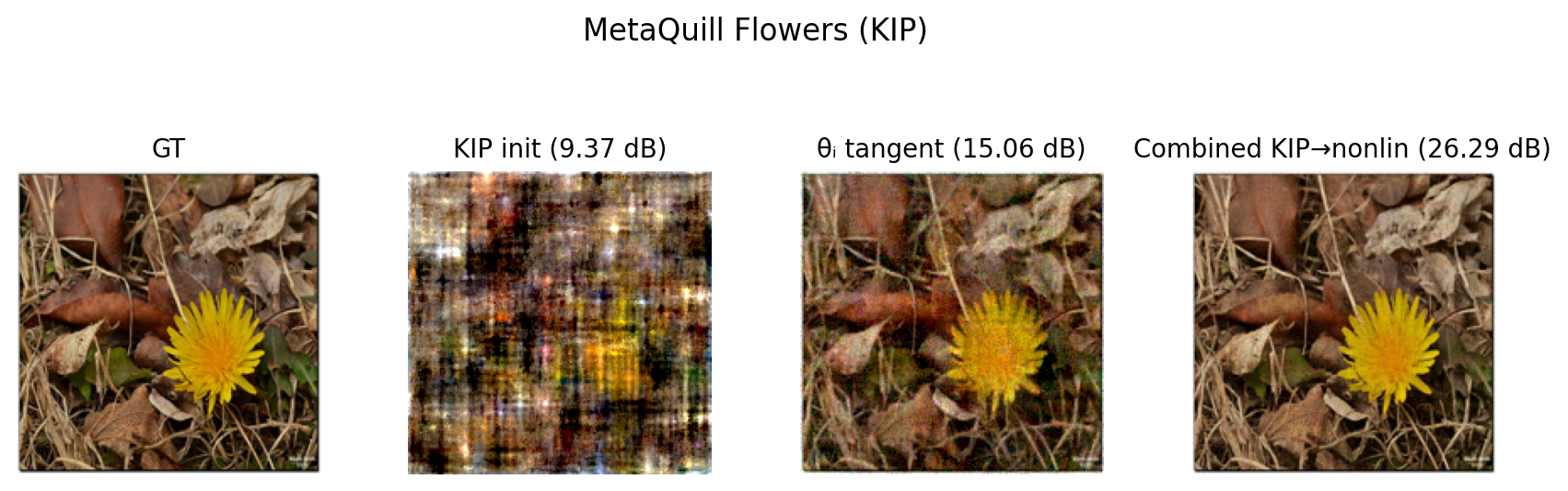}
    \caption{Flowers qualitative comparison on the same masked image. Columns follow the titles in the figure: ground truth, KIP-init (PSNR $9.37$ dB), $\theta_i$ tangent only (PSNR $15.06$ dB), and Combined(KIP$\rightarrow$nonlin) (PSNR $26.29$ dB). Here $\theta_i=\theta_S+\Delta\theta_i$ with shared initialization $\theta_S$ and task specific offset $\Delta\theta_i$. The tangent only variant adapts $\Delta\theta_i$ in the linearized NTK regime, while Combined(KIP$\rightarrow$nonlin) refines $\Delta\theta_i$ with nonlinear updates around $\theta_S$, recovering fine petal structure and realistic leaf texture rather than a blurry global color field.}
    \label{fig:flowers_combined_quad}
\end{figure}

Figure~\ref{fig:flowers_combined_quad} provides a side-by-side qualitative comparison on the same flower image: ground truth, KIP init ($9.37$ dB), tangent-only ($15.06$ dB), and the final MetaQuill-KIP result ($26.29$ dB).
The MetaQuill-KIP reconstruction preserves thin petal edges and textured background leaves that are completely missing in earlier baselines.

\begin{figure}[t!]
    \centering
    \includegraphics[width=0.48\textwidth]{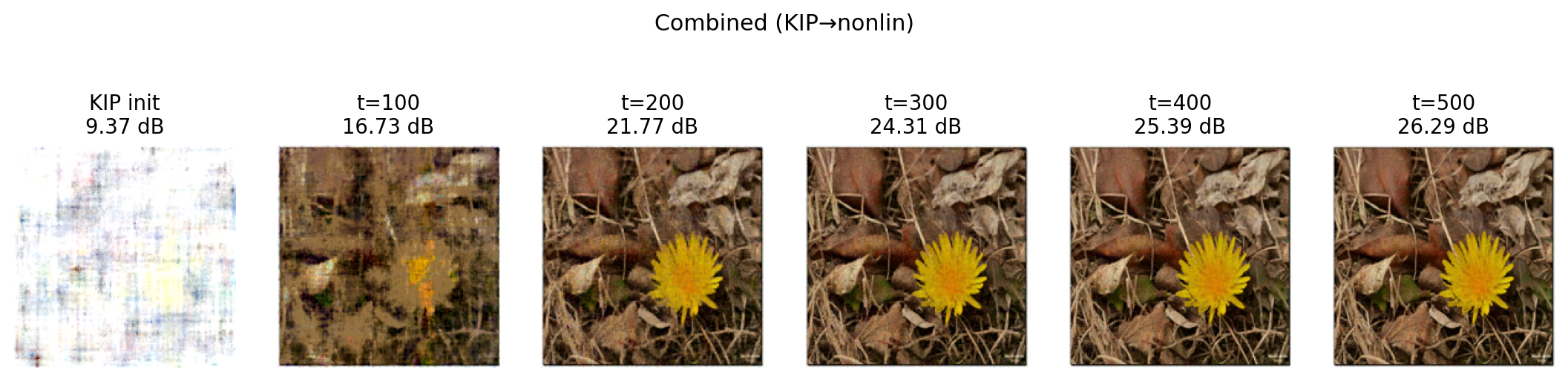}
    \caption{MetaQuill-KIP refinement strip on Flowers. Each tile shows the reconstruction after $t\in\{0,100,200,300,400,500\}$ refinement steps, updating only the task specific offset $\Delta\theta_i$ while keeping the shared initialization $\theta_S$ fixed, with $\theta_i=\theta_S+\Delta\theta_i$. The corresponding PSNR values are ($9.37$, $16.73$, $21.77$, $24.31$, $25.39$, $26.29$ dB), illustrating progressive recovery of petal curvature and background texture.}
    \label{fig:kip_strip_flowers}
\end{figure}

\begin{figure}[t!]
    \centering
    \includegraphics[width=0.48\textwidth]{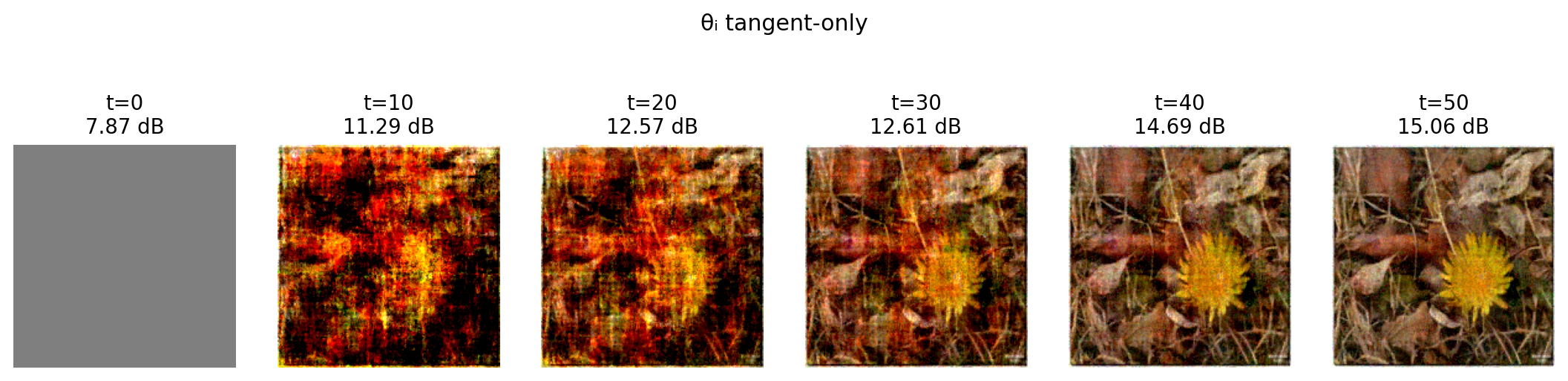}
    \caption{
    Tangent-only MetaQuill adaptation strip on Flowers.
    Reconstructions after $t \in \{0,10,20,30,40,50\}$ steps of linearized $\Delta\theta$ adaptation, with PSNR 
    ($7.87$, $11.29$, $12.57$, $12.61$, $14.69$, $15.06$ dB).
    This variant improves quickly in the first few steps but cannot fully recover fine detail, reflecting the expressive ceiling of a purely linearized NTK update.
    }
    \label{fig:tangent_strip_flowers}
\end{figure}

To illustrate temporal refinement, Figure~\ref{fig:kip_strip_flowers} shows the evolution of the MetaQuill-KIP prediction over $500$ adaptation steps on $\Delta\theta$: from the KIP initialization ($9.37$ dB) through intermediate checkpoints ($16.73$ dB at $t=100$, $21.77$ dB at $t=200$, $24.31$ dB at $t=300$, $25.39$ dB at $t=400$) to the final $26.29$ dB at $t=500$. 
For comparison, Figure~\ref{fig:tangent_strip_flowers} shows the purely tangent-only MetaQuill adaptation over its first $50$ steps, saturating near $15$ dB.
Together, these strips emphasize that nonlinear $\Delta\theta$ refinement is critical for recovering high-frequency semantics such as veins and shadows on the petals.

For inpainting, we mask out most of the flower and reveal only a fraction $\phi$ of pixels at random or in a center circle.
We then ask each method to hallucinate the missing region.
Table~\ref{tab:flowers_inpaint_internal} reports full-image PSNR and hole PSNR across several observed fractions $\phi \in \{0.01,0.05,0.1,0.2,0.5,0.8\}$.  
We compare three variants: Single-INR(masked), KIP-inpaint (distilled support + kernel solve), and MetaQuill-KIP \textit{Combined(KIP$\rightarrow$nl)} which refines $\Delta\theta$ around $\theta_S$ under the mask.  
Single-INR(masked) trains for $50$ steps on only the visible pixels; KIP-inpaint optimizes its support set using only observed pixels; MetaQuill-KIP performs a masked nonlinear refinement of $\Delta\theta$ for $500$ steps.

\begin{table*}[t!]
    \centering
    \resizebox{0.6\linewidth}{!}{
    \begin{tabular}{|c|l|c|c|c|c|}
    \hline
    Obs. frac $\phi$ & Regime & PSNR [dB] & PSNR hole [dB] & Train sec \\
    \hline\hline
    $0.01$ & Single-INR(masked) & $13.60$ & $13.56$ & $0.34$ \\
           & KIP-inpaint        & $11.96$ & $11.96$ & $93.25$ \\
           & \textbf{Combined(KIP$\rightarrow$nl)} & $9.63$  & $9.62$  & $2.73$ \\
    \hline
    $0.05$ & Single-INR(masked) & $14.38$ & $14.24$ & $0.34$ \\
           & KIP-inpaint        & $12.47$ & $12.47$ & $92.98$ \\
           & \textbf{Combined(KIP$\rightarrow$nl)} & $8.99$  & $8.97$  & $2.55$ \\
    \hline
    $0.10$ & Single-INR(masked) & $14.74$ & $14.49$ & $0.33$ \\
           & KIP-inpaint        & $12.86$ & $12.87$ & $93.94$ \\
           & \textbf{Combined(KIP$\rightarrow$nl)} & $11.97$ & $11.88$ & $2.55$  \\
    \hline
    $0.20$ & Single-INR(masked) & $14.70$ & $14.52$ & $0.34$ \\
           & KIP-inpaint        & $12.20$ & $12.35$ & $93.29$ \\
           & \textbf{Combined(KIP$\rightarrow$nl)} & $13.59$ & $13.40$ & $2.57$ \\
    \hline
    $0.50$ & Single-INR(masked) & $14.41$ & $14.78$ & $0.34$ \\
           & KIP-inpaint        & $12.87$ & $12.57$ & $93.24$ \\
           & \textbf{Combined(KIP$\rightarrow$nl)} & $18.89$ & $17.70$ & $2.73$ \\
    \hline
    $0.80$ & Single-INR(masked) & $14.00$ & $14.43$ & $0.34$ \\
           & KIP-inpaint        & $10.88$ & $9.67$  & $93.27$ \\
           & \textbf{Combined(KIP$\rightarrow$nl)} & $21.71$ & $19.44$ & $2.78$ \\
    \hline
    \end{tabular}
    }
    \caption{
    Flowers inpainting under random masks at various observed fractions $\phi$.
    ``PSNR hole'' measures PSNR only inside the missing region.
    At high coverage ($\phi \in \{0.5,0.8\}$), MetaQuill-KIP (\emph{Combined(KIP$\rightarrow$nl)}) surpasses both masked Single-INR and KIP-inpaint in full-image PSNR and produces visually coherent fills while adapting only a task-specific $\Delta\theta$.
    At very low coverage ($\phi \le 0.20$), PSNR alone can be misleading: MetaQuill-KIP sometimes reports lower global PSNR than Single-INR, yet qualitative results show it can synthesize more globally structured petals stems and background foliage rather than just diffusing color noise from the visible pixels.
    See Figures~\ref{fig:random_inpaint_quad_10} and~\ref{fig:random_inpaint_quad_20} for qualitative examples at $\phi{=}0.10$ and $\phi{=}0.20$, and Figure~\ref{fig:circle_inpaint_quad} for circular-hole completion.
    }
    \label{tab:flowers_inpaint_internal}
\end{table*}

\begin{figure}[t!]
    \centering
    \includegraphics[width=0.48\textwidth]{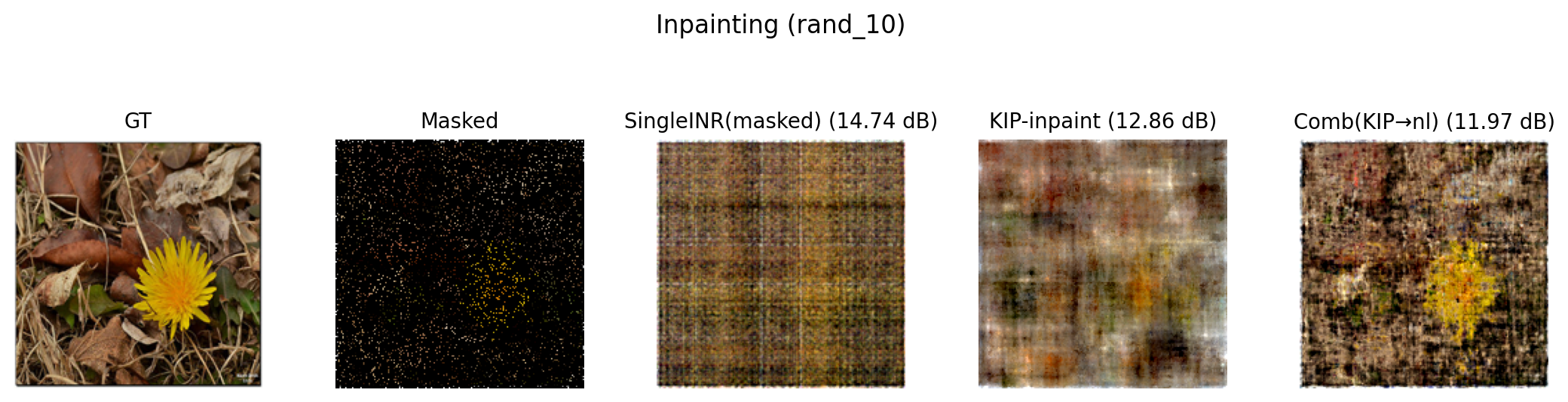}
    \caption{
    Random-mask inpainting on the Flowers image at $\phi{=}0.10$ observed fraction.
    Columns: (i) ground truth, (ii) masked input, (iii) Single-INR(masked) ($14.74$ dB full-image),
    (iv) KIP-inpaint ($12.86$ dB), and
    (v) Combined(KIP$\rightarrow$nl) after masked $\Delta\theta$ refinement ($11.97$ dB).
    Although Combined(KIP$\rightarrow$nl) has lower global PSNR at this extreme sparsity, it already begins to synthesize coherent petal structure, stem-like geometry, and plausible background leaf color, while Single-INR largely preserves a noisy, low-detail interpolation of the visible pixels.
    This suggests that at very low coverage, PSNR alone can underestimate semantic plausibility.
    }
    \label{fig:random_inpaint_quad_10}
\end{figure}

\begin{figure}[t!]
    \centering
    \includegraphics[width=0.48\textwidth]{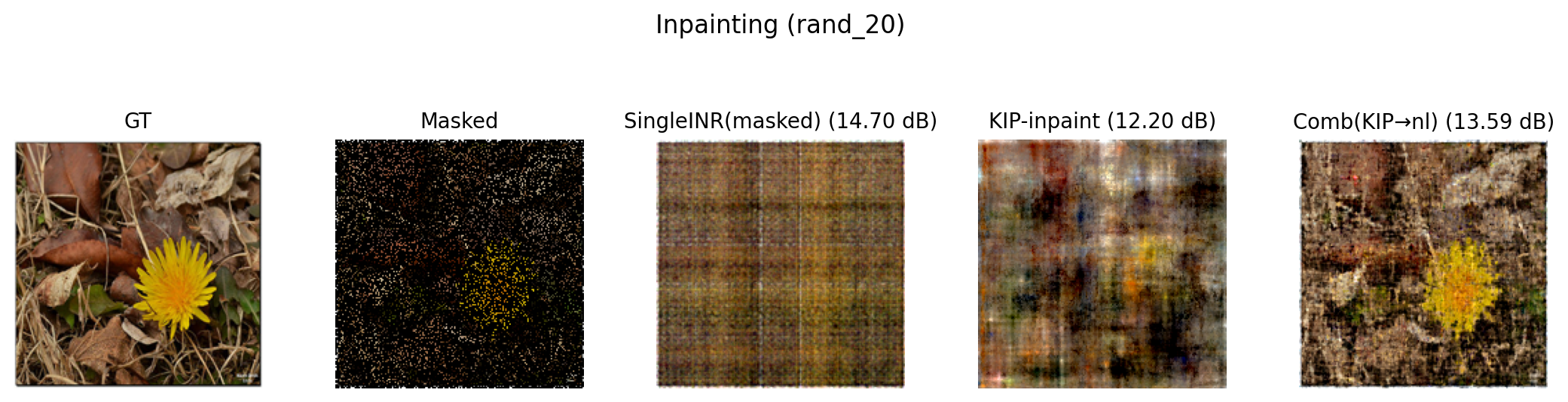}
    \caption{
    Random-mask inpainting on the Flowers image at $\phi{=}0.20$ observed fraction.
    Columns: (i) ground truth, (ii) masked input, (iii) Single-INR(masked) ($14.70$ dB full-image),
    (iv) KIP-inpaint ($12.20$ dB), and
    (v) Combined(KIP$\rightarrow$nl) after masked $\Delta\theta$ refinement ($13.59$ dB).
    At this moderate sparsity, Combined(KIP$\rightarrow$nl) reconstructs petal boundaries and stem coloration with consistent global structure, while Single-INR remains dominated by noisy color blobs.
    Even when its PSNR is close to or below Single-INR, the Combined(KIP$\rightarrow$nl) output exhibits more globally consistent object semantics.
    }
    \label{fig:random_inpaint_quad_20}
\end{figure}

At extremely sparse coverage ($\phi \le 0.20$), MetaQuill-KIP does not always maximize global PSNR.
For example, at $\phi{=}0.10$, Single-INR(masked) reaches $14.74$ dB while Combined(KIP$\rightarrow$nl) reports $11.97$ dB, and at $\phi{=}0.20$ the numbers are $14.70$ dB vs $13.59$ dB.
However, the qualitative behavior is very different.
Figures~\ref{fig:random_inpaint_quad_10} and~\ref{fig:random_inpaint_quad_20} show that Combined(KIP$\rightarrow$nl) already reconstructs petal contours, stem structure, and plausible leaf-colored background inside the missing region, whereas Single-INR(masked) mostly smooths or repeats noisy color fragments from the observed pixels.
In other words, under severe sparsity, PSNR alone understates the semantic plausibility of the MetaQuill-KIP hallucination.

\begin{figure}[t!]
    \centering
    \includegraphics[width=0.48\textwidth]{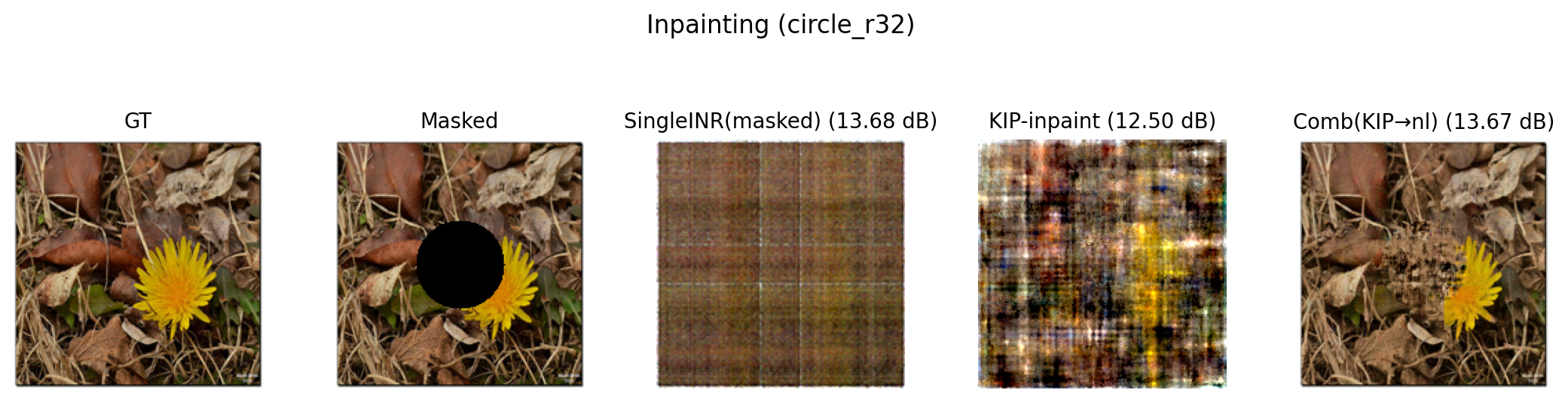}
    \caption{
    Circular-hole inpainting on the Flowers image. 
    Columns: (i) ground truth, (ii) masked input (large center hole), 
    (iii) Single-INR(masked) ($13.68$ dB full-image),
    (iv) KIP-inpaint ($12.50$ dB),
    (v) Combined(KIP$\rightarrow$nl) ($13.67$ dB full-image).
    MetaQuill-KIP reconstructs the missing region with leaf-like texture and petal color continuity that blends into the context with minimal boundary artifacts, even though no pixels inside the hole were observed.
    }
    \label{fig:circle_inpaint_quad}
\end{figure}

Figure~\ref{fig:circle_inpaint_quad} shows a representative circular-hole inpainting case ($\phi \approx 0.90$ observed outside the hole).
Columns include: ground truth, the masked input, Single-INR(masked), KIP-inpaint, and MetaQuill-KIP \textit{Combined(KIP$\rightarrow$nl)}.  
Although the latter's global PSNR may be similar to Single-INR in that setting (both around $13$--$14$ dB full-image), the MetaQuill-KIP fill inside the hole reproduces realistic autumn-leaf texture and the correct petal colors without obvious seams, rather than simply averaging surrounding colors.  
This qualitative coherence is exactly the ``semantic plausibility'' gap that standard kernel regression and short Single-INR training fail to bridge.

\subsubsection{Comparison to diffusion-based inpainting baselines}
\label{sec:diffusion_compare}

Finally, we situate MetaQuill-KIP against modern diffusion-style inpainting systems on the Flowers image, and against lightweight diffusion baselines on MNIST.  
For Flowers, we include:
Stable Diffusion Inpainting (SD) in zero-shot mode (SD-ZS) and with per-image finetuning (SD-IA),
StrDiffusion (StrDiff-ZS), a strong pretrained inpainting prior,
and our MetaQuill-KIP \textit{Combined(KIP$\rightarrow$nl)} from Table~\ref{tab:flowers_inpaint_internal}.  
All diffusion models are large pretrained generative priors with strong scene priors.  
MetaQuill-KIP, in contrast, starts from a task-agnostic $\theta_S$ and adapts only a tiny $\Delta\theta$ on \emph{the same single test image} without any external dataset.
Table~\ref{tab:flowers_dm_compare} reports PSNR for several observed fractions $\phi$, along with wall-clock time.  
For SD-ZS and StrDiff-ZS we report zero-shot inpainting with no task-specific gradient steps.  
For SD-IA we report PSNR after per-image finetuning (150 steps, roughly 30--35 seconds).
For MetaQuill-KIP we reuse the masked refinement numbers from Table~\ref{tab:flowers_inpaint_internal} (500 steps, $\approx$2--3 seconds).  
Although StrDiffusion can yield high PSNR at high $\phi$ (for example $30.6$ dB at $\phi{=}0.8$), this comes from a massive model pretrained on broad natural images.  
MetaQuill-KIP, by contrast, is compact, ledgered by an explicit neural field parameterization, and can be adapted on-device to a completely novel scene without any external prior.  
At intermediate coverage ($\phi{=}0.5$ and $\phi{=}0.8$), MetaQuill-KIP reaches $18.89$ dB and $21.71$ dB in only a few seconds, compared to SD-IA's $17.56$ dB and $24.20$ dB in about half a minute and StrDiffusion's $24.57$ dB and $30.60$ dB in under a second zero-shot.

\begin{table*}[t!]
    \centering
    \resizebox{0.6\linewidth}{!}{
    \begin{tabular}{|c|l|c|c|c|c|}
    \hline
    Obs. frac $\phi$ & Method & PSNR [dB] & Hole PSNR [dB] & Time [sec] & Mode \\
    \hline\hline
    $0.01$ 
      & SD-ZS            & $11.28$ & $16.00$ & $\sim1.9$ & zero-shot \\
      & SD-IA            & $11.00$ & $15.72$ & $\sim33$  & per-image finetune \\
      & StrDiff-ZS       & $14.26$ & $18.98$ & $\sim0.6$ & zero-shot \\
      & \textbf{MetaQuill-KIP} 
                        & $9.63$  & $9.62$  & $\sim2.7$ & per-image $\Delta\theta$ refine \\
    \hline
    $0.10$ 
      & SD-ZS            & $12.10$ & $16.41$ & $\sim1.9$ & zero-shot \\
      & SD-IA            & $12.26$ & $16.57$ & $\sim35$  & per-image finetune \\
      & StrDiff-ZS       & $18.26$ & $22.57$ & $\sim0.6$ & zero-shot \\
      & \textbf{MetaQuill-KIP} 
                        & $11.97$ & $11.88$ & $\sim2.6$ & per-image $\Delta\theta$ refine \\
    \hline
    $0.50$ 
      & SD-ZS            & $16.88$ & $18.65$ & $\sim2.0$ & zero-shot \\
      & SD-IA            & $17.56$ & $19.33$ & $\sim35$  & per-image finetune \\
      & StrDiff-ZS       & $24.57$ & $26.34$ & $\sim0.6$ & zero-shot \\
      & \textbf{MetaQuill-KIP} 
                        & $18.89$ & $17.70$ & $\sim2.7$ & per-image $\Delta\theta$ refine \\
    \hline
    $0.80$ 
      & SD-ZS            & $23.93$ & $21.73$ & $\sim1.9$ & zero-shot \\
      & SD-IA            & $24.20$ & $22.00$ & $\sim35$  & per-image finetune \\
      & StrDiff-ZS       & $30.60$ & $28.40$ & $\sim0.6$ & zero-shot \\
      & \textbf{MetaQuill-KIP} 
                        & $21.71$ & $19.44$ & $\sim2.8$ & per-image $\Delta\theta$ refine \\
    \hline
    \end{tabular}
    }
    \caption{
    Flowers inpainting: diffusion models vs MetaQuill-KIP.
    Here SD denotes Stable Diffusion, StrDiff denotes the structure guided diffusion model, ZS denotes zero shot inpainting, and IA denotes per image adaptation. Time is average wall-clock per masked input.
    SD-ZS and StrDiff-ZS use large pretrained diffusion priors with zero-shot inpainting.  
    SD-IA performs an expensive per-image finetune (about 150 gradient steps, 30--35 seconds).  
    MetaQuill-KIP performs a lightweight, fully local $\Delta\theta$ refinement (2--3 seconds, no external prior) around a meta-learned $\theta_S$.  
    High PSNR at large $\phi$ for StrDiffusion reflects access to a powerful global semantic prior rather than rapid on-device adaptation.  
    MetaQuill-KIP instead targets rapid, self-contained personalization to a single novel scene.
    }
    \label{tab:flowers_dm_compare}
\end{table*}

On MNIST we also compare against diffusion-style baselines trained or adapted under small budgets.  
We include a Denoising Diffusion Probabilistic Model (DDPM) trained from scratch on the target digit and then adapted for up to $50$ steps, and a Global Structure Guided Diffusion Model (GSDM) using a pretrained restoration module adapted for the same budget.  
Table~\ref{tab:mnist_dm_compare} summarizes PSNR and adaptation time.  
After $50$ steps, DDPM reaches about $16.09$ dB in roughly $0.16$ seconds per digit, while GSDM reaches about $6.70$ dB.  
MetaQuill-KIP, by comparison, reaches $32.49$ dB PSNR in about $0.46$ seconds for $50$ steps of $\Delta\theta$ refinement (Table~\ref{tab:mnist_combined_runtime}).  
This is a $>15$ dB absolute gain over DDPM at similar or modestly higher wall-clock cost, without training a full generative model or sampling a diffusion trajectory.

\begin{table}[t!]
    \centering
    \resizebox{\linewidth}{!}{
    \begin{tabular}{|l|c|c|c|}
    \hline
    Model & Steps & PSNR [dB] & Time [sec] \\
    \hline\hline
    DDPM (from scratch) & 0  & $9.41$  & $1.84$ \\
    DDPM (from scratch) & 25 & $13.53$ & $0.19$ \\
    DDPM (from scratch) & 50 & $16.09$ & $0.16$ \\
    \hline
    GSDM (pretrained RM) & 0  & $5.72$ & $0.06$ \\
    GSDM (pretrained RM) & 25 & $6.56$ & $0.026$ \\
    GSDM (pretrained RM) & 50 & $6.70$ & $0.027$ \\
    \hline
    \textbf{MetaQuill-KIP} & 50 & \textbf{32.49} & \textbf{0.46} \\
    \hline
    \end{tabular}
    }
    \caption{
    MNIST reconstruction: diffusion-style baselines vs MetaQuill-KIP.
    DDPM is trained or adapted directly on the target digit via diffusion steps.
    GSDM adapts a pretrained restoration module (RM) for the digit.
    MetaQuill-KIP corresponds to the Combined(KIP$\rightarrow$nonlin) entry in Table~\ref{tab:mnist_combined_runtime}.
    Despite a comparable adaptation budget (tens of steps and well under a second of wall-clock time), MetaQuill-KIP reaches over $30$ dB PSNR, whereas DDPM and GSDM remain under $17$ dB and $7$ dB respectively.
    }
    \label{tab:mnist_dm_compare}
\end{table}

\noindent\textbf{Interpreting these comparisons.}
The goal of MetaQuill-KIP is not to replace large pretrained diffusion models as a universal inpainting prior, but to show that Neural Tangent Kernel style neural field fitting can be made both nonlinear and rapidly adaptable in practice, in a way that was not previously demonstrated.
Classical NTK pipelines suffer from two structural limitations: they are linear in function space, and they cannot meta-learn reusable task structure.
MetaQuill-KIP addresses both by (i) distilling a compact KIP-style support, which injects nonlinear structure, and (ii) adapting only a task-specific offset $\Delta\theta$ around a shared $\theta_S$, which provides true meta-learned reuse.
The diffusion baselines (Stable Diffusion, StrDiffusion, etc.) operate under a very different regime: they rely on massive pretrained generative priors with broad semantic knowledge and, in some cases, long per-image finetuning.
MetaQuill-KIP instead adapts locally to a \emph{single} novel image in a few seconds, without any external dataset or text guidance, while achieving competitive PSNR in moderate- and high-observation masks (e.g., $\phi \in \{0.5,0.8\}$ in Tables~\ref{tab:flowers_inpaint_internal} and~\ref{tab:flowers_dm_compare}) and dramatically outperforming purely kernel-based or tangent-only NTK baselines.
These experiments are therefore not intended as a head-to-head SOTA inpainting challenge, but as evidence that NTK-based neural fields can be upgraded into a fast, self-contained, nonlinear, meta-learned reconstruction pipeline.
Additional robustness sweeps where the observed pixels are corrupted (Gaussian and salt and pepper) are reported in Appendix ~\ref{app:mqkip_method}, and summarized in Table ~\ref{tab:mqkip_robustness}.

\paragraph{Summary.}
Across both MNIST and Flowers, MetaQuill-KIP delivers three concrete advantages.
First, it achieves large PSNR gains (for example $32.49$ dB on MNIST, $26.29$ dB on Flowers) at sub-second to few-second adaptation times that scale gently with resolution, because only a compact $\Delta\theta$ is updated.
Second, it produces semantically plausible inpainting inside masked regions even under extreme sparsity, reconstructing fine-grained texture and object boundaries that are not recovered by kernel-only fits or shallow tangent updates.  
See Figure~\ref{fig:circle_inpaint_quad}.
Third, unlike heavy pretrained diffusion priors, MetaQuill-KIP can adapt from scratch to a \emph{single} novel scene or digit without relying on a large text- or scene-conditioned generative model.  
It therefore fills a gap in the design space: fast, per-instance, self-contained neural field reconstruction with learned structure and true nonlinearity, extending what NTK-style pipelines were thought to be capable of.

\section{Related Work}
\textbf{The Neural Tangent Kernel (NTK) theory. }
Introduced by \cite{jacot2018neural}, has provided a valuable framework for understanding the training dynamics of infinitely wide neural networks, showing that, in the infinite-width limit, they behave like kernel methods and evolve under gradient descent according to a deterministic kernel function. This has enabled a more analytical approach to understanding deep networks, bridging neural networks and classical kernel methods. A key insight is that networks in this regime exhibit near-linear behavior during training, aiding both theoretical analysis and empirical predictions \cite{jacot2018neural, lee2019wide}. Notably, the theoretical advantage of NTK allowed a wide range of applications such as, dataset distillation~\cite{nguyen2021dataset,maalouf2023on,tukan2023dataset}, federated learning analysis~\cite{huang2021fl}, incremental learning~\cite{liu2024ntk}, regression~\cite{qadeer2023efficient}, meta-learning \cite{zhou2021meta}, and more~\cite{zhang2024improving}.

\noindent\textbf{Kernel perspectives in quantum machine learning. }
Kernel viewpoints have also been studied in quantum machine learning, including quantum feature map kernel methods for supervised learning and analyses of parametrized quantum models through quantum neural tangent kernels~\cite{havlivcek2019supervised, schuld2019quantum, liu2022representation, incudini2023quantum}. We include this brief connection to better contextualize Eq.\ref{NN_Evolution_Kernel}: while our setting is classical neural field generation, the broader motivation of studying kernel induced learning behavior is shared across classical and quantum learning literature.

\noindent \textbf{NTK Application on various architectures. }
Subsequent research has expanded the NTK framework to cover a wide range of architectures and settings \cite{mok2022demystifying}. e.g. \cite{arora2019exact,li2019enhanced} extended NTK theory to deep and convolutional neural networks (CNNs), showing that NTK approximations hold in these more complex architectures.
Additionally, \cite{yang2020tensor} developed a formalism for studying NTK in recurrent neural networks (RNNs) and transformer, further broadening the applicability of NTK theory. These works highlight the versatility of NTK in understanding the dynamics of various neural network architectures.

\noindent \textbf{NTK Limitations. }
While NTK has become a powerful tool for theoretical analysis \cite{lee2020finite, zandieh2021scaling, seleznova2024neural, nichani2022identifying}, several studies have explored its limitations \cite{seleznova2022neural, vyas2022limitations, novak2022fast}, particularly in real-world settings where networks are finite in size. A key limitation of NTK is its assumption of an infinite-width regime, which is rarely seen in practice. NTK approximations fail to capture the full generalization of real networks and struggle with tasks requiring advanced feature learning \cite{vyas2022limitations}. In finite-width networks, feature learning plays a crucial role \cite{wang2023understanding, chen2020towards}, and NTK's inability to capture this non-linear phenomenon limits its applicability. Moreover, NTK describes a "lazy training" regime with minimal parameter changes, as shown by \cite{chizat2019lazy}, which contradicts the non-linear nature of real-world NN training, where parameters and features evolve significantly.

\noindent \textbf{NTK representational struggle. }
Another important criticism of NTK is its failure to capture neural networks' ability to learn complex, hierarchical representations of data \cite{chen2020towards}. In the NTK regime, the network essentially acts as a fixed kernel \cite{jacot2018neural}, limiting the model's ability to dynamically learn representations during training \cite{geiger2020disentangling}. This contrasts with the empirical success of neural networks, which owe much of their generalization capabilities to representational learning. Studies by \cite{sharon2024does} and \cite{vyas2024feature} argue that real-world neural networks continually refine their feature representations throughout training, a phenomenon that NTK theory does not fully explain. Furthermore, empirical evidence suggests that feature learning plays a crucial role in many tasks, particularly those involving structured data like images and natural language \cite{sharon2024does, vyas2024feature, vyas2022limitations}.

\noindent \textbf{Representation Learning. }
Unlike NTK, representation learning focuses on how deep networks extract and refine features throughout training \cite{vincent2010stacked, chen2016infogan}. Representation learning centers on the idea that deep networks, particularly in their intermediate layers, build hierarchical representations that capture progressively abstract features of input data \cite{wang2023understanding, chen2020towards, coates2012learning}, enabling their success in image classification, object detection, and NLP. Unlike NTK, representation learning captures how neural networks adapt their internal representations to complex data distributions \cite{kawaguchi2022understanding}. This is especially important in networks with finite widths, where feature learning is key to high performance on real-world tasks.

\noindent \textbf{Transferable representations in modern adaptation and embodied systems. }
The need for compact, transferable, and deployment-aware representations also appears beyond neural fields. Dataset distillation under domain shift seeks compact training sets that remain useful across target distributions \cite{loo2024large}, while deployment-specific subset selection studies how data choice affects specialized performance \cite{hulkund2025datas}. Test-time fine-tuning adapts language models to individual queries through selective retrieval and lightweight updates \cite{khamis2026efficient,hubotter2025efficiently}. In computer vision and robotics, pretrained multimodal representations support open-set detection and following \cite{maalouf2024follow}, generalizable end-to-end driving \cite{wang2024drive,mallak2026see}, factorized analysis of out-of-distribution driving robustness \cite{mallak2026robustness}, open-set 3D mapping \cite{jatavallabhula2023conceptfusion}, and language-conditioned or decentralized aerial navigation \cite{chahine2026flex,chahine2025decentralized}. Related studies also evaluate multimodal language models as driving world models \cite{shiva2025probing} and combine pretrained video-language and language models for training-free video summarization \cite{barbara2025prompts}. Although these systems operate in different application domains, they share our interest in representations that transfer, adapt efficiently, and remain useful under distribution shift.

\section{Conclusions}
This work begins from a limitation of standard NTK pipelines for neural fields: they adapt only via linearized weight updates around the current parameters, and they do not accumulate reusable structure that transfers across tasks. We make this concrete by analyzing three methods, all introduced in this work.
\emph{NTK-KIP} learns, for each target scene or image, a compact “support set’’ of spatial coordinates (and optionally pixel values) so that finite-width NTK kernel regression can reconstruct or inpaint the full signal. This overcomes a core NTK weakness: fixed, non-adaptive sampling, and enables strong inpainting with extremely sparse observations. However, NTK-KIP must be solved from scratch for every new instance, which limits responsiveness at test time and prevents true meta-learning.
\emph{MetaQuill} addresses that second point directly: it meta-learns a shared initialization $\theta_S$ across many tasks, and then adapts a new task by optimizing only a tiny task-specific $\Delta\theta$, rather than retraining the whole network. This turns a neural field into a fast, reusable prior with learned transferable features: we get sub-second adaptation and a lightweight representation of task identity. But because MetaQuill’s adaptation is effectively tangent-space (NTK-like) around $\theta_S$, it remains locally linear. In heavily occluded or masked regions, that linearity caps expressiveness and hurts hole filling.
\emph{MetaQuill\mbox{-}KIP} is our answer to both limitations at once. It first uses a KIP-style non-linear warm start that captures high-level structure and is not restricted to a purely linear NTK response, and then continues with MetaQuill’s efficient $\Delta\theta$ refinement around the learned $\theta_S$. This provides a unified mechanism for non-linearity, feature learning, and fast adaptation: non-linear expressivity via the KIP warm start, learned transferable features via the shared initialization $\theta_S$, and fast per-task refinement by updating only $\Delta\theta$ rather than the full network. In practice, this yields high-PSNR reconstructions and semantically plausible inpainting from extreme sparsity (down to $\sim$1\% observed pixels) in only tens to hundreds of lightweight refinement steps, just a few seconds of wall-clock, while strong diffusion baselines such as Stable Diffusion with per-image finetuning or StrDiffusion require tens of seconds of sampling/adaptation to reach comparable fidelity.
In summary, NTK-KIP solves NTK’s lack of non-linear expressivity but is instance-specific and slow; MetaQuill solves fast adaptation and reusability but is locally linear; MetaQuill-KIP merges both. The result is an NTK-driven neural field pipeline that is (a) non-linear where it matters, (b) meta-learned and reusable across tasks, and (c) fast enough for practical few-shot reconstruction and inpainting.

\noindent\textbf{Future work. }
We view MetaQuill-KIP as a first step toward a unified NTK framework that is both non-linear and explicitly meta-learned. NTK-KIP supplies distilled, task-specific non-linear support sets; MetaQuill supplies a fast, reusable meta-initialization. Their combination suggests an NTK pipeline with learnable features, non-linear expressivity, and practical wall-clock efficiency. We hope this direction motivates further work on NTK methods that behave less like fixed kernels and more like adaptable representation learners.

\bibliographystyle{IEEEtran}
\bibliography{main}
\newpage

\vspace{-40pt}
\begin{IEEEbiography}[{\includegraphics[width=1in,height=1.25in,clip,keepaspectratio]{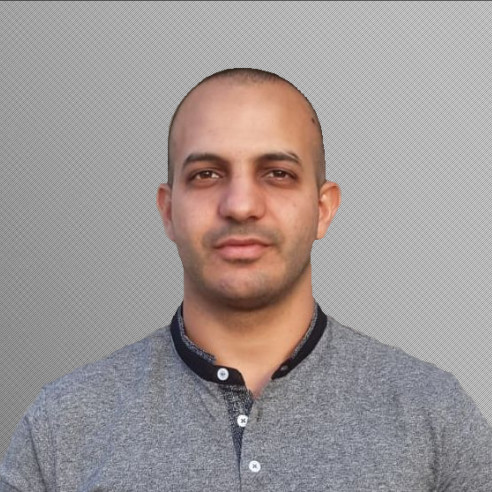}}]{Amir Mallak}
is a Ph.D. researcher at the University of Haifa. He holds a B.Sc. in Electrical and Computer Engineering from Ben Gurion University, and an M.Sc. in Computer Science from the University of Haifa. His research focuses on generalizable, efficient, and reliable AI systems, spanning machine learning, computer vision and spatial intelligence, robotics, multimodal foundation models, algorithms, and machine learning theory.
\end{IEEEbiography}

\vspace{-40pt}
\begin{IEEEbiography}[{\includegraphics[width=1in,height=1.25in,clip,keepaspectratio]{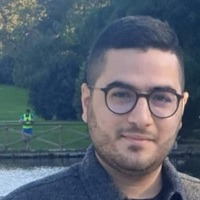}}]{Alaa Maalouf}
is the Neubauer Assistant Professor of Computer Science at the University of Haifa. He is also a Research Affiliate at MIT CSAIL, an Associate at Harvard SEAS, and a Principal Investigator with Project CETI. Before that, he was a postdoctoral researcher at MIT CSAIL and received his Ph.D. from the University of Haifa. He received the Maof Scholarship for Excellent Young Researchers from the Council for Higher Education, a Neubauer Family Foundation Fellowship for excellent young faculty, and the 2019 NeurIPS Outstanding Paper Award Honorable Mention. Maalouf’s research lies at the intersection of machine learning, computer vision, robotics, and large-scale data.
\end{IEEEbiography}

\vspace{-40pt}
\begin{IEEEbiography}[{\includegraphics[width=1in,height=1.25in,clip,keepaspectratio]{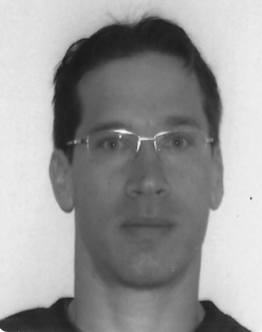}}]{Lior Wolf}
is the CEO at Mentee Robotics and a full professor in the School of Computer Science at Tel-Aviv University, Israel. He conducted postdoctoral research at prof. Poggio's lab at the Massachusetts Institute of Technology and received his PhD degree from the Hebrew University, under the supervision of Prof. Shashua. He is an ERC grantee and has won the ICCV 2001 and ICCV 2019 honorable mention, and the best paper awards at ECCV 2000 and ICANN 2016. His research focuses on computer vision and deep learning.
\end{IEEEbiography}

\vspace{-40pt}
\begin{IEEEbiography}[{\includegraphics[width=1in,height=1.25in,clip,keepaspectratio]{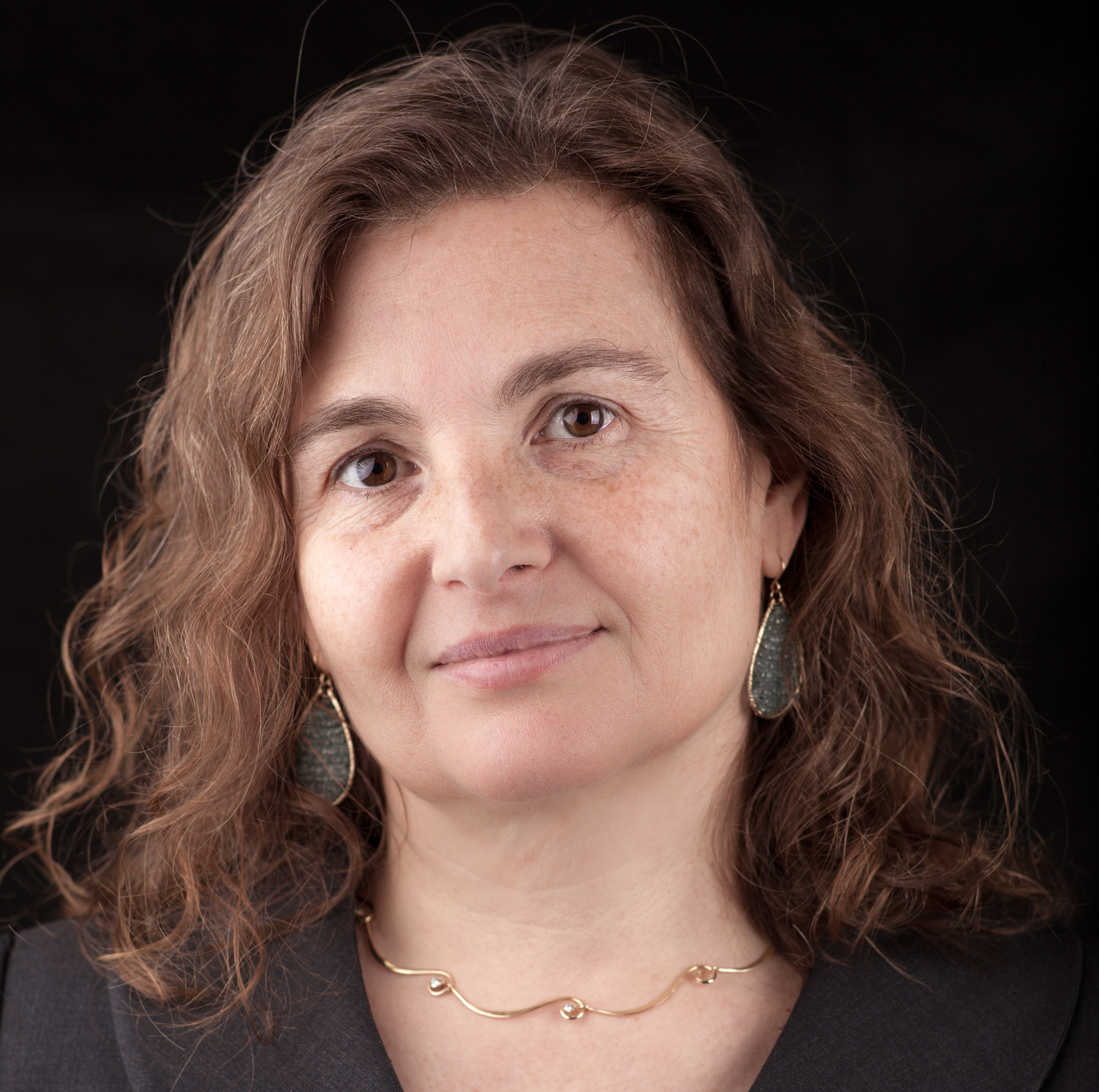}}]{Daniela Rus}
is the Andrew (1956) and Erna Viterbi Professor of Electrical Engineering and Computer Science and Director of the Computer Science and Artificial Intelligence Laboratory (CSAIL) at MIT. Rus’s research interests are in robotics, mobile computing, and data science. Rus is a Class of 2002 MacArthur Fellow, a fellow of ACM, AAAI and IEEE, and a member of the National Academy of Engineering, and the American Academy for Arts and Science. She earned her PhD in Computer Science from Cornell University.
\end{IEEEbiography}

\vspace{-40pt}
\begin{IEEEbiography}[{\includegraphics[width=1in,height=1.25in,clip,keepaspectratio]{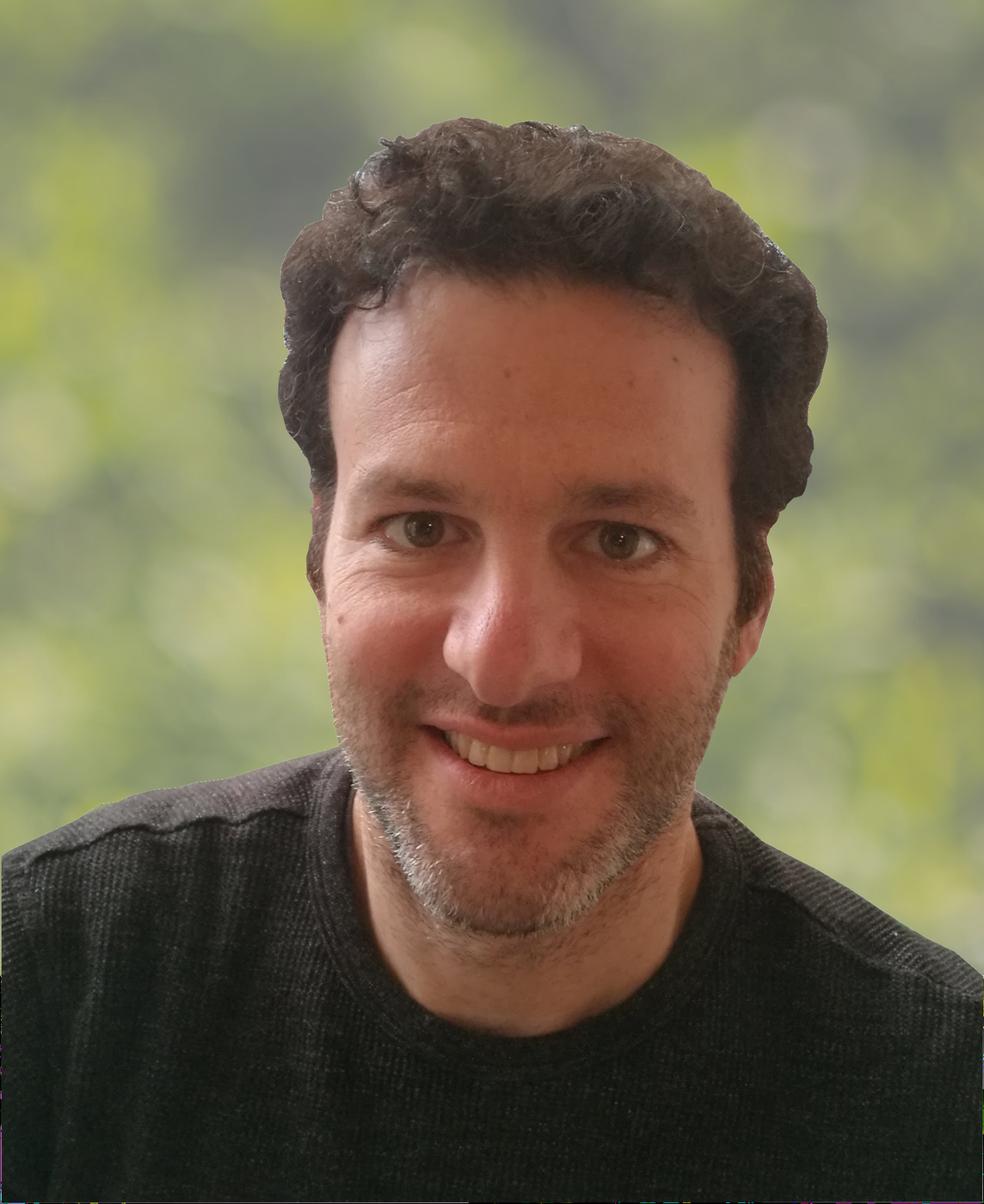}}]{Dan Rosenbaum}
is a Senior Lecturer in the Department of Computer Science at the University of Haifa. He is working on machine learning and computer vision, and specifically on 3D scene understanding and generative approaches that model vision as an inverse problem. Before joining the University of Haifa Dan completed his PhD in 2016 at the Hebrew University of Jerusalem, advised by Prof. Yair Weiss, and then worked as a Research Scientist at DeepMind in London between 2016 and 2021.
\end{IEEEbiography}

\newpage

\appendices

\section{List of Acronyms and Abbreviations}
\label{app:acronyms}
For readability, Table~\ref{tab:acronym_list} consolidates the acronyms and abbreviations used throughout the main manuscript and appendix.

\begin{table*}[t]
\centering
\small
\setlength{\tabcolsep}{4pt}
\renewcommand{\arraystretch}{1.1}
\begin{tabular}{p{0.15\textwidth} p{0.31\textwidth} p{0.15\textwidth} p{0.31\textwidth}}
\hline
\textbf{Acronym} & \textbf{Meaning} & \textbf{Acronym} & \textbf{Meaning} \\
\hline
NTK & Neural Tangent Kernel &
NF & Neural Field \\
INR & Implicit Neural Representation &
NN & Neural Network \\
MC & Matrix completion &
MLP & Multi Layer Perceptron \\
KRR & Kernel Ridge Regression &
KIP & Kernel Inducing Points \\
NTK-KIP & Our NTK based KIP method for neural fields &
MetaQuill & Meta learned initialization and task adaptive tangent space refinement method \\
MetaQuill-KIP & Combined method that uses KIP style distilled support and MetaQuill style adaptation &
GT & Ground Truth \\
PE & Positional Encoding &
RPE & Raw Positional Encoding \\
FPE & Fourier Positional Encoding &
GD & Gradient Descent \\
MSE & Mean Squared Error &
PSNR & Peak Signal to Noise Ratio \\
RGB & Red Green Blue &
DM & Diffusion Model \\
DM\_W & DM trained on NTK coefficient representations (W) rather than directly on GT images &
DDPM & Denoising Diffusion Probabilistic Model \\
GSDM & Global Structure-guided Diffusion Model &
StrDiffusion & Structure-guided Diffusion Model \\
SD & Stable Diffusion &
MNIST & Modified National Institute of Standards and Technology dataset \\
CIFAR-10 & Canadian Institute For Advanced Research-10 class dataset &
CelebA & Large-scale Celebrity faces Attributes dataset \\
QML & Quantum Machine Learning &
SOTA & State of the Art \\
\hline
\end{tabular}
\caption{Consolidated list of acronyms and abbreviations used in the main manuscript and appendix. We include common method names, kernel terms, optimization terms, and datasets to improve readability.}
\label{tab:acronym_list}
\end{table*}

\section{Determining the frequency bands k value} \label{Ablation}
\begin{figure*}
    \centering
        \includegraphics[width=0.7\textwidth]{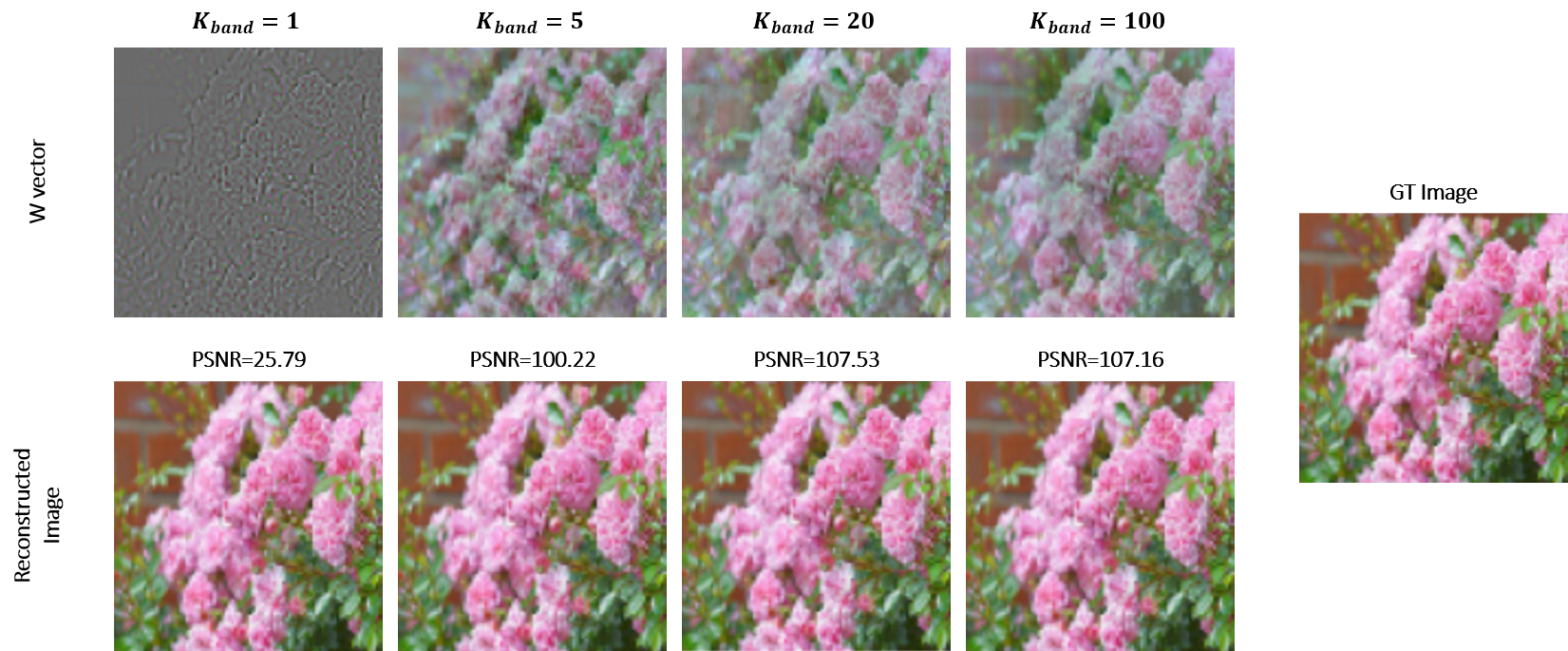}
        
        \caption{Optimal FPE Band according to the $\mathit{W}$ vector}
    \label{Optimal_K_Band}
\end{figure*}

In our experiments, we use FPE with frequency bands of $\mathit{k} = 20$ for the NF architecture’s input positional encoding. This value was determined from an experiment to identify the optimal FPE band. An illustration of the effect of $\mathit{k}$ is shown in Fig.~\ref{Optimal_K_Band}.

At $\mathit{k} = 20$, the $\mathit{w}$ vector achieves the highest accuracy with sufficient detail ($PSNR \approx 105.35$), closely matching the target image’s representation through a near-linear transformation. To further illustrate the optimal Fourier frequency $\mathit{k}_{band}$ selection in terms of $PSNR$ accuracy and feature efficiency, we present a graph of Fourier $\mathit{k}_{bands}$ versus $PSNR$ (for the reconstructed image), with a cubic interpolation to reveal the continuous trend.

\begin{figure}
    \centering
\includegraphics[width=0.48\textwidth]{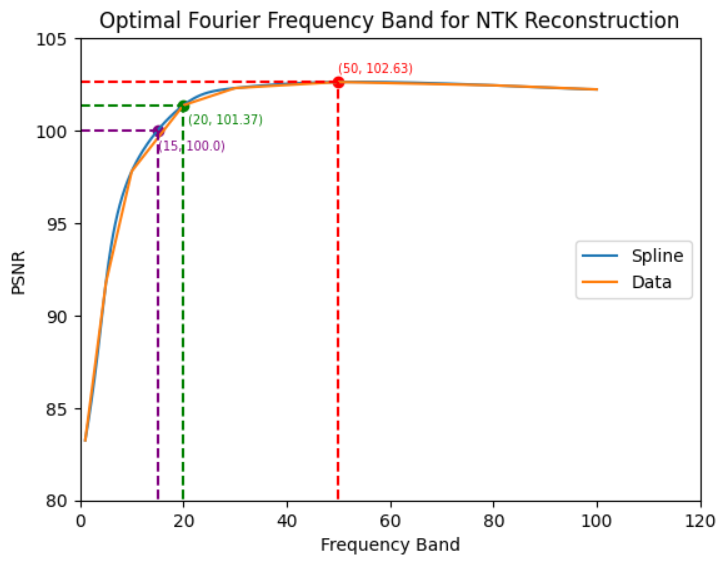}
        
        \caption{Optimal Fourier Frequency $\mathit{K}_{band}$ with Elbow Value}
    \label{Elbow_K_Band_Graph}
    \end{figure}

In FPE, each positional encoding (PE) 2D coordinate is transformed into $d + 2 \cdot d \cdot k$ Fourier features, where $d$ is the PE dimensionality, and $k$ is the number of Fourier bands (e.g., $d=2$, $k=20$). Thus, the number of $\mathit{k}_{bands}$ impacts the Fourier feature space by approximately $\sim4 \times \mathit{k}_{bands}$. For instance, increasing $\mathit{k}_{bands}$ by 10 adds 40 positional input features. The optimal $\mathit{k}_{bands}$ yields an elbow point in the graph with minimal bands. We calculate this elbow value across Fourier frequencies and include it in the graph, as shown in Fig.~\ref{Elbow_K_Band_Graph}. 

We can notice that the maximum $PSNR = 102.63$ is at $\mathit{k}_{band} = 50$. The Elbow and the optimal chosen Fourier Frequency $\mathit{k}_{band}$ is $\mathit{k}_{band} = 15$.
In this paper though, we ended up choosing $\mathit{k}_{band} = 20$ for the experiments. It yields slightly higher $\mathit{PSNR}$ and has a bit more feature representation.

\section{NTK Kernel Illustration}
\begin{figure}
    \centering
        \includegraphics[width=0.48\textwidth]{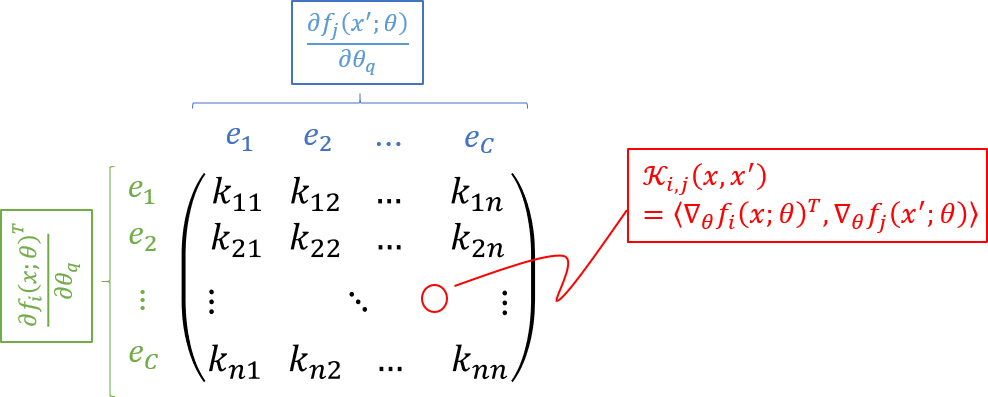}
        \caption{Neural Tangent Kernel Matrix Illustration}
    \label{NTK_Matrix_Illustrarion}
\end{figure}

To better understand the NTK kernel representation, we delve into its core and present theoretical findings and explanations.
From equation~\ref{NN_Evolution_Kernel}, if to consider a specific cell \([(i, j)\, |\, 1 \leq i,j \leq C]\) in the NTK,
\begin{align*}
\mathcal{K}_{(\mathit{i}, \mathit{j})}(\mathit{x}, \mathit{x}^{'} ;\, \theta) &= \nabla{_\theta}\mathit{f}_{\mathit{i}}(\mathit{x}\, ;\, \theta)^T \ \cdot \ \nabla{_\theta}\mathit{f}_{\mathit{j}}(\mathit{x}^{'} ;\, \theta) = \\ &= \sum_{\mathit{q}=\mathit{1}}^{\mathit{Q}} \frac{\partial \mathit{f}_{\mathit{i}}(\mathit{x}\, ;\, \theta)^\mathit{T}}{\partial \theta_{\mathit{q}}} \cdot \frac{\partial \mathit{f}_{\mathit{j}}(\mathit{x}^{'} ;\, \theta)}{\partial \theta_{\mathit{q}}},
\end{align*}

where $\mathit{Q}$ is the total number of parameters in the NN $\mathit{f}$. See Fig. \ref{NTK_Matrix_Illustrarion} for illustration.

Calculating correlations through inner products of neural network gradients compresses meaningful information, neglecting dynamic architectural effects and resulting in a low-dimensional projection rather than a robust representation. While NTK offers advantages in training flexibility, it lacks effective feature extraction and true learning capability. To explore these limitations, we propose the concept of "NTK Representational Linearity", which addresses the fundamental constraints in NTK’s approach. This analysis contributes to understanding and mitigating limitations in the NTK theorem, highlighting its impact on representation learning.

\section{Functa via NTK} \label{functa_NTK_appendix}

It is common practice in deep learning to represent a measurement of the world on a discrete grid, e.g. a 2D grid of pixels. However, the underlying signal represented by these measurements is often continuous, e.g. the scene depicted in an image, or the transformed Fourier of an audio signal, etc.

A powerful frequency-continuous alternative is then to represent these measurements using an NF, a neural function trained to output the appropriate measurement value for any input spatial location 
$NF: l \rightarrow v$ (a location to value transformation).

In Functa paper \cite{dupont2022data}, the authors take this idea to a higher approach (Regarding Meta-Learning): what would it take to perform deep learning on these functions (NF) instead, treating them as data?
In this context it's referred to the data as Functa.

In this experiment, in order to test NTK's representation and efficiency, we decided to refer to the Functa paper as a comparison. In this paper, the authors transform each data point in the dataset to a function (NF Model) and manage to distillate this function representation (hence the dataset) up to 
$\sim 0.5\%$ from the original dataset size with great accuracy. Meaning, they managed to distillate the data by more than two orders of magnitude while maintaining the interesting features to be able to reconstruct the distilled NF.

This scale of feature learning has radical importance in representation learning, and one of the methods to be able to determine the quality and ability of the conducted method to adapt and actual include and learn new features.

In order to test this, we conducted an NTK for representation learning via Functa experiment. Our downstream chosen task is MC, and more specifically, Inpainting.

 For illustration of Functa, see Fig. \ref{Functa_NTK_Illustration}.

In NTK Functa, the process in slightly different. Instead of training our generative model (Diffusion Model - DM) on the NFs new dataset. We train the generative model (DM) on an infinite-width NF. And the way to make this feasible is via NTK kernel trick, Fig. \ref{Functa_NTK_Illustration}. Thus, calculating $\mathcal{K}_{train}$ and $\mathcal{K}_{test}$ NTK Kernels for all the dataset (shared for all the NFs), and calculating and training the DM on the $W$ vectors (derived from the kernel and labels, Eq. \ref{W_Vector_MC}).

As stated above, after calculating the NTK kernels, we'll calculate the $\mathcal{W}$ vectors (as explained earlier via $\mathcal{K}_{train}$ , Eq. \ref{W_Vector_MC}) as such: $\mathcal{W} = \mathcal{K}_{\mathit{t}}(\mathit{x}, \mathit{x}^{'} ; \theta)^\mathit{-1} \mathcal{Y}\ |\ \mathcal{K}: \mathbb{R}^d \ \times\  \mathbb{R}^d \ \rightarrow{} \ \mathbb{R}^{\mathit{C} \times \mathit{C}}$ (such that we're taking the full PE vector (This means, all of its elements representing all of the matrix coordinates), And the $\mathcal{Y}$ vector as whole. Namely, containing all the labels (observed and non-observed ones)). And replacing the learned NFs with these $\mathcal{W}$ vectors.

Note,
Though this is an Inpainting downstream task, we're taking the whole image pixels (and coordinates) as observed at this stage (creating the $\mathcal{W}$ vectors), due to the nature of Functa application, where the INR model needs first to learn the distribution of the dataset before being tested and generate novel data points.

After completing this, as the case with classic Functa, we'd actually achieve a whole new representation of the dataset. Replacing each dataset sample with a $\mathcal{W}$ vector. Turning the dataset of samples into a dataset of $\mathcal{W}s$.

After calculating the $\mathcal{W}s$ representations, we'll train a generative-diffusion model on those $\mathcal{W}s$ to be able to generate novel $\mathcal{W}$ vectors. Thus, generating new INRs which can handle other real-world tasks (e.g. Super-resolution, continuous-unseen scenery samples, etc.).

By doing so, we're training the DM (diffusion model) to fit the probability distribution of such $\mathcal{W}s$. Letting it learn how the probability of such $\mathcal{W}s$ representational vectors, for this specific trained task, supposed to look like.

At inference,
We'll generate a novel $\mathcal{\hat{W}}$ vector via our DM, and predict the missing pixels via the pre-calculated NTK ($\mathcal{K}_{test}$ to be specific) - see Eq. \ref{Y_Reconstruct_MC} (Yet, here, the observed pixels are the whole image. Thus, $\mathcal{K}_{test} = \mathcal{K}_{train}$). Formally,
\begin{enumerate}
    \item $\{\mathcal{W}_i\}_{i=0}^{N-1} = \mathcal{K}_{train}^{-1} \times \{\mathcal{Y}_i\}_{i=0}^{N-1} \mid N: |D|,\, D: Dataset$
    \item $Train\, DM_W\, on\, \{\mathcal{W}_i\}_{i=0}^{N-1}$
    \item $\hat{\mathcal{W}} = I[DM_W] \mid I: Inverse\, from\, trained\, DM_W$
    \item $\hat{\mathcal{Y}} = \mathcal{K}_{test} \hat{\mathcal{W}}$
\end{enumerate}

Configuring the Loss Function of the DM,
Each Diffusion model block consists of a Unet NN architecture with attention. We train the Unet model to identify and output the noise associated to its input image at each time step t ($t \in [0, T-1]$, where $T$ if the total number of steps in the diffusion process) accordingly, for various time steps.
Formally,

\begin{align*}
    \mathcal{L} = ||\epsilon - U_\theta(x_t, t)||_2^2
\end{align*}
S.T.
\begin{align*}
    &\epsilon \sim N(0, 1) \\&
    x_t = \sqrt{{\overline{\alpha_t}}} \cdot x_, +\, \sqrt{{1 - \overline{\alpha_t}}} \cdot \epsilon \\&
    U_\theta(x_t, t) = \hat{\epsilon}_{t-1}\, |\, \hat{\epsilon}_{t-1}: Predicted\, noise\, at\, step\, t-1
\end{align*}

The experiment was conducted on both MNIST and Celeb-A datasets. In each, we trained two DMs. First, directly on the image doamin, and second, on the $W$s dataset.
Example results of the calculated $\{\mathcal{W}\}_i \mid i \in \{0, ...,  N-1\}$, can be seen in Fig, \ref{W_vs_GT_CelebA_MNIST}.

The trained DM will generating a novel data distribution sample via its learnt Unet parameters. But what if we want to use it for Inpainting or Matrix Completion (MC) tasks?
In that case, we'll have an incomplete$\backslash$masked matrix (e.g. Image), and we would want the DM to output the complete hidden-pixels predicted image. Similarly for completing$\backslash$predicting the corrupted pixels$\backslash$values of a masked $\mathcal{W}$
vector.
Thus, given a corrupted image$\backslash$$\mathcal{W}$ vector, we pass it through the trained DM ($DM,\, DM_W$ accordingly) and use its observed pixels as guidance in every step $t \in [0, T-1]$ of the Diffusion process (For each $Unet_\theta$).
Formally,
\begin{enumerate}
\item $\hat{\epsilon}_{t-1} = U_\theta(x_t, t)$
\item $\hat{x}_0(x_t, t) = \frac{1}{\sqrt{\overline{\alpha_t}}}(x_t\, -\, \sqrt{1 - \overline{\alpha_t}} \cdot \hat{\epsilon}_t)$
\item $\mathcal{L} =  \sum_{obs}||\mathcal{Y}_{obs} - \hat{x}_0||_2^2\ |\, obs: Observed\, pixels$
\item $\nabla P(y|x) = -\frac{\partial{\mathcal{L}}}{\partial{x_t}}$
\\Sample $x_{t-1}$ from the new probability distribution,
\item $x_{t-1} \sim N(\mu_{t-1} + \triangle P(y|x), \sigma_{t-1})$
S.T.
\begin{itemize}
    \item $\mu_{t-1}(x_t, t) = \frac{1}{\sqrt{{\alpha_t}}}(x_t\, -\, \frac{1-\alpha_t}{\sqrt{1 - \overline{\alpha_t}}} \cdot \hat{\epsilon}_{t-1})$
    \item $\sigma_{t-1} = \sqrt{\overline{\beta}_t} \cdot z\ |\, z_{[T-1, 2]} \sim N(0, 1),\, z_1 = 0$
    \item $\overline{\beta}_t = \beta_{t-1} \cdot \frac{1-\hat{\alpha}_{t-2}}{1-\hat{\alpha}_t}$
    \item $\alpha_t = 1-\beta_t\ ;\ \overline{\alpha}_t = \prod_{i=1}^{T} \alpha_i\ ;\ \beta = [10^{-4}, 2 \cdot 10^{-2}]_T$
\end{itemize}

Thus, resulting in,\\
$x_{t-1} = \frac{1}{\sqrt{{\alpha_t}}}(x_t\, -\, \frac{1-\alpha_t}{\sqrt{1 - \overline{\alpha_t}}} \cdot \hat{\epsilon}_{t-1})\, +\, S \cdot \nabla P(y|x)\, +\, \sqrt{\overline{\beta}_t} \cdot z$
\end{enumerate}

In order to compare the representational performance of the NTK method, let's consider the comparison between the $\mathcal{W}$ vector representation and the original image. For this task, we trained two Diffusion Models, one on the GT images and the other on the created NTK $\mathcal{W}$ vectors.
The results in Fig. \ref{Guided_DM_&_WDM} show the end downstream task of Inpainting and MC, inverse guided-sample of the DM and the reconstructed image.

We can notice the representational difference between training and performing directly in the image space and when the NTK space is considered as a representational mean. Many conclusions can be drawn from these results.

In the GT images space, the results are much smoother and more refined. Also, in the bottom-left ($2$ mnist digit) GT image example, we can see that in the NTK representational space ($W$ space), the $W$ vector is noise sensitive (as stated and shown in Section \ref{NTK_Robustness}) and hence affecting the performance of the Diffusion Model. The process of de-noising in the diffusion trajectory is rather complex and an additional noise in the learned representations could only make it more difficult to inverse. Furthermore, after the diffusion model has generated a new sample (as in the bottom-left ($2$ digit) example), if this sample is noisy (which as stated before, the $W$ vector representation doesn't make this easier), the reconstruction image from this $W$ vector will be radically blurry (as can be seen in the reconstructed image part). Again, due to $W$ vector noise sensitivity.

Another conclusion which can be made, by examining the bottom-right result. We can see that when the diffusion model was trained on the original GT images (image space), the resulted guidance was much easier, without much of a hyperparameter effort, and with great accuracy in each result. Yet, when it was trained on the $W$ vectors (NTK space), the guided regularization hyperparameter wasn't consistent for all samples. For some the DM could conduct a good inverse guidance, but for other samples, the guidance was disrupted and the inverse wasn't an accurate one. Leading to the observation that this NTK representation doesn't bring any extra interested features or information in its core, but rather a degraded one.

We can notice that even in the upper results, when the DM could reconstruct the guided novel sample via the NTK method, the reconstruction suffers from noise.

If to consider Inference run-time wise efficiency,
Let's compare the representational size and computation in both methods (Functa vs NTK), considering the basic NF which was used in both algorithms.
In Functa, the basic NF which was used in the                      paper, as shown above (without the FPE), consists of $5 \times 256$ hidden layers and an RGB output ($1\times 3$). Thus, if to calculate the Inference run-time complexity required for the algorithm (Neglecting the bias term),
\begin{align*}
    &\underbrace{\overbrace{(h \cdot w) \times d}^{PE} \cdot l}_{1^{st}\, hidden\, layer}\, +\, \underbrace{(l \times l \times N_l)}_{mid\, hidden\, layers}\, +\, \underbrace{(l \times c)}_{final\, output\, layer} = \\& \underbrace{=}_{in\, our\, case} (32 \cdot 32) \times 2 \cdot 256\, +\, (256 \times 256 \times 4)\, +\, (256 \times 3) = \\&
    \cong 8 \times 10^5 = \textbf{0.8M}
\end{align*}

In the NTK method, we'll mention the whole process in Inference. Yet, at the end, neglect the FPE and the kernel mapping calculation (it differs depending on model architecture, input data, algorithm, etc.). The process will be to calculate the FPE, using it to perform Kernel calculation (Calculating both $K_{train}$ and $K_{test}$), using $K_{train}$ to calculate $W$, and then calculating $\hat{Y}$ via $K_{test}$ (Considering the reconstruction complexity). We get,

\begin{align*}
    &\underbrace{\overbrace{(h \cdot w) \times d}^{PE} \times (d\, +\, 2 \cdot d \cdot \overbrace{k}^{Fourier\, bands})}_{FPE}\, +\, \underbrace{(K_{train}^{-1} \times Y)}_{W_{vector}}\, +\, \\& \underbrace{K_{test} \times W}_{Reconstruct\, \hat{Y}} = \\&
    = \underbrace{\overbrace{(h \cdot w) \times d}^{PE} \times (d\, +\, 2 \cdot d \cdot \overbrace{k}^{Fourier\, bands})}_{FPE} \\& +\, \underbrace{(\overbrace{\mathcal{O}(o^3)}^{K_{train}\, inverse\, \mathcal{O}(n^3)} \times \overbrace{(o \cdot o)}^{K_{train}\, structure} \times \overbrace{(o \times c)}^{Y\, structure})}_{W_{vector}} \\& +\, \underbrace{[\overbrace{[(h \cdot w) \times o]}^{K_{test}\, structure} \times \overbrace{(o \times c)}^{W_{vector}}]}_{Reconstruct\, \hat{Y}} \\& \underbrace{=}_{in\, our\, case} \underbrace{\overbrace{(32 \cdot 32) \times 2}^{PE} \times (2\, +\, 2 \cdot 2 \cdot \overbrace{20}^{Fourier\, bands})}_{FPE} \\& +\, \underbrace{(\overbrace{\mathcal{O}((32 \cdot 32)^3)}^{K_{train}\, inverse\, \mathcal{O}(n^3)} \times \overbrace{((32 \cdot 32) \times (32 \cdot 32))}^{K_{train}\, structure} \times \overbrace{((32 \cdot 32) \times c)}^{Y\, structure})}_{W_{vector}} \\& +\, \underbrace{[\overbrace{[(32 \cdot 32) \times (32 \cdot 32)]}^{K_{test}\, structure} \times \overbrace{((32 \cdot 32) \times c)}^{W_{vector}}]}_{Reconstruct\, \hat{Y}} = (*)
    \end{align*}

In the above calculation, we've calculated the Inference complexity term using the variable $c$. Here, $c$ is an indication for the last dimension of the matrix $Y$ (Which is used for the Training data phase (Calculating $W$)). While $Y$ is determined according to the downstream task which we're solving.
e.g. In case we're dealing with medical data for instance, and the $Y$ matrix $c$ dimension of a size $\sim \mathcal{O}(n)\, |\, n = h \cdot w$. Then,

\begin{align*}
    (*) &\overbrace{\geq}^{\triangle_3} \overbrace{10^5}^{FPE}\, +\, \overbrace{[10^9 + 10^9]}^{W_{vector}}\, +\, \overbrace{10^9}^{Reconstruc\, \hat{Y}} \\& \overbrace{\geq}^{\triangle_4} 2 \times 10^9 = \textbf{2G}
\end{align*}
And in case $Y$ matrix is an RGB image. Thus, $c = 3$. Then,
\begin{align*}
    (*) &\overbrace{\geq}^{\triangle_5} \overbrace{10^5}^{FPE}\, +\, \overbrace{[10^9 + 10^6 \cdot 3]}^{W_{vector}}\, +\, \overbrace{[10^6 \cdot 3]}^{Reconstruc\, \hat{Y}} \\& \overbrace{\geq}^{\triangle_4} 2 \cdot [10^6 \cdot 3] = 2 \cdot 3 \times 10^6 = \textbf{6M}
\end{align*}
S.T.
\begin{align*}
    \triangle_3: General\, Matrix\, Multiplication - \mathcal{O}(n^3)
\end{align*}
\begin{align*}
    \triangle_4: &Neglecting\, FPE\, and\, Kernel\, Inverse\, \\& 
(Both\, computed\, only\, once)
\end{align*}
\begin{align*}
    \triangle_5: Matrix,\, Vector\, Multiplication - \mathcal{O}(n^2)
\end{align*}

Explaining $\triangle_4$,
\begin{itemize}
    \item We neglected the inverse calculation of $FPE$ and $K_{train}$. This is due to the fact that both could be calculated only once at the beginning of the experiment and then be used in all the individual NFs calculations (The NTK Kernel is shared among the dataset samples).  This due to the fact that the $FPE$ are fixed to all positional inputs, and $K_{train}$ is fixed as well for all the corresponding $W$s (In an NF model, the $PE$ input is fixed for all images - when they're at the same size, and the model's architecture is similar to all Multi-NFs. Thus, the NTK Kernel calculation is fixed since its inputs are the PEs data samples and the NF architecture).
    \item We neglected the Kernel mapping function time complexity. This is due to the fact that again, as stated in the previous point, the Kernel matrices (in our experiment) can be calculated only once at the beginning and then be used and shared among all the dataset samples. Another reason which this could help, the kernel mapping matrix differs depending on model architecture, input data, calculation algorithm, etc.
\end{itemize}

We can see the radical difference in between the two representational methods in terms of Inference efficiency (One or even two order of magnitude difference to the favor of the NF). The $W$ vector's representation size $w = k^{-1} \times y \rightarrow (h \cdot w) \times c$ is similar to the target image's size, also $(h \cdot w) \times c$. And the results, yet, falls short.

The NTK method suffers from inefficiency when compared to other methods (as in the Functa paper, and even the direct raw data input space) and this inefficiency is shown in three main constellations:
\begin{enumerate}
\item \textbf{Representational structure efficiency.}
\item \textbf{Representational effectiveness.}
    \item \textbf{Time-complexity and computational flops efficiency.}
\\\end{enumerate}

To conclude,
NTK is unsuitable for practical representation tasks due to weaknesses in structure, representation, and runtime, posing limitations for applications like model scaling and edge AI (energy-efficient systems). We observe three main weaknesses in the NTK method:
\begin{enumerate}
    \item \textbf{Representational structure efficiency:} The NTK matrix $W$ has a representational size of $\mathbb{R}^{(h \cdot w) \times c}$, matching the ground truth (GT) image size but lacking efficient structure.
    \item \textbf{Representational effectiveness:} Although $W$ matches the image size, it fails to achieve effective representation in terms of performance results.
    \item \textbf{Time-complexity:} The NTK method exhibits high time complexity and computational cost, being at least one order of magnitude less efficient in runtime.
\end{enumerate}

\section{Novel Approaches of Representation Learning via NTK}

\subsection{Representation Learning via NTK Novel Algorithms} \label{NTK_Novel_Algorithms}
In this paper, we outline the previous limitations, and propose a transformative path forward in applying NTK to representation learning, proposing the integration of our two novel algorithms. Each algorithm independently addresses a unique, previously unsolved challenge in NTK, and their combined potential could fundamentally overcome existing limitations, paving the way for a significant leap in the field.

The results of our novel algorithms,
\begin{itemize}
    \item Fixing Representational Linearity - NTK KIP.
    \item NTK Algorithm which suggests a new Efficient meta-learning approach.
\end{itemize}

\subsubsection{\textbf{NTK-KIP}} \label{NF_KIP_Algorithm_Appendix}
In an effort to solve the previous discussed issues,                      linearity, efficiency, and NTK's kernel poor representation (as appears in the hollow experiment), we came up with the idea of NTK KIP.

We've noticed that the $w$ vector suffers from the linearity problem (Section \ref{NTK_Natural_Limitations}, \textit{NTK representational linearity}) with respect to the target image, As we recall $W = K_{train} Y_{train} \mid Y_{train}: GT\, Observed\, Labels$.

Hence, if to handle this, we need for $K_{train}$ (NTK Kernel) to be $Y_{GT}$ dependent, which will lead for $W$ to become $Y$ dependent as well. And the way to assure this is via the input coordinates. If we can update the coordinates (or at least some of them) according to $Y$ then our PE will become $labels$ dependent and hence $K$ (which uses our NF input's PE) will also be $labels$ dependent, leading that $W$ will be non-linear with respect to $Y$. Formally, let $N$ be the total number of images in our dataset. Given image $i \in [1,\, ...,\, N]$ of size $(h,\, w,\, c)$, let $P = h \cdot w$ be the total number of pixels in $i$, $p \in [1,\, ...,\, P]$ be a pixel in $i$, $x \in \mathbb{R}^{N \times 2}$ be the PE vector, $y \in \mathbb{R}^{(P \times c)}$ be the image labels (values). We then define $x_i^p \in \mathbb{R}^2$ to be the PE coordinate which corresponds to pixel $p$ in image $i$, and $y_i^p$ to be the value (label) of pixel $p$ in image $i$. In the Raw PE (RPE) example, we optimize $x_i^p$ and $y_i^p$ to minimize the reconstruction error for the full image:
\begin{align*}
    \arg \min_{x_i^p, y_i^p}
    \left\| \mathcal{K}(x, x_i^p) \; \mathcal{K}(x_i^p, x_i^p)^{-1} \; y_i^p   - y \right\|
\end{align*}

This representation is therefore no longer a simple linear mapping of the signal, and can potentially lead to extracting more meaningful and useful information from the image.

This algorithm will also improve NTK's kernel representation (as can be seen in the $hollow$ experiment). The PEs will be updated via Gradient Descent according to a reconstruction loss objective of the hypothesis $\hat{Y}$ model output. This will lead to not only better feature representation, but also data $(Y_{GT})$ distillation. When the distilled information will be "held" in the PEs, and they will be updated is such a manner that the NTK Kernel can be built as a better representation concept (from the leaned objective), and will be used as an actual mean for learning (via Kernel gradient backpropagation) and updating features accordingly.

Previous research was conducted in similar areas, where NTK's kernel was used also as a mean for inducing points, yet, in a classification task (an image as the input, and the output is the class hypothesis) \cite{nguyen2020dataset}.
In our case, the input is the positional coordinates and the output is an RGB value. In this experiment we aim to distillate our dataset (positional coordinates) as much as possible (At least one order of magnitude, with the goal for two orders).

We tested this method when the trained PEs were once the raw coordinates, and again as the FPEs (Fourier Positional Encodings). In both experiments we enable labels learning (yields better results).
The results of both experiments can be seen in Fig. \ref{NF_KIP_PEs_experiment}.

In this experiment, for the Raw PEs KIP, one can see that a steep degradation in the performance appears once down scaling the distillation percentage of the observed pixels in the GT image. We can gain relatively good results at $80\% \sim 50\%$ observed pixels ($PSNR \cong 36.74 \sim 23.72$).
Regarding the FPEs KIP experiment, we can notice a much better improvement regarding feature learning, Kernel representation, and distilled Inpainting reconstruction.
Though in the FPE we're using $k$ bands, thus expending the input coordinates features representation, it's still considering only the mentioned distilled percentage of the $labels$. The learning and image representation will take place on extra features, yet, the distilled representational coordinates will be maintained.

The $k$ band which was used in the FF transformation is 
$k=20$ (as shown in Fig. \ref{Elbow_K_Band_Graph}), which means, as shown in Appendix \ref{Ablation}, that each PE 2D coordinate is transformed into a $d\, +\, 2 \cdot d \cdot k\, |_{d=2,\, k=20} = 82$ Fourier Features. Hence, the total number of features is $Obs{\%} \cdot F_{features} = x\% \cdot 82$. If to take a $5\%$ distilled dataset for instance, the No. of FF will then be $410\%$ of the whole dataset.
This is still a limitation of this algorithm. For this is higher than the Raw PEs percentage and about $\times 4$ larger representation than the raw image itself (image representation number of pixels). Yet, the $labels$ percentage of the target image still holds, and is still the original distilled percentage.
Another thing to be noticed, after applying the Fourier encodings and learning a distilled representation of its transformation, the resulted representation will not be interpreted and might not represent a 2D spatial target position as the RPEs.
These findings lead to an understanding that a significant amount of trained and learned features are required in order to truly grasp and absorb the distilled latent information via the NTK method. Hence, driving the need for a true learning-capable NTK algorithm (See Section \ref{MetaQuill_Algorithm}).

To conclude the above in terms of Raw PE vs FPE with respect to Distillation percentage, see Table \ref{PE_(Raw and FPE)_vs_Distillation}.

In order to be able to have a better perspective regarding the high dimensional model hyperparameters and to project their main harmonic to a hyperplane, we've selected the Fourier $k_{bands}$ and Distillation percentage (FPE and Target image) parameters to examine their effect on the Final model's PSNR. Results in Fig. \ref{fig:kip_surface_fourier}.

We conducted the experiment examining a grid of $25$ samples in total from each param ($5$ values per parameter - $k_{bands}$ and $Distillation$). In order to fit and approximate in relatively high accuracy the probability hidden function of this distilled space, we used a radial basis function. And more specific, Quintic function.
We also added the projection of the approximated space function on each 2D plane.

If we examine the plane projection on the $X-Z$ axes, we can see the same 2D graph shape as in the experiment with the Fourier $k_{bands}$ (See Appendix \ref{Ablation}, Fig. \ref{Elbow_K_Band_Graph}).

In this sweep grid, PSNR is highest at low distillation and decreases as distillation becomes more aggressive, while increasing $K_{\mathrm{bands}}$ improves PSNR until saturation.
This is as expected. The distillation is at its lowest (distillation range: $95\% \sim 60\%$), $k_{bands}$ is at the maximum of the saturated 2D planar.

As an example for the FPE representation size, let's consider a transformation consists of $k_{bands}=30 \rightarrow FF=122$ (Per positional encoding), and at $50\%$ distillation. Then, the representational size,
\begin{align*}
    representation\_size &= distillation \cdot image\_size \cdot FF \\& = 50\% \cdot h \cdot w \cdot 122 = 61 \cdot h \cdot w \\& = 61 \times image\_size
\end{align*}

We were able to achieve $labels$ percentage of about $2$ orders of magnitude $(\sim 95\%)\, distillation$ lower than the GT image. Yet, the representation size (Total feature count) is still relatively large. Thus, concluding that the NTK method still falls short and relatively struggle with extracting and learning meaningful features.

To conclude,
\noindent \textbf{Neural Field Kernel Inducing Points (NTK-KIP).} In order to fix the problem of NTK's $\mathcal{W}$ Representational Linearity and also Representational Efficiency, we developed a novel algorithm NTK KIP (Kernel Inducing Points), which by updating the spacial coordinates input of the NF via NTK's Kernel derivative while minimizing an objective loss, not only creating a distilled input (And hence, label) prior data, and thus a more efficient NTK $\mathcal{W}$ representation. But also, a Non-Linear $\mathcal{W}$ representational vector,
$\mathcal{W} = \mathcal{K}_{\mathit{t}}(\mathit{x}, \mathit{x}^{'} ; \theta)^\mathit{-1} \mathcal{Y}\\$
s.t.
$\mathit{x}=\mathcal{G}(\mathcal{Y})\, |\, \mathcal{G}: non-linear\, mapping$\\
$\rightarrow \, \mathcal{K} = \mathcal{F}[\mathit{g}(\mathcal{Y})]\, |\, \mathcal{F}: kernel\, function$ \\
$\rightarrow \, \mathcal{W}:\, non-linear\, mapping\, of\, \mathcal{Y}$

With this, we managed to create via the Raw Positional Encoding (RPE) an input data distillation, and hence also $\mathcal{W}$ distillation, of $\mathit{20}$\% with above $\mathit{36}$ \textit{PSNR}, and $\mathit{50}$\% with above $\mathit{23}$ \textit{PSNR}. And via the Fourier Positional Encoding (FPE), $\mathit{80}$\% with above $\mathit{31}$ \textit{PSNR}, and $\mathit{95}$\% with above $\mathit{20}$ \textit{PSNR}.
We discovered that though this algorithm can create a more distilled and a non-linear representation, it suffers from the lack of learninng and features extraction which NTK's representational nature suffers from. Hence, we created the following algorithm (Section \ref{MetaQuill_Algorithm}) to solve the above NTK problem.

\subsubsection{\textbf{MetaQuill Algorithm}} \label{MetaQuill_Algorithm_Appendix}
We've discussed the lack and limitations of the NTK method as a representational concept. Some of which reasons had to do with the Infinite limit, and the lack of feature learning and the actual evolution of NN complex architectures over time.

In an effort to bridge these gaps, we came up with a novel idea (which to our knowledge has never been discovered) which combines Finite-width NTK, Model Feature learning, and a novel algorithm which achieves the benefits of MAML (Mode Agnostic meta-learning) without its inner loop. We also supply a theoretical analysis for our novel algorithm.

For this algorithm development, we consider the Functa (See Section \ref{NTK_Functa}) discussed downstream task (NF NTK's representation via DM). The main idea is to try and develop a new algorithm which takes into consideration the Feature Learning of an NF model (Thus, gaining learning capabilities for NTK), makes better usage of the learned data (Learning on only a small subset of the data. Tackling Efficiency), and builds a whole new concept which has the benefits of Meta-learning (Features Learning. e.g. MAML), yet in a more efficient manner (Specifically without MAML's inner loop).

Specifically, in the Functa DM (Diffusion Model) downstream task, we explained how, as inspired by the Functa paper, our dataset is being transferred into a dataset of functions (NFs), and then we train a DM on those NFs for various tasks and purposes (Generating new distributions for yet unseen tasks, such that it can be controlled to our own purposes and task dependent. e.g. super-resolution, matrix completion, etc.). Each function (NF) learns a single image, thus transforming the images dataset into an NFs dataset (Creating Multi-NF).

\paragraph{\textbf{Algorithm}}
In this Algorithm, we'll be using Finite-NTK rather than the Infinite theorem.

\noindent{\textbf{Finite-Width Kernel Learning}}
It has been observed that the performance of the infinite width NTK can be quite poor compared to finite width networks \cite{lee2020finite}. This is because finite-width networks can exhibit feature learning, whereas for infinite-width networks, the underlying feature map is fixed.
In the infinite width, the feature map $\phi(x)$ is given by the Tangent feature map $\varphi(x) = \nabla_{\theta}f_\theta(x)$ . As the width of $\theta$ goes to infinity. Because $\theta$ is infinite, the feature map is infinite, so we use the kernel trick to perform regression on this feature set.
Let $K(x,\, x') = \mathbb{E}_{\theta \sim P(\theta)} [\nabla_\theta f_\theta(x)^T \cdot \nabla_\theta f_\theta(x')]$ , as $w \rightarrow \infty$ , for initialization distribution $P(\theta)$.
It has been observed that while the infinite width NTK is an inaccurate model of network training, the NTK theory can still be used, but instead using a time and data-dependent finite NTK. Specifically, it has been observed that the training process of a NN $f(\theta)\, |\, \theta: NN\, parameters$, can be approximated by a first-order Taylor expansion around a working point $\theta_{\tau}$, S.T.
\begin{align*}
\theta_{\tau}\, |\, \tau \in \{T\},\, &\tau: infinitesimal\, No.\, epochs,\, \\& T: The\, training\, phase\, of\, NN
\end{align*}

First-order Taylor expansion,
\begin{align*}
    f_{\theta_{t +\, \tau }}(x) \approx f_{\theta_{\tau}}(x)\, +\, (\theta_{t\, + \tau} - \theta_{\tau})^T \cdot \nabla_{\theta_{\tau}}f_{\theta_{\tau}}(x)
\end{align*}

Namely, we train for $\tau$ epochs, then perform a first-order Taylor expansion. This is like training a linear model with a learned feature map $\nabla_{\theta_{\tau}}f_{\theta_{\tau}}(x)$ , with a resulting finite NTK $K_{\tau}(x, x') = \nabla_{\theta_{\tau}}f_{\theta_{\tau}}(x)^T \cdot \nabla_{\theta_{\tau}}f_{\theta_{\tau}}(x')$ .
The quantity $f_{\theta_{\tau}}(x)\, +\, (\theta_{t\, + \tau} - \theta_{\tau})^T \cdot \nabla_{\theta_{\tau}}f_{\theta_{\tau}}(x)$ can be computed exactly using forward-mode auto-diff. We could also consider a similar training paradigm, in which we omit the zeroth order contribution, which we called centered training,
\begin{align*}
    &f_{\theta_{t +\, \tau }}(x) - f_{\theta_{\tau}}(x) \approx (\theta_{t\, + \tau} - \theta_{\tau})^T \cdot \nabla_{\theta_{\tau}}f_{\theta_{\tau}}(x)
    \end{align*}
    \begin{equation}
    \label{centered_learning}
    \rightarrow f_{Centered,\, \triangle \theta,\, \theta_{\tau}}(x) = \triangle \theta^T \cdot \nabla_{\theta_{\tau}}f_{\theta_{\tau}}(x)
\end{equation}

Note that in centered training, we have two free parameters - $\triangle \theta$, and $\theta_{\tau}$ . $\theta_{\tau}$ defines our feature map, while $\triangle \theta$ is our linear weight (The Shared-base weight). 
As far as we know, no one has considered training these two parameters simultaneously, as $\theta_{\tau}$ is typically fixed, but we will be learning both here.

\noindent{\textbf{Why Finite-Width NTKs matters?}}
As mentioned earlier, finite-width NTKs can benefit from feature learning. Feature learning allows networks to quickly learn without much modification of parameters, like meta-learning.
The Functa paper \ref{NTK_Functa} has a network with shared weights $\theta_w$ , and per-image biases $\theta_b^i$ . While this may be motivated by the need to reduce the dimensionality of the new data samples representations (As the training-set for the diffusion model), it additionally has the benefit of leverage the shared learned features in $\theta_w$ .
Contrasting this with the infinite-width Functa idea - there is no feature learning (As can bee seen from the results in Section \ref{NTK_Functa}). That is, we are modeling the $NF$ training process as training infinite width models on each image independently with the frozen infinite-width NTK feature set, while Functa is modelling training NFs jointly, with weight sharing in $\theta_w$ . However, this may be too inflexible, ideally, we want to adjust all the weights of the network, but only adjust them slightly so that we are approximately in the fine-tuning regime. The problem is that there are too many parameters to do this.
This highlights the main difficulty in balancing Functa and the Infinite-width Functa. Infinite-width methods give a tractable way to have more parameters by working in kernel space, but cannot benefit from feature learning. In contrast, Functa has too few parameters, but gets feature learning, and we cannot increase the number of free parameters in a tractable way.

\noindent{\textbf{MetaQuill Core Algorithm}} \label{MetaQuill_Core_Algorithm}
To fix this, we propose the MetaQuill Algorithm, which leverages the Feature Learning of neural networks and the parameter flexibility of NTK kernels while achieving meta-efficiency using only a fraction of the dataset. The goal of this algorithm is to integrate Finite-Width NTK, Model Feature Learning, and a novel approach that retains the benefits of Model Agnostic meta-learning (MAML) but with greater efficiency.

We demonstrate our findings on the Functa problem example. Yet, the algorithm is general and holds for any other variate of NNs and downstream tasks.

With regards to Functa,
This will become MetaQuill Functa, which uses the Kernel Trick with a learned feature map. Specifically, from Eq. \ref{centered_learning} we have the following model for each NF in the dataset (indexed by i),

\begin{equation}
\label{NF_i}
    NF_i(x) = \triangle \theta_i^T \cdot \nabla_{\theta_S} f_{\theta_S}(x)
\end{equation}

That is, we have per-image weights $\triangle \theta_i$ , but a shared feature map given by $\theta_S$ . Now we want to define a diffusion model on the per-image parameters $\triangle \theta_i$ , but that has high dimensionality. To work around this, we use the kernel trick, and define the weight-to-kernel space transform (and the reverse transform).
\textbf{Transforming from weight space to kernel space (and vice-versa)}. To proceed, we first consider the task of training an NF on a fixed set of input points $X$. Let $Y_i$ be the labels for a specific data point. We assume we are working in the Fourier embedding space, and also that these are shared for every image. Our image NF loss is,

\begin{align*}
    \mathcal{L}_i = \sum_{x^p,\, y^p\, \in\, X,\, Y_i} \frac{1}{2} \cdot (NF_i(x^p) - y_i^p)^2
    \end{align*}
    \begin{equation}
        \label{MetaQuill_NF_Loss}
    \rightarrow \mathcal{L}_i = \sum_{x^p,\, y^p\, \in\, X,\, Y_i} \frac{1}{2} \cdot (\triangle \theta_i^T \cdot \nabla_{\theta_S} f_{\theta_S}(x^p) - y_i^p)^2
\end{equation}

Where $p$ indexes the positions in the image (So, $x^p$ would be $p^{th}$ coordinate which corresponds to $p^{th}$ pixel in the label-image, and $y_i^p$ would be the $p^{th}$ pixel of the $i^{th}$ image).
Solving for the optimal $\triangle \theta$, we have,

\begin{equation}
\label{delta_theta_i}
    \triangle \theta_i = \sum_{w_i^p,\, x^p} w_i^p \cdot \nabla_{\theta_S} f_{\theta_S}(x^p)
\end{equation}

That is, the optimal $\triangle \theta$ is given by a linear combination of gradients at $X$. These $w_i^p$ values have a closed form solution, $w_i = K_{\theta_S}^{-1} \cdot y_i$ .

Where $w_i$ is the vector containing $w_i^p$, and $K_{\theta_S}$ is the finite-width NTK kernel matrix at the index points $X$, with finite NTK kernel $K_S(x, x') = \nabla_{\theta_S}f_{\theta_S}(x)^T \cdot \nabla_{\theta_S}f_{\theta_S(x')}$ .
This has lower dimensionality than $\triangle \theta$ , so we can create a data-set and train the diffusion model in this space.
This allows for us to transform from $\triangle \theta$ space to $w$ space by multiplying by the sum of input point gradients.
Also, assuming that $X$ is the same for every data point (i.e. every image has the same size, and is sampled at the same points), then we only need to compute $K_{\theta_S}$ and its inverse once.

\noindent{\textbf{MetaQuill Method}. }
The Algorithm's Method (For MetaQuill Functa) is as such,

\begin{enumerate}
    \item Train INRs for the training set with shared/per-image parameters given by Eq. \ref{MetaQuill_NF_Loss}. Additionally, add $L_2$ penalties $\triangle \theta_i$ so that the shared feature map does most of the work. The goal of this is to obtain $\theta_S$ . This can be done with only a subset of the data.
    \item Compute $K_{\theta_S}$ using the $\theta_S$ recovered from step 1.
    \item Compute per-image $w_i$ for the rest of the training-set.
    \item Train a diffusion model on those $w_i$s .
To generate novel models and hence images or scenes, we transform back to weight-space for fast forward passes (so that we could sample a lot of data points without compute more finite-NTKs which are slow),
\item Synthesize a new $\hat{w}$ using the diffusion model.
\item Compute $\triangle \theta_i = \sum_{\hat{w}_i^p,\, x^p} \hat{w}_i^p \cdot \nabla_{\theta_S} f_{\theta_S}(x^p)$ to transform from kernel space to weight space (Eq. \ref{delta_theta_i}).
\item Sample new points (e.g. for up-resolution) using $ NF_{new}(x) = \hat{\triangle \theta_i^T} \cdot \nabla_{\theta_S} f_{\theta_S}(x)$ (Eq. \ref{NF_i}).
\end{enumerate}

Formally,
\begin{algorithm}[H]
\caption{MetaQuill.}\label{alg:alg1}
\begin{algorithmic}
\STATE 
\STATE {\textsc{TRAIN}} $INR_{\theta_S}$
\STATE \hspace{0.5cm}
    $\theta_S \leftarrow  {L}_i = \sum_{x^p,\, y^p\, \in\, X,\, Y_i} \frac{1}{2} \cdot (\triangle \theta_i^T \cdot \nabla_{\theta_S} f_{\theta_S}(x^p) - y_i^p)^2\, +\, \norm{\triangle \theta_i}^2$
\STATE \hspace{0.5cm} $K_{\theta_S} \leftarrow NTK_{Finite}(\theta_S)$
\STATE \hspace{0.5cm} $w_i = K_{\theta_S}^{-1} \cdot y_i$
\STATE 
\STATE {\textsc{Generate new NFs}}
\STATE \hspace{0.5cm} $\hat{w} \leftarrow DM\, Inverse$
\STATE
\STATE {\textsc{Transform from Kernel to weight space}}
\STATE \hspace{0.5cm} $\triangle \theta_i = \sum_{\hat{w}_i^p,\, x^p} \hat{w}_i^p \cdot \nabla_{\theta_S} f_{\theta_S}(x^p)$
\STATE
\STATE {\textsc{sample a new NF}}
\STATE \hspace{0.5cm} $ NF_{new}(x) = \hat{\triangle \theta_i^T} \cdot \nabla_{\theta_S} f_{\theta_S}(x)$
\end{algorithmic}
\label{alg1}
\end{algorithm}

\paragraph{\textbf{Results}}
When considering the following novel algorithm for our Neural Field's Inpainting task, the practical experiment as follows,

Training a shared NF for a single image (See Fig. \ref{Training_Shared_NF}, top row - digit 5).
When training a Multi-NF on a subset of the training dataset (100 images in our example), one shared NF ($NF_S$) learning both $\theta_S$ and $\triangle \theta_i$ simultaneously for 100 images (See Fig. \ref{Training_Shared_NF}, bottom section).

After achieving the shared NF base parameters $(\theta_S)$, achieved via Eq. \ref{MetaQuill_NF_Loss}. We train a new NF from the shared obtained $NF_S$ model, creating $NF_i$ (For random image $i$) to obtain $\triangle \theta_i$ (See Fig. \ref{Train_NFi_Reconstruct_via_w}, top section - digit 7).

We can see that so far, the algorithm is working excellently, yielding high accuracy results, all with efficient calculations, run-time, and resources (only 100 images of the original    samples dataset).
Furthermore, as explained in Appendix \ref{MetaQuill_Core_Algorithm} - \textit{MetaQuill Core Algorithm} , we can notice that the algorithm achieves similar MAML meta-learning method, yet, without its (MAML's) inner loop. It makes MAML inner loop process redundant, which is a huge leap. This leads to a great efficiency in run-time space.

For the next phases (After calculating $K_{\theta_S}$) of calculating the $w_i$ vector (considering its corresponding $Y_i$ sample image), we're using Finite-width NTK algorithm. The results can be seen in Fig. \ref{Train_NFi_Reconstruct_via_w}, bottom section.

We can see that in contrary of the infinite width and time NTK's $W_i$ vector calculation (where there's no feature learning and the parameters of the model aren't taken under consideration), when using the Finite-NTK with Feature Learning, the $W_i$ vector loses representation power. When exploring why is that happening, it was discovered that the Kernel matrix ($K_{\theta_S}$ calculated via Finite-NTK on NFs) is ill-conditioned (with high condition number). This means, such a matrix is almost singular, and the computation of its inverse, or solution of a linear system of equations is prone to numerical errors. Thus, leading to blurry reconstructed images.

From the conducted experiments we found that though our novel suggested algorithm achieved efficient feature-learning and the desired results. Due to NTK's insufficient representational nature, even in the Finite-Feature Learning NTK approach, when applying the NTK method (In order to use it as a multi-model NF replacement. See Section \ref{NTK_Functa}), the kernel matrix suffers from high condition number (Being close to an Ill-condition matrix. Thus, approximating it to singularity). Therefore, leading to a blurry and inaccurate image reconstruction. Hence, we derive that while NTK could be used as a mean for approximately mimicking NN Model's evolution, at its core it lacks the ability to truly distillate the features which were learned in the training phase of the model and grasping the distribution of the dataset needed for the different desired tasks (It lacks the ambient spanning base representation of NN mapping functions space - Here, trained shared Neural Field model space). A major factor affecting this, is NTK's Representational Linearity (Linearity of the $W$ vector. See Section \ref{NTK_Natural_Limitations} - \textit{NTK Representational Linearity}), which the $NF\, KIP$ algorithm (See Section \ref{Neural Field Kernel Inducing Points (NTK-KIP)}) addresses and solves.
Therefore, we believe that a combined algorithm of the two novel suggested algorithms $(NF\, KIP,\, and\, MetaQuill)$
can potentially achieve groundbreaking results, gaining NTK Learning capabilities, and Non-linear NTK representation mapping, all while achieving Efficiency, and Meta-learning.

To coclude,
\textbf{NTK Algorithm which suggests a new Efficient meta-learning approach.} This algorithm tackles a major issue regarding the NTK method, NTK Learning and Feature Extraction. Thus, giving the NTK method an actual feature learning capabilities.
All while suggesting a new efficient Meta-Learning approach which is similar to MAML \cite{finn2017model}, yet, without the inner loop phase.
We found this very promising as we managed to train a shared NF model with only a subsection of the data (100 images), and adapt it to new samples straight from the first gradient update.

The above algorithm though looks promising, still suffers from the NTK's Representational Linearity problem, and from an Ill-condition matrix (With relatively high condition number), resulting in blurry reconstruction images.
Thus, we believe a unified algorithm which combines the above two novel approaches, can potentially achieve a new novel NTK method which benefits from Learning capabilities, Non-linear representation, and a much more Efficient core representation.

We hope these findings not only advance understanding in NTK-based representation learning but also inspire the research community to develop more robust theoretical frameworks and innovative methods. By addressing key limitations and opening new avenues for exploration, we aim to spur future work that further strengthens and expands the theoretical foundations of representation learning.

\subsubsection{\textbf{MetaQuill-KIP Algorithm}} \label{app:mqkip_method}
MetaQuill-KIP represents each task $i$ by a shared initialization $\theta_S$ and a compact task specific offset $\Delta\theta_i$, so that $\theta_i = \theta_S + \Delta\theta_i$.
At test time, instead of training a full INR from scratch, we optimize only $\Delta\theta_i$ while keeping $\theta_S$ fixed.

\paragraph{Tangent (NTK linearized) prediction}
Given coordinates $x$ (Fourier positional encodings in our experiments), the linearized prediction around $\theta_S$ is
\[
\hat{y}_{\text{tan}}(x) = J_{\theta_S}(x)\,\Delta\theta_i,
\]
where $J_{\theta_S}(x)$ is the Jacobian of the INR output with respect to parameters at $\theta_S$.
This is exactly the tangent output used in our implementation. 

\paragraph{KIP warm start for $\Delta\theta_i$}
To obtain an informative initialization for $\Delta\theta_i$, we run a KIP style inducing point optimization on a single target image.
We optimize a compact support set of coordinates $X_S$ (and optionally pseudo labels $Y_S$) so that a kernel ridge regression solve on that distilled support predicts the target well.
For inpainting, the KIP objective is restricted to observed pixels only, matching the inpainting protocol.
This stage yields a task specific offset $\Delta\theta_i^{(0)}$ that is already structure aligned, but still limited by the linearized NTK regime.

\paragraph{Nonlinear refinement around $\theta_S$}
Starting from $\Delta\theta_i^{(0)}$, we refine only $\Delta\theta_i$ using the full nonlinear model output
\[
\hat{y}_{\text{nonlin}}(x) = f_{\theta_S+\Delta\theta_i}(x),
\]
and optimize a masked reconstruction loss on the observed pixels (plus a mild $\ell_2$ regularizer on $\Delta\theta_i$).
This final stage injects true nonlinearity while preserving fast adaptation, since only $\Delta\theta_i$ is updated.

\paragraph{Why robustness matters}
In real sensing pipelines, the observed pixels can be corrupted by sensor noise or impulse artifacts.
We therefore evaluate robustness when only the observed pixels are perturbed, while the target ground truth remains clean.
To isolate observation noise from the intrinsic inpainting difficulty, we keep the evaluation target fixed and corrupt only the observed pixel values supplied to each method. We focus on two common sensing abstractions: additive Gaussian noise and impulse salt and pepper corruptions, both applied only on the observed set while leaving the hole unobserved.

\begin{figure*}[t!]
\centering
\includegraphics[width=0.98\textwidth]{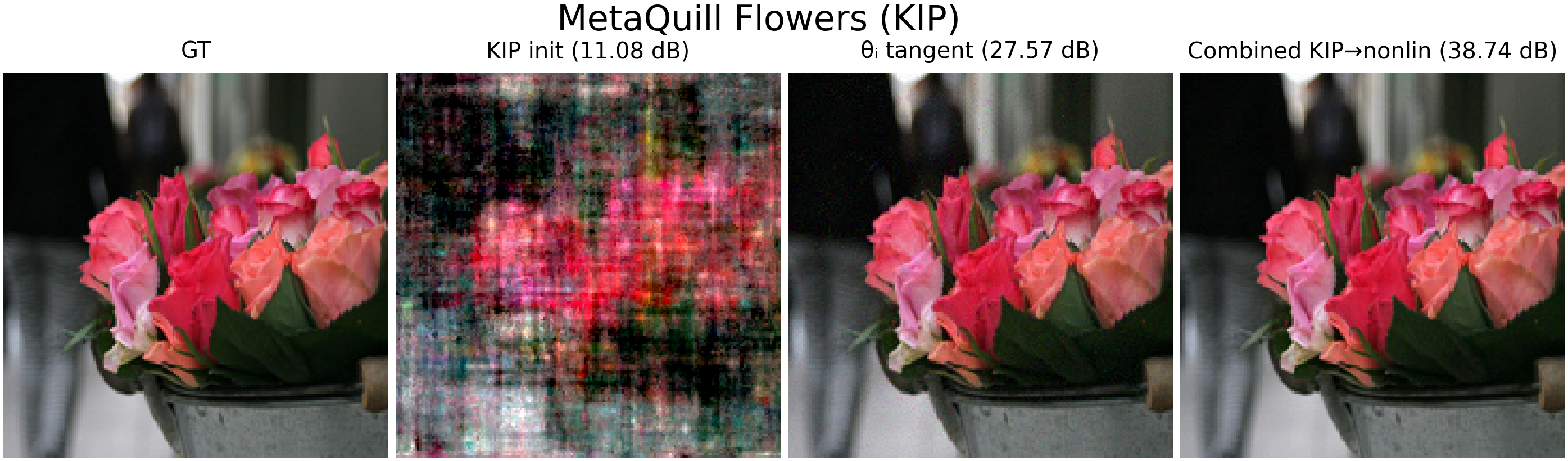}
\caption{Robust inpainting qualitative overview on Flowers. From left to right: ground truth, KIP init (PSNR $11.08$ dB), $\theta_i$ tangent only (PSNR $27.57$ dB), and Combined(KIP$\rightarrow$nonlin) (PSNR $38.74$ dB). Here $\theta_i=\theta_S+\Delta\theta_i$ and only $\Delta\theta_i$ is adapted while $\theta_S$ is fixed.}
\label{fig:mqkip_robust_overview}
\end{figure*}

\begin{figure*}[t!]
\centering

\makebox[0.48\textwidth]{\textbf{Gaussian} $\sigma=0.02$}
\hfill
\makebox[0.48\textwidth]{\textbf{Salt and pepper} $p=0.02$}

\vspace{0.4em}

\begin{minipage}[c]{0.04\textwidth}
\centering
\rotatebox{90}{\textbf{$\phi=0.10$}}
\end{minipage}
\hfill
\begin{minipage}[c]{0.94\textwidth}
\includegraphics[width=0.48\textwidth]{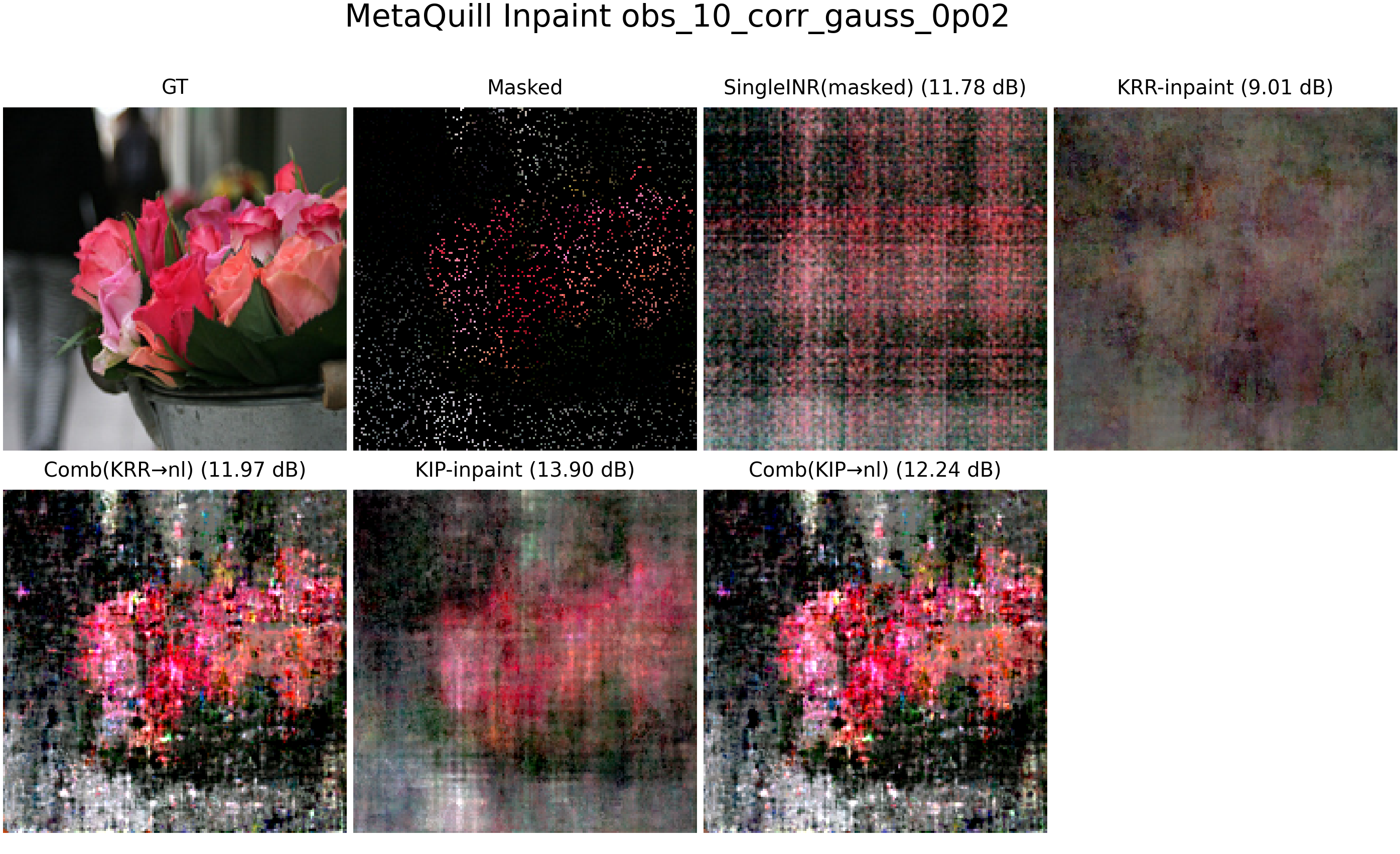}
\hfill
\includegraphics[width=0.48\textwidth]{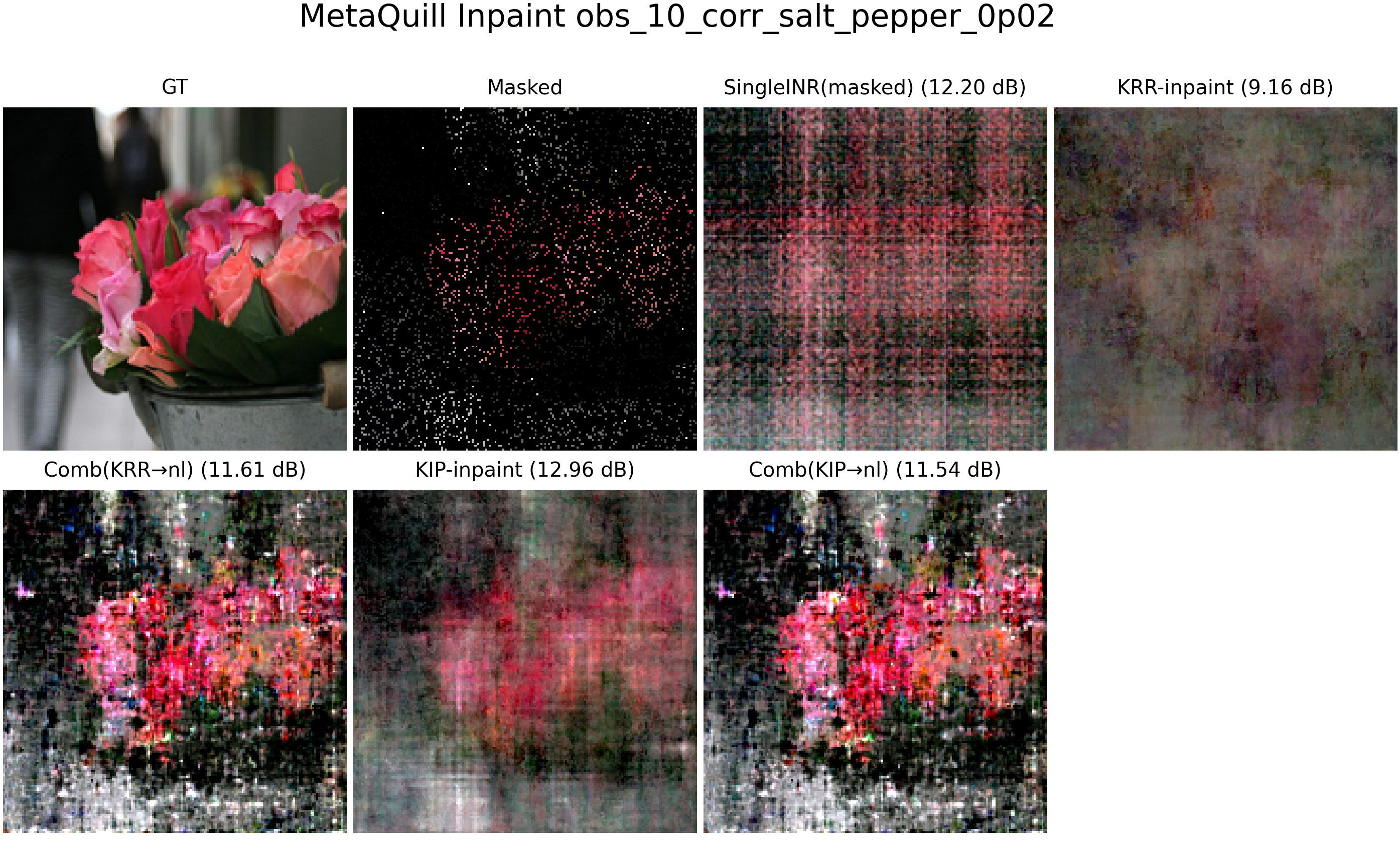}
\end{minipage}

\vspace{0.7em}

\begin{minipage}[c]{0.04\textwidth}
\centering
\rotatebox{90}{\textbf{$\phi=0.20$}}
\end{minipage}
\hfill
\begin{minipage}[c]{0.94\textwidth}
\includegraphics[width=0.48\textwidth]{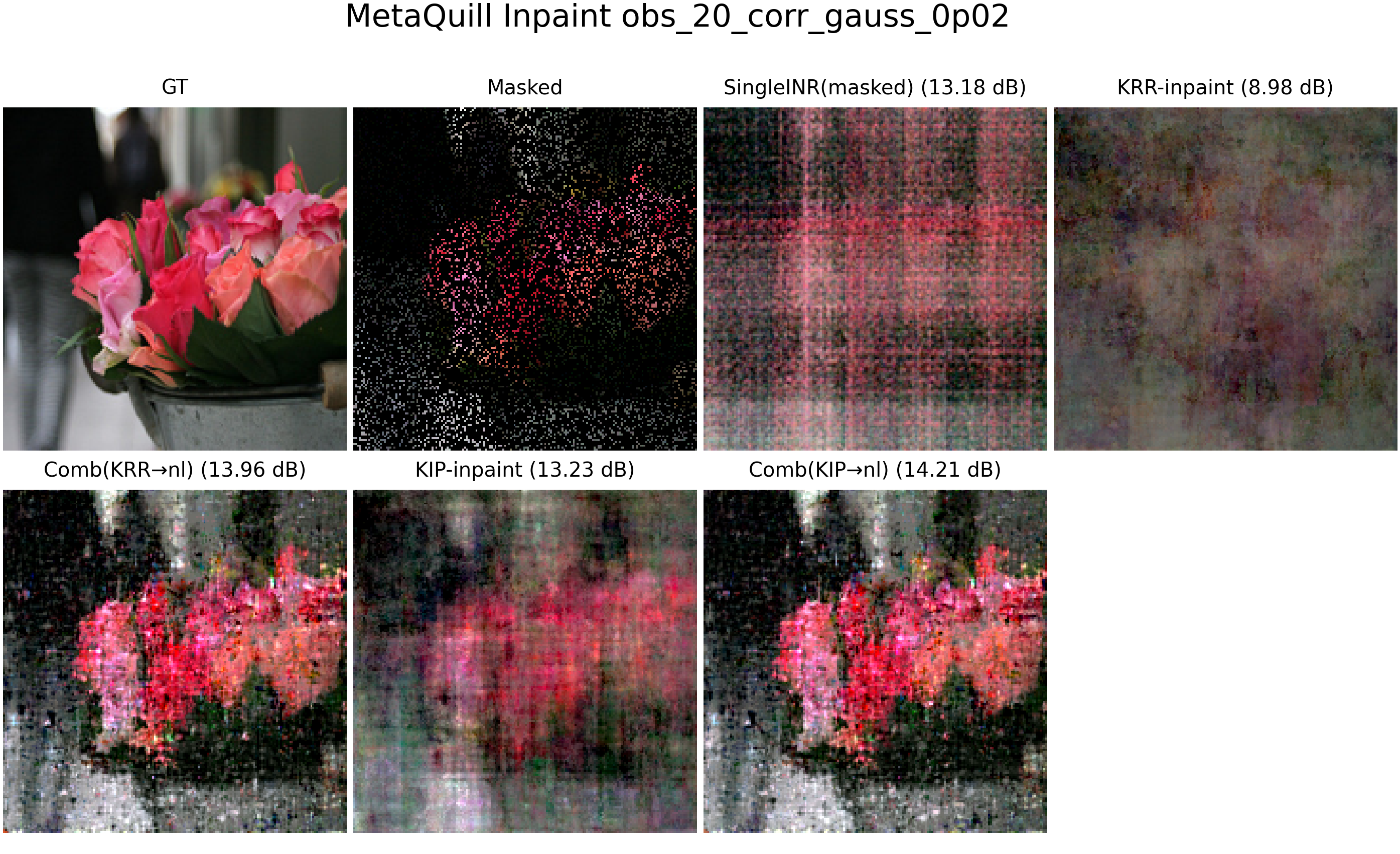}
\hfill
\includegraphics[width=0.48\textwidth]{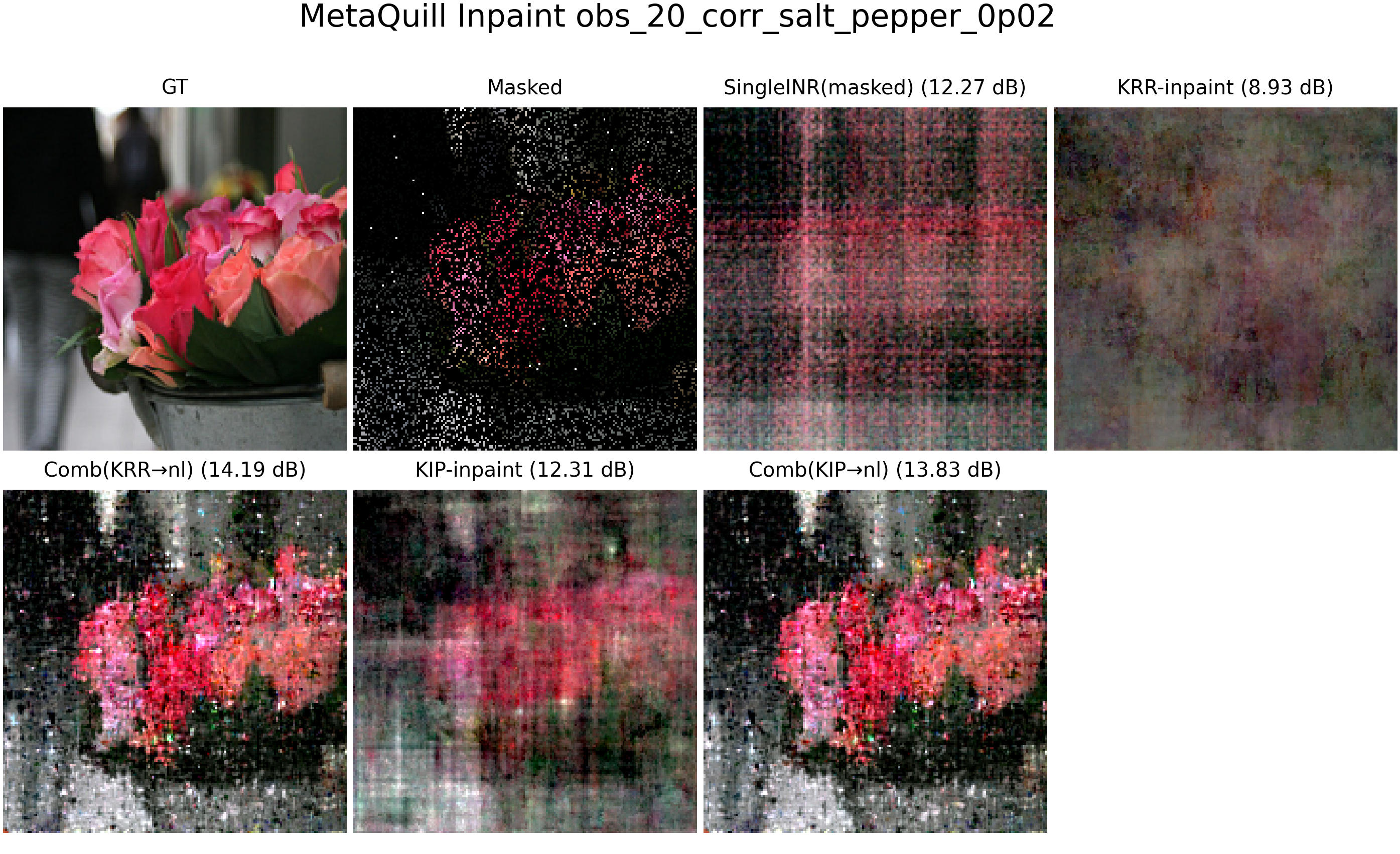}
\end{minipage}

\vspace{0.7em}

\begin{minipage}[c]{0.04\textwidth}
\centering
\rotatebox{90}{\textbf{$\phi=0.40$}}
\end{minipage}
\hfill
\begin{minipage}[c]{0.94\textwidth}
\includegraphics[width=0.48\textwidth]{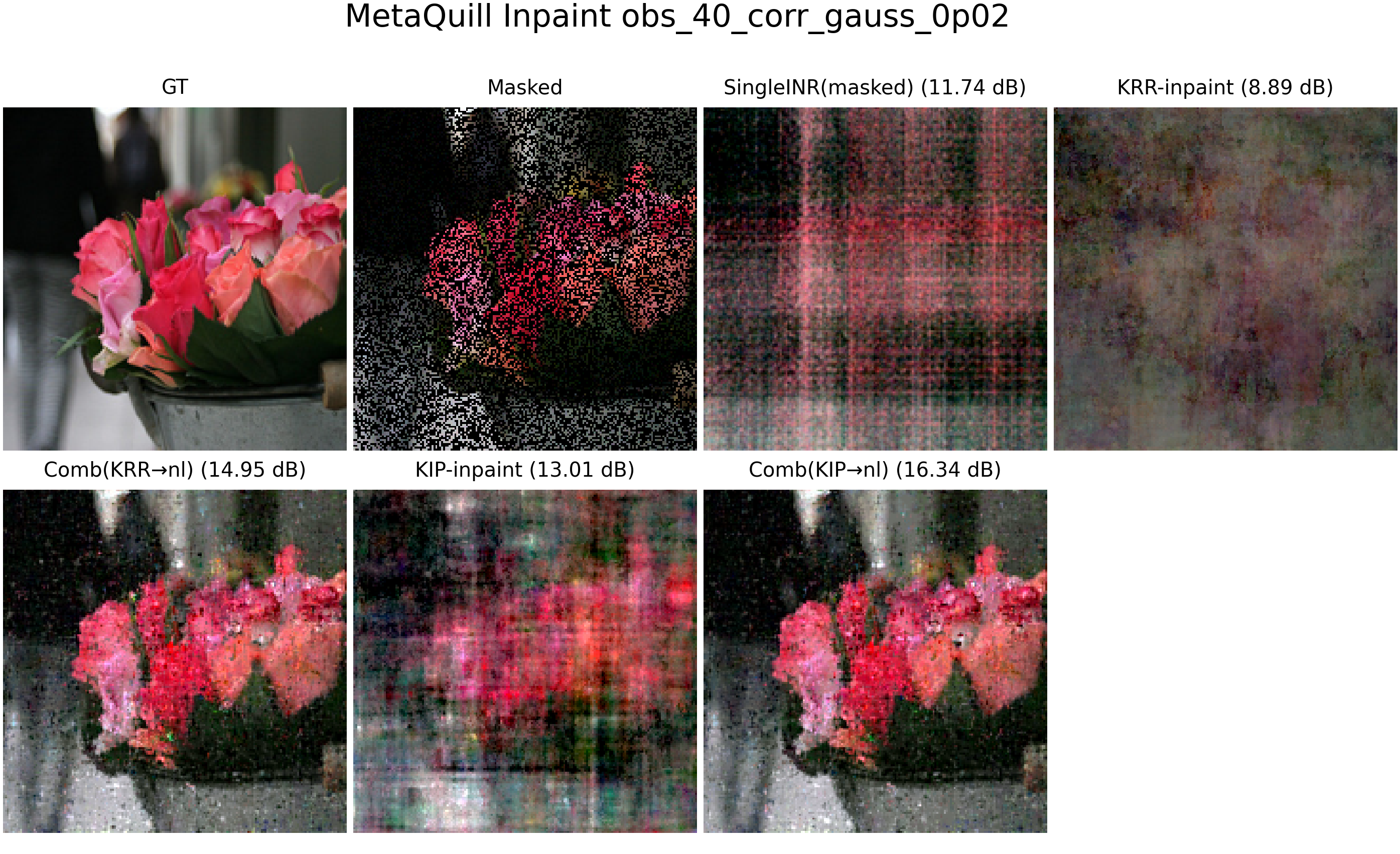}
\hfill
\includegraphics[width=0.48\textwidth]{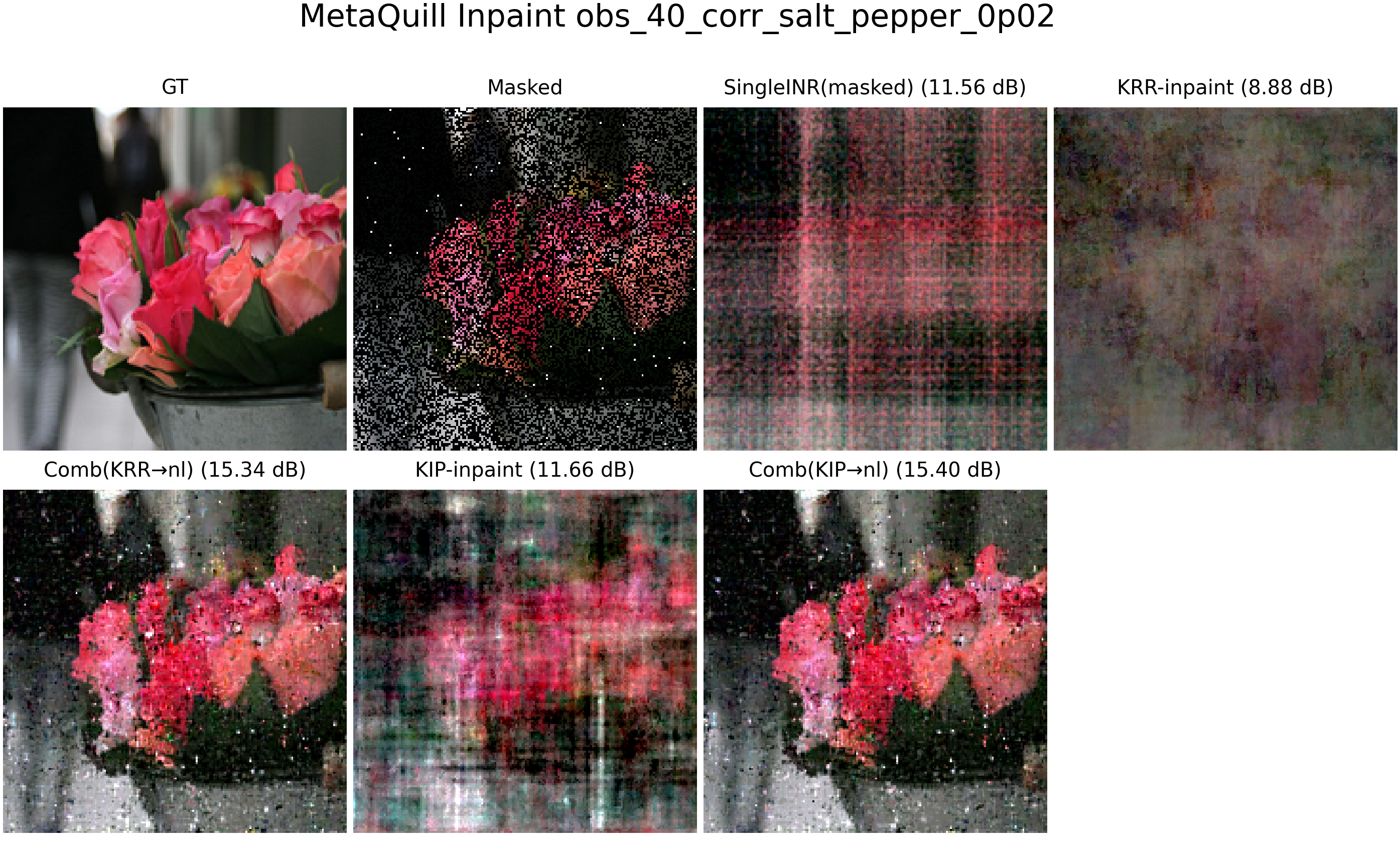}
\end{minipage}

\caption{Robustness to corruption of observed pixels on Flowers random mask inpainting. The left column uses Gaussian corruption with $\sigma=0.02$ and the right column uses salt and pepper corruption with $p=0.02$, both applied only on observed pixels. Each panel uses the same internal layout: first row from left to right is ground truth, masked input, Single-INR (masked), and KRR inpaint; second row from left to right is Combined(KRR$\rightarrow$nl), KIP inpaint, and Combined(KIP$\rightarrow$nl).}
\label{fig:mqkip_robust_grid_3x2}
\end{figure*}

\textbf{MetaQuill-KIP robustness under observed pixel corruptions. }
\label{app:mqkip_robust}
We evaluate robustness of inpainting when the observed pixels themselves are corrupted. This setting is practically relevant in sensor noise and imperfect measurements, and it complements the clean mask experiments in the main text. We use random masks with observed fraction $\phi \in \{0.10,0.20,0.40\}$ on Flowers, and we apply corruption only on the observed pixels. For Gaussian corruption we add i.i.d. noise $\epsilon \sim \mathcal{N}(0,\sigma^2)$ to observed pixels only, where $\sigma \in \{0.02,0.05,0.10\}$ is measured in the stored Flowers pixel scale $[-0.5,0.5]$ and values are clipped back to that range.
For salt and pepper corruption we flip an observed pixel to either $-0.5$ or $0.5$ with probability $p \in \{0.02,0.05,0.10\}$, again applied only on observed pixels.
Qualitative robustness examples are shown in Figure~\ref{fig:mqkip_robust_overview}, with additional snapshots at $\sigma=0.02$ and $p=0.02$ for $\phi \in \{0.10,0.20,0.40\}$ in Figure~\ref{fig:mqkip_robust_grid_3x2}.

We report PSNR on the full image as well as PSNR restricted to the hole. Both metrics follow our implementation convention of normalizing the compared region by its own intensity range before computing mean squared error, which can make the full image PSNR and hole PSNR numerically close in some regimes. We therefore use hole PSNR mainly as a complementary diagnostic and rely on the qualitative panels to assess structural plausibility under heavy sparsity and corruption.

Table~\ref{tab:mqkip_robustness} summarizes results for Single-INR (masked), NTK-KIP inpaint, and MetaQuill-KIP (Combined(KIP$\rightarrow$nl)). Across most corruption settings, MetaQuill-KIP degrades gracefully as noise increases and benefits more from higher observed fraction. In several low coverage cases, quantitative differences are small, while qualitative results indicate that MetaQuill-KIP preserves more globally coherent structure, consistent with the observation that PSNR alone can be misleading under extreme sparsity.

\begin{table*}[t!]
\centering
\setlength{\tabcolsep}{5pt}
\renewcommand{\arraystretch}{1.15}
\begin{tabular}{c l c c c}
\toprule
Obs.\ frac $\phi$ & Corruption on observed pixels & Single-INR(masked) & KIP-inpaint & Combined(KIP$\rightarrow$nl) \\
\midrule
\multirow{7}{*}{0.10} & Gaussian $\sigma=0.02$ & 11.78/11.63 & 13.90/13.87 & 12.24/12.19 \\
 & Gaussian $\sigma=0.05$ & 11.96/11.80 & 12.93/12.91 & 12.35/12.30 \\
 & Gaussian $\sigma=0.10$ & 11.32/11.18 & 12.74/12.72 & 10.71/10.68 \\
 & Salt and pepper $p=0.02$ & 12.20/12.04 & 12.96/12.93 & 11.54/11.50 \\
 & Salt and pepper $p=0.05$ & 11.81/11.64 & 12.82/12.80 & 10.50/10.46 \\
 & Salt and pepper $p=0.10$ & 11.75/11.57 & 12.81/12.78 & 11.21/11.16 \\
\midrule
\multirow{7}{*}{0.20} & Gaussian $\sigma=0.02$ & 13.06/13.01 & 13.41/13.40 & 13.56/13.55 \\
 & Gaussian $\sigma=0.05$ & 12.56/12.51 & 13.65/13.64 & 12.86/12.84 \\
 & Gaussian $\sigma=0.10$ & 12.12/12.08 & 12.62/12.61 & 12.51/12.49 \\
 & Salt and pepper $p=0.02$ & 12.15/12.10 & 12.48/12.48 & 13.19/13.17 \\
 & Salt and pepper $p=0.05$ & 12.55/12.50 & 13.18/13.17 & 13.18/13.16 \\
 & Salt and pepper $p=0.10$ & 12.05/12.00 & 11.44/11.43 & 11.26/11.24 \\
\midrule
\multirow{7}{*}{0.40} & Gaussian $\sigma=0.02$ & 13.16/13.24 & 14.34/14.34 & 17.38/17.29 \\
 & Gaussian $\sigma=0.05$ & 13.35/13.42 & 15.01/15.01 & 16.56/16.47 \\
 & Gaussian $\sigma=0.10$ & 13.18/13.26 & 14.31/14.31 & 14.26/14.21 \\
 & Salt and pepper $p=0.02$ & 12.97/13.05 & 12.98/12.99 & 16.43/16.35 \\
 & Salt and pepper $p=0.05$ & 12.86/12.95 & 14.14/14.14 & 15.76/15.69 \\
 & Salt and pepper $p=0.10$ & 12.73/12.82 & 13.51/13.51 & 14.73/14.67 \\
\bottomrule
\end{tabular}
\caption{Robustness to observed pixel corruptions on Flowers random mask inpainting. Each entry reports PSNR on the full image and PSNR restricted to the hole as psnr/psnr\_hole. Corruptions are applied only on observed pixels, while evaluation uses the clean ground truth.}
\label{tab:mqkip_robustness}
\end{table*}

\end{document}